\documentclass{article}

\usepackage{microtype}
\usepackage{graphicx}
\usepackage{booktabs} % for professional tables

\usepackage{hyperref}

\usepackage[accepted]{icml2026}

\usepackage{amsmath}
\usepackage{amssymb}
\usepackage{mathtools}
\usepackage{amsthm}

\usepackage[capitalize,noabbrev]{cleveref}

\theoremstyle{plain}
\newtheorem{theorem}{Theorem}[section]

\newtheorem{lemma}[theorem]{Lemma}
\newtheorem{corollary}[theorem]{Corollary}
\theoremstyle{definition}
\newtheorem{definition}[theorem]{Definition}
\newtheorem{assumption}[theorem]{Assumption}
\theoremstyle{remark}

\usepackage[textsize=tiny]{todonotes}
\usepackage{xspace}
\usepackage{bbm}
\usepackage{bbold}
\usepackage{enumitem}
\usepackage{tabularx}
\usepackage{multirow}
\usepackage{array}

\icmltitlerunning{Memory Savings at What Cost? A Study of Alternatives to Backpropagation}

\usepackage{url}
\usepackage[utf8]{inputenc}
\usepackage[T1]{fontenc}
\usepackage{amsfonts}
\usepackage{nicefrac}
\usepackage[table, dvipsnames]{xcolor}

\usepackage{tcolorbox}
\usepackage{diagbox}
\usepackage{caption}
\usepackage{subcaption}
\usepackage{lipsum}
\usepackage{wrapfig}
\usepackage{pifont}

\newcommand{\cO}{\mathcal{O}}

\newcommand{\cL}{\mathcal{L}}
\newcommand{\cN}{\mathcal{N}}

\newcommand{\bR}{\mathbb{R}}
\newcommand{\bE}{\mathbb{E}}
\newcommand{\bw}{\pmb{w}}
\newcommand{\bv}{\pmb{v}}
\newcommand{\by}{\pmb{y}}

\newcommand{\jvp}{\texttt{jvp}\xspace}
\newcommand{\fmad}{\textsc{FmAD}\xspace}
\newcommand{\rmad}{\textsc{RmAD}\xspace}
\newcommand{\cmark}{\textcolor{Green}{\ding{51}}}%
\newcommand{\xmark}{\textcolor{red}{\ding{55}}}%

\definecolor{pastelgreen}{RGB}{177, 237, 232}
\definecolor{pastelyellow}{RGB}{255, 239, 159}
\definecolor{pastelred}{RGB}{255, 166, 158}
\definecolor{darkgray}{RGB}{184, 189, 181}
\definecolor{lightdarkgray}{RGB}{210, 212, 200}
\definecolor{lightgray}{RGB}{229, 229, 229}
\newtcbox{\swatchbox}[1][]{on line, box align=base, nobeforeafter,
  width=1.5ex, height=1.5ex,
  boxrule=0.4pt, arc=1pt,
  colframe=black, colback=#1,
  left=0pt, right=0pt, top=0pt, bottom=0pt,
  valign=center, baseline
}

\begin{document}

\twocolumn[
\icmltitle{Memory Savings at What Cost?\\A Study of Alternatives to Backpropagation}



\icmlsetsymbol{equal}{*}

\begin{icmlauthorlist}
\icmlauthor{Kunjal Panchal}{umass}
\icmlauthor{Sunav Choudhary}{adobe}
\icmlauthor{Yuriy Brun}{umass}
\icmlauthor{Hui Guan}{umass}
\end{icmlauthorlist}

\icmlaffiliation{umass}{College of Information and Computer Sciences, University of Massachusetts, Amherst}
\icmlaffiliation{adobe}{Adobe, San Jose}

\icmlcorrespondingauthor{Kunjal Panchal}{kpanchal@umass.edu}

\icmlkeywords{Machine Learning, ICML}

\vskip 0.3in
]



\printAffiliationsAndNotice{}  % leave blank if no need to mention equal contribution

\begin{abstract}
    
Forward-mode automatic differentiation~(\fmad) and zero-order~(ZO) optimization are increasingly proposed as memory-efficient, backpropagation-free alternatives for large language model (LLM) fine-tuning, yet their benefits are typically evaluated only against standard backpropagation (BP), omitting memory-efficient variants such as activation checkpointing.
We present a unified theoretical and empirical comparison of BP, checkpointed BP, \fmad, and ZO for LLM and vision-language model training, showing that while \fmad and ZO reduce activation memory, they trade memory for higher computational cost and longer wall-clock time to convergence, resulting in lower accuracy and slower training, especially under constrained perturbation budgets.
Across models, BP with checkpointing outperforms \fmad and ZO variants, including variance-reduced methods, achieving up to 31.1\% higher accuracy, 34.8\% faster convergence, and 3.8$\times$ fewer computations at comparable memory usage, while also revealing instability-related failure modes in \fmad and ZO.
Overall, our results correct a one-sided benchmarking narrative by showing that memory-efficient methods entail fundamentally different trade-offs, and that ignoring these distinctions has led to misleading conclusions about LLM optimization in prior work.
Our source code is available at~\url{https://github.com/Astuary/Gradient_Estimation_Methods}.

\end{abstract}

\section{Introduction}
\label{sec:introduction}

Backpropagation~(BP)~\citep{rumelhart1986learning} is the standard algorithm for gradient computation in deep learning and remains the dominant approach for training and fine-tuning large language models (LLMs), due to its convergence efficiency and widespread support in automatic differentiation frameworks such as PyTorch~\citep{paszke2017pytorch} and JAX~\citep{bradbury2018jax}.
However, BP incurs high memory overhead when training large models such as LLMs, as it must store intermediate activations for the backward pass.
To address this limitation, recent research has explored alternative gradient estimation methods for LLM optimization, including forward-mode automatic differentiation (\fmad)~\citep{baydin2017autodiff, baydin2022gradients, panchal2024thinking} and zero-order (ZO) optimization~\citep{richardson1955finitedifferences, malladi2023mezo}, which approximate gradients using directional derivatives or finite-difference evaluations under random weight perturbations.
These methods are often promoted as memory-efficient or hardware-friendly alternatives to BP for LLM training and fine-tuning, particularly in resource-constrained or non-differentiable settings~\citep{panchal2024thinking, xu2024fwdllm, malladi2023mezo}.

% limitations of existing approach 
Despite growing interest, prior work on \fmad and ZO for LLM training and fine-tuning suffers from two critical limitations that leave their practical value inadequately understood.
First, existing comparisons~\citep{gautam2024mezosvrg, zhang2024revisitingzo} often overlook activation checkpointing~\citep{chen2016activationcheckpointing}, a widely supported and effective BP variant for large-scale models that substantially reduces memory usage by recomputing rather than storing intermediate activations.
Second, as shown in Table~\ref{tab:existing-metho-comparison}, key considerations for LLM optimization (such as computational cost and wall-clock time to convergence) are frequently omitted, leaving even comparisons against vanilla BP incomplete.
This one-sided narrative of ZO and \fmad as superior to BP in LLM settings motivates our study: we aim to provide a comprehensive account of these trade-offs, encompassing not only memory usage but also convergence speed and overall computational efficiency.

\begin{table*}[t]
% \vspace{-0.3cm}
    \centering
    \footnotesize
    \tabcolsep=0.05cm
    \rowcolors{2}{lightgray}{white}
    \caption{While some existing research empirically compares vanilla backpropagation (\textsc{BP-Vanilla}) across multiple metrics including memory usage, convergence time, and computational cost, they examine only a subset of these criteria, and notably, none include comparisons with backpropagation using checkpointing (\textsc{BP-Checkpointing}). We omit accuracy as it is evaluated in all the studies. }
    \renewcommand{\arraystretch}{1.2}
    \begin{tabular}{lccc>{\raggedright\arraybackslash}m{10.5cm}}
    \toprule 
        \textsc{Methods} & \begin{tabular}[c]{@{}c@{}}\textsc{Conv.}\\ \textsc{Time}\end{tabular} & \textsc{Memory} & \begin{tabular}[c]{@{}c@{}}\textsc{Comp.}\\\textsc{Cost}\end{tabular} & \quad  \textsc{Contributions} \\ 
    \midrule
        \begin{tabular}[c]{@{}l@{}}\textsc{MeZO}\\\citep{malladi2023mezo}\end{tabular}  & \xmark & \cmark & \xmark & ZO uses 12$\times$ less memory than Vanilla BP while achieving accuracy within 5\%.\\
        \begin{tabular}[c]{@{}l@{}}\textsc{MeZO-SVRG}\\\citep{gautam2024mezosvrg}\end{tabular}  & \xmark & \cmark & \xmark & Enhances the convergence accuracy of \textsc{MeZO} through variance reduction, improving accuracy by up to 20\%. 
        \\
        \begin{tabular}[c]{@{}l@{}}{Revisiting ZO}\\\citep{zhang2024revisitingzo}\end{tabular}  & \xmark & \cmark & \xmark & Benchmarks ZO optimization in LLM fine-tuning, along with proposing novel techniques that enhance accuracy over MeZO by up to 3\%. 
        \\
        \begin{tabular}[c]{@{}l@{}}\textsc{ZOSparse}\\\citep{guo2025zosparse}\end{tabular} & \cmark & \xmark & \xmark & ZO fine-tuning achieves full ZO accuracy by updating just 0.1\% of sensitive parameters, with up to 2.5× speedup.
        \\
        \begin{tabular}[c]{@{}l@{}}\textsc{Spry}\\\citep{panchal2024thinking}\end{tabular} & \cmark & \cmark & \xmark & Distributes trainable parameters across federated clients, improving \fmad's convergence speed by up to 20$\times$ and final accuracy by up to 13\% compared to ZO. \\
        \begin{tabular}[c]{@{}l@{}}\textsc{FoMoH}\\\citep{cobb2024secondorderforwardmodeautomaticdifferentiation}\end{tabular}  & \xmark & \xmark & \xmark & Introduces forward-mode second-order optimization; improves accuracy by 1--3\% compared to first-order \fmad on logistic regression and CNN tasks.
        \\
        This work  & \cmark & \cmark & \cmark &  First to evaluate how BP with checkpointing fares in the three-way tradeoff vs.\ variance-reduced ZO and \fmad. \\
    \bottomrule 
    \end{tabular}
    \label{tab:existing-metho-comparison}
\end{table*}

This paper addresses the above-mentioned limitations through a comprehensive study of BP, \fmad, and ZO approaches in the context of LLM training and fine-tuning.
We first outline the expected trade-offs among convergence behavior, memory consumption, and computational cost as functions of model dimensionality $d$ and the number of perturbations per iteration $n$, quantities that are particularly large in modern LLMs.
These theoretical results suggest that, while \fmad and ZO may reduce memory under certain regimes (e.g., when perturbations are evaluated sequentially), they face scalability challenges for LLM optimization, including higher per-iteration computational cost, $\mathcal{O}(nd)$, and slower convergence in high dimensions or under limited perturbation budgets.
In contrast, BP with activation checkpointing is expected to achieve favorable convergence with comparable memory usage.

\begin{figure}
    \centering
    \includegraphics[width=\linewidth]{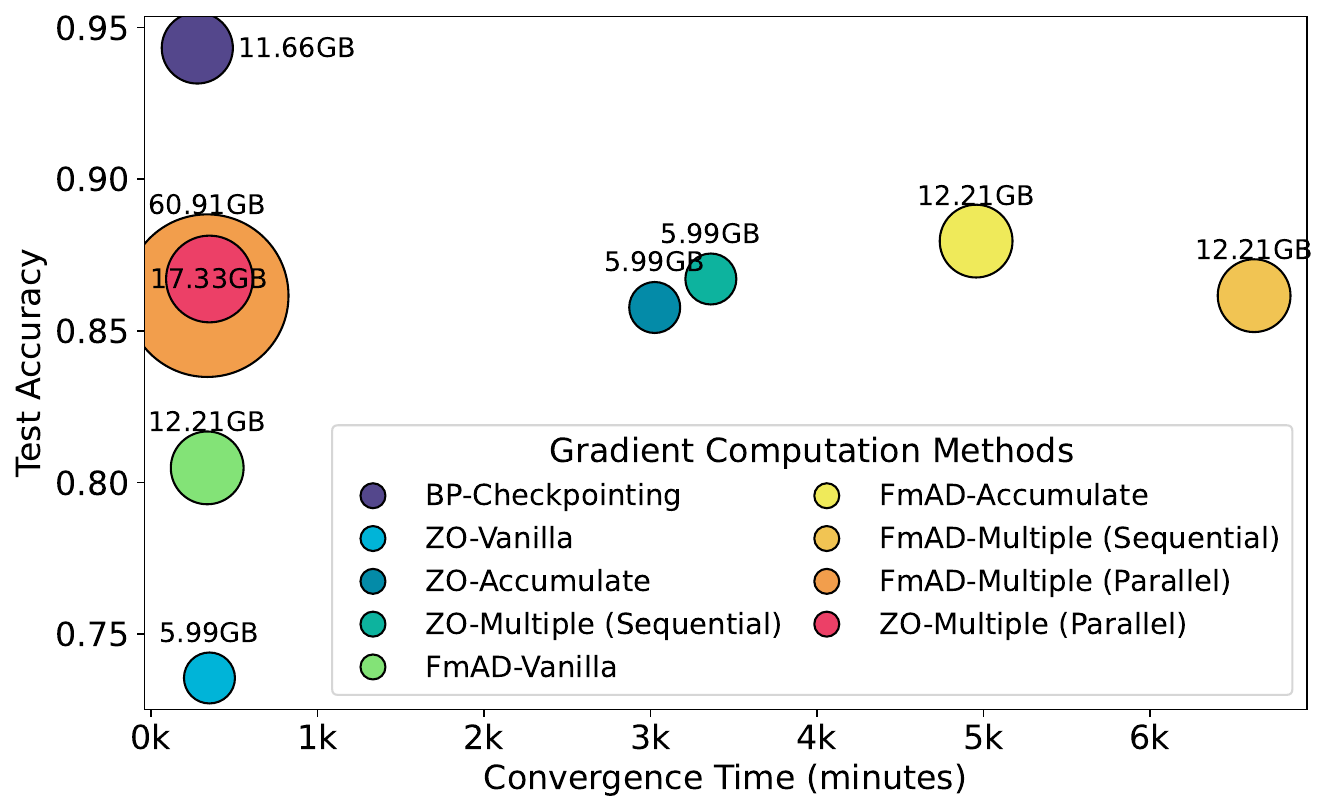}
    \caption{
    The three-way trade-off between accuracy, convergence time, and memory consumption during training of \textsc{Llama 3.1} (8B) on the AGNews dataset.
    The circle radii are proportional to the memory consumption.
    \textsc{BP-Checkpointing} achieves highest accuracy with lowest convergence time using comparable memory to \fmad and ZO variants. 
    \S~\ref{sec:empirical-evaluation} describes these methods in detail.}
    \label{fig:three-way-tradeoff}
    \vspace{-0.38cm}
\end{figure}

We then conduct extensive empirical evaluations on large language and vision-language models across text classification, text generation, and visual question answering tasks. 
We compare BP with checkpointing against a wide range of \fmad and ZO variants (including 
SVRG~\citep{liu2018zosvrg}, 
multiple perturbations per iteration~\citep{feng2024baffle}), 
and our own enhanced versions with variance reduction: gradient accumulation and adaptive perturbation sampling.
As illustrated in Figure~\ref{fig:three-way-tradeoff}, BP with checkpointing consistently achieves higher accuracy and faster convergence, while using memory on par with \fmad and ZO variants.

Beyond standard performance metrics (accuracy, memory, and convergence time), we also perform a dedicated study of specific failure modes in \fmad and ZO, focusing on instabilities in Jacobian-vector products ({\jvp}s) that can arise under adaptive optimizers and hinder convergence.  
This analysis provides insight into why these gradient estimation methods behave unpredictably in practice and complements our broader evaluation of their scalability and reliability.

These findings lead to a critical insight: 
despite recent enthusiasm for forward-mode and zero-order methods~\citep{panchal2024thinking, malladi2023mezo, liu2018zosvrg, gautam2024mezosvrg}, they remain fundamentally constrained by their inability to efficiently scale to large models. 
Rather than serving as alternatives to backpropagation, they operate as inefficient approximations that trade off accuracy or convergence speed for marginal memory reductions. 

This paper's main contributions are: 
\vspace{-0.3cm}
\begin{itemize}[leftmargin=0.35cm,nolistsep, noitemsep]
    \item A theoretical analysis of the convergence rates, memory cost, and compute complexity of BP, \fmad, and ZO under \textit{a common theoretical framework}, highlighting their three-way trade-offs for LLM training.

    \item A comprehensive empirical study across diverse LLM and vision-language tasks.
    We show that \textsc{BP} with checkpointing consistently achieves 4.5--31.1\% higher accuracy, 21.4--34.8\% faster convergence, and 3.2--3.8$\times$ lower computation cost than \fmad and ZO variants, while using comparable memory.

    \item The design and benchmarking of two new variance reduction methods for \fmad and ZO.
    These methods improve accuracy by 7.5--14.0\%, but still fall short of BP's overall efficiency, introducing overheads in either convergence time or memory.

    \item An analysis of \fmad's and ZO's failure modes, including high-dimensional perturbations, noisy {\jvp}s, and optimizer-dependent instabilities (e.g., abrupt \jvp spikes under adaptive optimizers like \textsc{AdamW}) that destabilize training and degrade convergence.

\end{itemize}

\section{Background}
\label{sec:background}

This section reviews three gradient computation techniques central to our study:
(a)~reverse-mode automatic differentiation (\rmad, of which backpropagation is a special case),
(b)~forward-mode automatic differentiation~(\fmad), and
(c)~zero-order~(ZO) finite-difference methods.
For an in-depth survey of these approaches, we refer readers to~\citet{baydin2017autodiff}. 
Appendix~\ref{sec:related-work} reviews related work in detail.
Details on signal propagation mechanism of these methods are in Appendix~\ref{adx:signal-propagation}.

The three methods described below operate on a function $f$, which in deep learning corresponds to a neural network and can be non-convex. 
This function $f$ is composed of nested functions $f_i$, $i\in [p]$, where $p$ is the number of layers given a neural network. Each nested function produces intermediate activations $y_i = f_i(w_i, y_{i-1})$, given weights $w_i$ and previous activations $y_{i-1}$, where $y_0 = x$ is the input.
The weights are represented by the vector $\bw = {w_1, w_2, \dots, w_p}$, where each $w_{[1, \dots, p]} \in \bR^{[m_1, \dots, m_p]}$. 
The total number of trainable parameter is $d = \sum_{i=1}^p m_i$.
The intermediate activations are $\by = {y_1, \dots, y_p}$.
The final output is $y = y_p = f(\bw, x) \in \bR^q$, where typically $q \ll m_i, \forall i \in [p]$. 
The loss function $\cL(y,\hat{y}) \in \bR$ measures the difference between the predicted output $y$ and the true target values $\hat{y}$. 

\textbf{Reverse-mode Auto Differentiation~(\rmad).}
\rmad computes gradients by propagating sensitivities 
(which is the rate at which the output of a function changes with respect to a given intermediate value)
backward through the neural network.
\rmad relies on vector-Jacobian product~(\texttt{vjp}), where the \emph{Jacobian} represents partial derivatives of an intermediate activation $y_i$ with respect to weights $w_{i-1}$, denoted $\frac{\partial y_i}{\partial w_{i-1}}$, and the \emph{vector} is the activation gradient $\frac{\partial f}{\partial y_i}$.  
\rmad starts by setting $\frac{\partial f}{\partial y_p}$ to 1, and propagating $\frac{\partial f}{\partial w_{i-1}} = {\frac{\partial f}{\partial y_{i}}}\frac{\partial y_i}{\partial w_{i-1}}$ and $\frac{\partial f}{\partial y_{i-1}} = \frac{\partial f}{\partial y_i} \frac{\partial y_i}{\partial y_{i-1}}$, for $i \in [2, p]$, backwards.
The final result is the weight gradient $\frac{\partial f}{\partial \bw}$, formed from a series of \texttt{vjp} computations.

Backpropagation~\citep{rumelhart1986learning} (BP) is a specific case of \rmad tailored for neural networks. 
While \rmad's backward pass begins by $\frac{\partial f}{\partial y_p}$ set to 1, BP initializes from the gradient of the loss function: $\frac{\partial \cL}{\partial y_p}$, which provides a semantically meaningful signal for optimization.  
The backward phase is preceded by a forward pass that computes the activations and the loss $\cL$.

\textbf{Forward-mode Auto Differentiation (\fmad).} 
\fmad propagates directional derivatives through the neural network to compute Jacobian-vector products (\texttt{jvp}). 
\fmad analyzes how a small perturbation $\bv$ in the weights $\bw$ affects the outputs. 
Starting from $\delta y_1 = \frac{\partial y_1 }{\partial w_1} v_1$, \fmad propagates changes forward as: 
{\footnotesize
\begin{align}
\delta y_i = \frac{\partial y_i}{\partial w_i} v_i + \frac{\partial y_i}{\partial y_{i-1}} \delta y_{i-1}, \;\; \text{for }i \in [2, p] \label{eq:fmad-one-layer-computations}
\end{align}}% 
until the final scalar perturbation in the loss $\delta \cL$ is computed.
Here, the \emph{Jacobian} term $\frac{\partial y_i}{\partial w_i}$ reflects sensitivity to weight changes, and the perturbation \emph{vector} is $v_i \in \bv$, where $\bv$ is typically sampled from $\cN(0, I_d)$.
The scalar $\delta \cL$ is referred to as the \jvp. Weight gradients (also called forward gradients) are computed as $\frac{\partial \cL}{\partial w_i} = \jvp \cdot v_i$.
In contrast to BP, which propagates $\frac{\partial \cL}{\partial y_i}$ backward ($i$ from $p$ to $1$), \fmad propagates $\frac{\partial y_i}{\partial w_j}$ forward ($i$ from $1$ to $p$, for all $j$). 

\textbf{Zero-order~(ZO) Finite Differences.}
ZO optimization estimates gradients using only function $f$ evaluations, with no first-order derivative information required. 
These methods, including finite differences~\citep{richardson1955finitedifferences, malladi2023mezo}, perturb the weights and approximate gradients through changes in the loss values of the perturbed function evaluations.
Given a perturbation direction $\bv \sim \cN(0, I_d)$, the gradient with respect to $w_i$ is approximated via: 
{
\footnotesize
\begin{align}
\frac{\partial \cL}{\partial w_i} \approx \frac{\cL(f(\bw + \epsilon \bv, x), \hat{y}) - \cL(f(\bw - \epsilon \bv, x), \hat{y})}{2\epsilon} \cdot v_i, \label{eq:zo-one-layer-computations} 
\end{align}
}%
where $\epsilon$ is a small step size. 
This symmetric difference estimator requires two sequential forward passes per perturbation direction.

\section{Convergence, Memory, and Compute Trade-offs}
\label{sec:theoretical-analysis}

We next review the theoretical characteristics of BP, \fmad, and ZO optimization, focusing on their convergence, memory, and computational profiles.
These methods have been analyzed individually in prior works~\citep{malladi2023mezo, gautam2024mezosvrg, chen2019zo, guo2025zosparse}, as well as in classic results on BP~\citep{bottou2018optimization, garrigos2024handbookconvergencetheoremsstochastic} and automatic differentiation.
The derivations of convergence bounds on a non-convex function $f$, for the three gradient computation approaches studied in this work are shown in Appendix~\ref{adx:proofs-convergence}. Analysis on computation complexity is in Appendix~\ref{adx:computational-complexity}. 
Here, we compile the theoretical results into a common comparative framework to highlight their trade-offs under shared assumptions. 

Table~\ref{tbl:theoretical-bounds} shows how convergence behavior is affected by key parameters, including the trainable parameter dimensionality $d$ and the number of perturbations per iteration $n$. 
Although \fmad and ZO can achieve memory savings in certain regimes, they incur higher per-iteration compute costs and slower convergence in high-dimensional or low-perturbation settings.
In contrast, BP (especially when paired with activation checkpointing) retains favorable convergence with competitive memory efficiency.
These theoretical results provide intuitions for our empirical analysis in \S~\ref{sec:empirical-evaluation}, where we quantify how these trade-offs manifest in large-scale training.
\begin{table*}[!h]
\centering
\footnotesize
\renewcommand{\arraystretch}{1.75}
\caption{
Big $\cO$ bounds of gradient computation methods on 
(a)~Convergence error,
(b)~Memory consumption, and
(c)~Compute cost (per-iteration).
Let 
$c$ denote the memory required to store activations for a single layer, 
$c_h$ the maximum per-layer activation memory, 
$p$ the total number of layers, and 
$d$ the number of trainable parameters.
While \textsc{BP} with checkpointing retains the fast convergence of BP with additional memory savings; both \fmad and ZO methods suffer from worse convergence and higher compute costs, with parallel variants further increasing memory consumption. 
}
\begin{tabular}{lccc}
\toprule
Method & Convergence Error & Memory & Compute \\
 \midrule
BP & \multirow{2}{*}{\begin{tabular}[c]{@{}l@{}}$\cO(1/T)$  with $\eta \leq \frac{1}{L}$\end{tabular}} & $\cO(cp)$ & $\cO(d)$ \\ \cmidrule{1-1}
\cmidrule{3-4}
\textsc{BP} (with checkpointing) &  & $\cO(c\sqrt{p})$ & $\cO(d \log p)$ \\
\midrule
\fmad (Parallel) & \multirow{2}{*}{\begin{tabular}[c]{@{}l@{}}$\cO\left(\frac{1}{T\left[ 1 - \frac{L \eta}{2} \left( 1 + \frac{d+1}{n}\right) \right]}\right)$ with $\eta < \frac{2}{L \left( 1 + \frac{d+1}{n}\right)}$\end{tabular}} & $\cO(nc_h)$ & $\cO(nd)$ \\ \cmidrule{1-1}
\cmidrule{3-4}
\fmad (Sequential) &  & $\cO(c_h)$ & $\cO(nd)$ \\
\midrule
ZO (Parallel) & \multirow{2}{*}{\begin{tabular}[c]{@{}l@{}} $\cO\left(\frac{1}{T\left[ 1 - \frac{L \eta}{2} \left( 1 + \frac{d+1}{n}\right) \right]}\right) + \frac{L d \eta^2}{2n} \cO(\epsilon^2)$ with $\eta < \frac{2}{L \left( 1 + \frac{d+1}{n}\right)}$
\end{tabular}} & $\cO(nc_h)$ & $\cO(nd)$ \\ \cmidrule{1-1}
\cmidrule{3-4}
ZO (Sequential) &  & $\cO(c_h)$ & $\cO(nd)$ \\
\bottomrule
\end{tabular}
\label{tbl:theoretical-bounds}
\end{table*}
We summarize theoretical comparisons of BP, \fmad, and ZO along three key axes:

\textbf{Observation 1: Accuracy.} \fmad and ZO introduce approximation noise and discretization effects, leading to higher convergence error than BP, especially in high-dimensional models or with limited perturbations.
\S~\ref{subsec:accuracy} empirically demonstrates that ZO suffers greater accuracy degradation than \fmad due to discretization error, and that both ZO and \fmad achieve lower accuracy than BP because of additional learning rate constraints, which are detailed in Appendix~\ref{adx:baselines-and-hyperparameters}.

\textbf{Observation 2: Convergence Speed.} Both \fmad and ZO require stricter learning-rate constraints than BP, which slows convergence as dimensionality grows or perturbation budgets shrink.
\S~\ref{subsec:wallclock-runtime} supports this observation by showing that \fmad and ZO converge more slowly and reach lower accuracy compared to BP.

\textbf{Observation 3: Memory-compute Trade-offs.} While able to reduce activation memory, \fmad and ZO incur $\cO(nd)$ compute cost per iteration, and face a fundamental trade-off: parallel perturbations reduce runtime but increase memory, whereas sequential perturbations conserve memory but slow training.
\S~\ref{subsec:memory-consumption} corroborates these memory bounds and shows a breakdown of the memory consumption.
\S~\ref{subsec:computation-cost} empirically validates the computation cost. 

\textbf{A Note on Non-differentiable and Black-box Settings.} 
While it's claimed that ZO has utility in settings with non-differentiable objectives~\citep{qiu2023nondiffzo, rando2023optimalnondiffzo} or limited model access~\citep{nikolakakis2022blackbox, lobanov2024blackbox}, its applicability to large-scale differentiable model training (including LLMs) is fundamentally constrained.
In true black-box scenarios, it is often infeasible to perturb weights or query loss values, making ZO methods impractical.
In contrast, first-order methods such as BP and \fmad require access to model internals and automatic differentiation support, challenges that are largely engineering in nature and increasingly well-supported by modern frameworks.
As such, the growing trend~\citep{gautam2024mezosvrg, guo2025zosparse} of applying ZO to train differentiable models like LLMs is misguided: the computational cost and degraded convergence significantly outweigh the memory gains.

\section{Empirical Evaluation}
\label{sec:empirical-evaluation}

This section empirically compares the variants of BP, \fmad, and ZO optimization. 
We evaluate these methods across multiple axes, including 
(a)~accuracy,
(b)~wallclock convergence time, 
(c)~memory consumption, and
(d)~computation cost.
For each of these dimensions, we also examine how different variance reduction strategies affect performance.
Last but not least, we empirically show that variance reduction methods and adaptive optimizers fail to make
\fmad and ZO converge reliably.

\subsection{Experimental Settings}
\label{subsec:experimental-setup}
\textbf{Datasets.}
We evaluate gradient computation methods across a diverse set of \textbf{5~text-based tasks} and \textbf{2~vision-based tasks}. 
The 5~text-based tasks are 
(a)~{Gsm8K} (text generation on math problems/next-word prediction)~\citep{cobbe2021gsm8k}, 
(b)~{MMLU} (multiple-choice question-answering covering various domains of knowledge)~\citep{hendryckstest2021mmlu}, 
(c)~{AGNews} (4-class news article text classification task)~\citep{zhang2015agnews}, 
(d)~{BoolQ} (boolean question-answering)~\citep{clark2019boolq}, and 
(e)~{MultiRC} (closed-book question-answering)~\citep{khashabi2018multirc}. 
The 2~vision-based tasks are both based on visual question-answering:
(a)~{VQAv2}~\citep{goyal2019vqav2}, and
% \textsc{GQA}~\citep{hudson2019gqa}, 
(b)~{TextVQA}~\citep{singh2019textvqa}.
Appendix~\ref{adx:datasets} describes the datasets in detail, including the train/test splits. 

\textbf{Models.}
Our evaluation uses 5 models with a varying number of total parameters (listed in parentheses).  
For text-based tasks, on the billion-parameters scale, we use \textsc{Llama~3.1} (8B)~\citep{grattafiori2024llama} 
and \textsc{OPT}~(1.3B, 6.7B, 13B)~\citep{zhang2022opt}.
Additionally, we include medium-sized language models  
\textsc{Bert} (110M, 340M)~\citep{devlin-etal2019bert}
and \textsc{RoBERTa} (125M, 355M)~\citep{liu2019robertarobustlyoptimizedbert} to analyze performance variations across model sizes.
For vision-based tasks, we use
\textsc{Qwen~2~VL} (7B)~\citep{qwen2025qwen25}. %,
To finetune these models, we use \textsc{QLoRA}~\citep{dettmers2023qlora}, where low-rank adapters are trainable while the rest of the weights are frozen and quantized to 4 bits. 
By default, we set the \textsc{LoRA} rank $r=1$ and scale $\alpha=1$ to minimize the number of trainable parameters for \fmad and ZO. 
Appendix~\ref{subsec:reducing_trainable_parameter_count} reports results on higher \textsc{LoRA} ranks and full finetuning setting. 

\textbf{Methods for Comparison.}
We categorize the 16 gradient computation methods our evaluation compares into three groups: 
(a)~\textbf{Backpropagation Methods:} 
\textsc{BP-Vanilla}~\citep{rumelhart1986learning} (the standard implementation that stores all intermediate activations), 
\textsc{BP-Checkpointing}~\citep{chen2016activationcheckpointing} (reduces peak memory consumption by storing only a subset of activations and recomputing the rest during the backward pass), 
and \textsc{BP-Accumulate} (uses gradient accumulation).
% to simulate large batch sizes under tight memory budgets). 
(b)~\textbf{Zero-order Methods}: 
\textsc{ZO-Vanilla}~\citep{chen2019zo} (use a single perturbation to estimate gradients as in Equation~\ref{eq:zo-one-layer-computations}), 
\textsc{MeZO}~\citep{malladi2023mezo} (incorporates a prompt-finetuning approach to convert classification tasks into next-word prediction with a constrained vocabulary),
\textsc{ZO-Accumulate} (applies gradient accumulation to reduce noise in gradient estimates),
\textsc{ZO-Multiple}~\citep{feng2024baffle} (averages gradient estimates from multiple perturbations per iteration to improve estimate stability), 
\textsc{ZO-Adaptive} (adaptively selects perturbation directions based on prior gradients), 
\textsc{ZO-SVRG}~\citep{liu2018zosvrg} (applies stochastic variance reduction to correct noisy gradients), 
and \textsc{ZO-Sparse}~\citep{guo2025zosparse} (only updates top-1\% parameters each iteration). 
(c)~\textbf{Forward-mode AD Methods:} 
\textsc{FmAD-Vanilla}~\citep{baydin2022gradients}, 
\textsc{FmAD-Accumulate},
\textsc{FmAD-Multiple}, 
\textsc{FmAD-Adaptive}, 
\textsc{FmAD-SVRG}, and
\textsc{FmAD-Sparse}. 
The \textsc{-Vanilla} suffix denotes the original implementation according to Equation~\ref{eq:fmad-one-layer-computations}, while the other variants mirror the corresponding ZO method in (b), adapting similar strategies for the forward-mode setting. 
Appendix~\ref{adx:baselines-and-hyperparameters} describes these methods and their hyperparameters in detail.

\textbf{Metrics.}
We evaluate the efficiency of the gradient computation methods using four metrics.
(a)~\textbf{Accuracy} at test-time assesses the efficacy of the learned models. 
(b)~\textbf{Wallclock convergence time} (in minutes) measures the time each approach takes to achieve stable-state accuracy.
(c)~\textbf{Peak memory consumption} (in GBs) quantifies the maximum memory consumed during training. 
(d)~\textbf{Computation cost} for each iteration and until convergence, in Tera Floating-Point Operations per Second (TFLOPs). 
Additionally, in our failure mode analysis, for ZO and \fmad approaches, Section~\ref{subsec:failure-mode-analysis} reports statistics, such as the mean of effective gradient norms and \jvp values across iterations, capturing the instability of estimated gradients and its impact on optimization dynamics.

\textbf{Libraries and Hardware.}
Our codebase is built using PyTorch~\citep{paszke2017pytorch}.   Quantization uses AutoGPTQ~\citep{frantar2022gptq}. 
We conducted all experiments involving billion-scale models across ZO and \fmad variants on a single Nvidia L40 GPU (48GB~RAM). 
For experiments on OPT (13B) model, we used one Nvidia A100 (80GB~RAM).
For \textsc{BERT} and \textsc{RoBERTa} models, we used an Nvidia 2080ti (11GB~RAM). 
We repeated each experiment three times with random seeds set to 0, 1, and 2 to ensure consistency and robustness.

\subsection{Comparison on Accuracy}
\label{subsec:accuracy}

\looseness-1
Accuracy is the primary metric of interest since any gradient computation method that reduces memory consumption or computational cost is of little practical value if it cannot match the predictive performance of BP.
Table~\ref{tbl:llama-accuracy} presents accuracy results and Appendix~\ref{adx:experimental_variance} shows 
variance across 3 executions. 

\begin{table*}[!t]
\centering
\footnotesize 
\renewcommand{\arraystretch}{1.2}
\caption{
BP, \fmad, and ZO variant accuracies (higher is better) across models and datasets. 
Subscripts show accuracy gaps from \textsc{BP-Vanilla/Chkpt} (\textsc{Chkpt} = \textsc{Checkpointing}). 
While BP remains the most accurate, \textsc{FmAD} and ZO variants \textsc{-Accumulate} and \textsc{-Multiple} offer notable gains over their \textsc{-Vanilla} forms but still lag behind BP, especially on generation tasks, such as GSM8K. 
Appendix~\ref{adx:experimental_variance} reports variance across runs.
Darker shade \swatchbox[darkgray] \phantom{} = range of high accuracies, 
lighter shade \swatchbox[lightgray] \phantom{} = range of moderate accuracies, 
unshaded = range of low accuracies.}
\begin{tabular}{l|lllll|ll}
\toprule
\multirow{2}{*}{\diagbox[width=\dimexpr \textwidth/5+8\tabcolsep\relax, height=0.94cm]{Method}{Model + Dataset}} & \multicolumn{5}{c}{\textsc{Llama 3.1 (8B)}} & \multicolumn{2}{|c}{\textsc{Qwen 2 VL (7B)}} \\
\cmidrule{2-8}
 & AGNews & BoolQ & MultiRC & GSM8K & MMLU & VQAv2 & TextVQA \\
\midrule
No Finetuning & {23.5} & {51.6} & {52.8} & {27.3} & {51.1} & {73.2} & {71.1} \\
\midrule
\textsc{BP-Vanilla/Chkpt} & \cellcolor{darkgray}{94.2} & \cellcolor{darkgray}{88.3} & \cellcolor{darkgray}{85.2} & \cellcolor{darkgray}{54.3} & \cellcolor{darkgray}{60.3} & \cellcolor{darkgray}{87.1} & \cellcolor{darkgray}{98.5} \\
\textsc{BP-Accumulate} & \cellcolor{darkgray}{93.8}\textsubscript{(-0.4)} & \cellcolor{darkgray}{87.9}\textsubscript{(-0.4)} & \cellcolor{darkgray}{83.3}\textsubscript{(-1.9)} & {33.1}\textsubscript{(-21.1)} & {53.8}\textsubscript{(-6.4)} & \cellcolor{darkgray}{86.3}\textsubscript{(-0.7)} & \cellcolor{darkgray}{97.1}\textsubscript{(-1.4)} \\
\midrule
\textsc{ZO-Vanilla} & {73.6}\textsubscript{(-20.6)} & {57.1}\textsubscript{(-31.1)} & {57.2}\textsubscript{(-28.0)} & \cellcolor{lightgray}{36.3}\textsubscript{(-17.9)} & {54.7}\textsubscript{(-5.6)} & {77.6}\textsubscript{(-9.4)} & {72.9}\textsubscript{(-25.6)} \\
\textsc{ZO-Accumulate} & \cellcolor{lightgray}{85.8}\textsubscript{(-8.4)} & \cellcolor{lightgray}{60.9}\textsubscript{(-27.3)} & \cellcolor{lightgray}{60.3}\textsubscript{(-24.8)} & {28.0}\textsubscript{(-26.3)} & \cellcolor{lightgray}{55.2}\textsubscript{(-5.1)} & {79.7}\textsubscript{(-7.4)} & {73.1}\textsubscript{(-25.4)} \\
\textsc{ZO-Multiple} & \cellcolor{lightgray}{86.7}\textsubscript{(-7.4)} & \cellcolor{lightgray}{60.0}\textsubscript{(-28.2)} & \cellcolor{lightgray}{61.0}\textsubscript{(-24.1)} & \cellcolor{lightgray}{35.8}\textsubscript{(-18.5)} & \cellcolor{lightgray}{56.8}\textsubscript{(-3.4)} & \cellcolor{lightgray}{81.5}\textsubscript{(-5.6)} & {74.7}\textsubscript{(-23.8)} \\
\textsc{ZO-Adaptive} & \cellcolor{lightgray}{81.5}\textsubscript{(-12.7)} & {57.4}\textsubscript{(-30.9)} & {59.0}\textsubscript{(-26.1)} & {30.2}\textsubscript{(-24.1)} & {52.6}\textsubscript{(-7.6)} & {79.5}\textsubscript{(-7.5)} & \cellcolor{lightgray}{79.1}\textsubscript{(-19.3)} \\
\textsc{ZO-SVRG} & \cellcolor{lightgray}{84.7}\textsubscript{(-9.5)} & \cellcolor{lightgray}{62.6}\textsubscript{(-25.7)} & \cellcolor{lightgray}{61.2}\textsubscript{(-23.9)} & {32.1}\textsubscript{(-22.2)} & \cellcolor{lightgray}{55.9}\textsubscript{(-4.3)} & {79.1}\textsubscript{(-7.9)} & {72.9}\textsubscript{(-25.6)} \\
\textsc{ZO-Sparse} & {64.5}\textsubscript{(-29.6)} & {53.2}\textsubscript{(-35.1)} & {55.3}\textsubscript{(-29.8)} & {29.1}\textsubscript{(-25.1)} & {51.4}\textsubscript{(-8.9)} & {78.6}\textsubscript{(-8.5)} & {73.8}\textsubscript{(-24.7)} \\
\textsc{MeZO} & \cellcolor{lightgray}{80.5}\textsubscript{(-13.7)} & {58.2}\textsubscript{(-30.1)} & \cellcolor{lightgray}{60.4}\textsubscript{(-24.8)} & \qquad --- & \qquad --- & \qquad --- & \qquad --- \\
\midrule
\textsc{FmAD-Vanilla} & \cellcolor{lightgray}{80.5}\textsubscript{(-13.7)} & \cellcolor{lightgray}{60.7}\textsubscript{(-27.6)} & \cellcolor{lightgray}{61.4}\textsubscript{(-23.8)} & \cellcolor{lightgray}{37.7}\textsubscript{(-16.6)} & \cellcolor{lightgray}{55.8}\textsubscript{(-4.5)} & \cellcolor{lightgray}{82.3}\textsubscript{(-4.8)} & \cellcolor{lightgray}{78.3}\textsubscript{(-20.2)} \\
\textsc{FmAD-Accumulate} & \cellcolor{lightgray}{88.0}\textsubscript{(-6.2)} & \cellcolor{lightgray}{70.3}\textsubscript{(-17.9)} & \cellcolor{lightgray}{71.2}\textsubscript{(-14.0)} & {30.8}\textsubscript{(-23.5)} & \cellcolor{lightgray}{57.1}\textsubscript{(-3.1)} & \cellcolor{lightgray}{83.7}\textsubscript{(-3.4)} & \cellcolor{lightgray}{80.9}\textsubscript{(-17.6)} \\
\textsc{FmAD-Multiple} & \cellcolor{lightgray}{86.2}\textsubscript{(-8.0)} & \cellcolor{lightgray}{64.4}\textsubscript{(-23.8)} & \cellcolor{lightgray}{65.4}\textsubscript{(-19.7)} & \cellcolor{lightgray}{40.5}\textsubscript{(-13.8)} & \cellcolor{lightgray}{57.7}\textsubscript{(-2.6)} & \cellcolor{lightgray}{82.9}\textsubscript{(-4.2)} & \cellcolor{lightgray}{79.1}\textsubscript{(-19.4)} \\
\textsc{FmAD-Adaptive} & {78.5}\textsubscript{(-15.7)} & {56.4}\textsubscript{(-31.9)} & {58.2}\textsubscript{(-27.0)} & \cellcolor{lightgray}{38.1}\textsubscript{(-16.2)} & \cellcolor{lightgray}{56.3}\textsubscript{(-3.9)} & \cellcolor{lightgray}{82.9}\textsubscript{(-4.1)} & \cellcolor{lightgray}{78.2}\textsubscript{(-20.3)} \\
\textsc{FmAD-SVRG} & \cellcolor{lightgray}{82.5}\textsubscript{(-11.7)} & \cellcolor{lightgray}{64.6}\textsubscript{(-23.7)} & \cellcolor{lightgray}{64.1}\textsubscript{(-21.0)} & \cellcolor{lightgray}{35.4}\textsubscript{(-18.9)} & \cellcolor{lightgray}{56.1}\textsubscript{(-4.2)} & \cellcolor{lightgray}{83.0}\textsubscript{(-4.0)} & \cellcolor{lightgray}{79.5}\textsubscript{(-19.0)} \\
\textsc{FmAD-Sparse} & {70.4}\textsubscript{(-23.8)} & {56.9}\textsubscript{(-29.4)} & {53.1}\textsubscript{(-32.1)} & {30.3}\textsubscript{(-23.9)} & {53.4}\textsubscript{(-6.8)} & {80.3}\textsubscript{(-6.7)} & \cellcolor{lightgray}{77.0}\textsubscript{(-21.5)} \\
\bottomrule
\end{tabular}
\label{tbl:llama-accuracy}
\end{table*}

\textbf{Backpropagation achieves significantly higher accuracy than \textsc{\fmad-Vanilla} and \textsc{ZO-Vanilla}.}
Backpropagation, both in its standard form (\textsc{BP-Vanilla}) and with checkpointing (\textsc{BP-Chkpt}), consistently achieves the highest accuracy across all tasks. 
Among the alternatives, the \textsc{-Vanilla} forms of \fmad and ZO are most directly comparable to BP. 
Due to the inherent randomness in their perturbation-based gradients (see \S~\ref{sec:theoretical-analysis} Obs~1), both \textsc{FmAD-Vanilla} and \textsc{ZO-Vanilla} lag behind BP by 4.5--27.5\% and 5.6--31.1\% across datasets, respectively.
Further, across all the datasets, \textsc{FmAD-Vanilla} outperforms \textsc{ZO-Vanilla}, with gains of 
1.1--6.9\%. 
This consistent margin illustrates \fmad's fundamental advantage: access to analytic first-order Jacobian-vector products (\jvp), over ZO's reliance on noisy finite-difference estimates.

\begin{figure}[!h]%{r}{0.65\textwidth}
    % \vspace{-0.55cm}
     \captionsetup[subfigure]{justification=centering}
     \centering
     \begin{subfigure}[b]{0.4\textwidth}
         \centering
         \includegraphics[width=\textwidth]{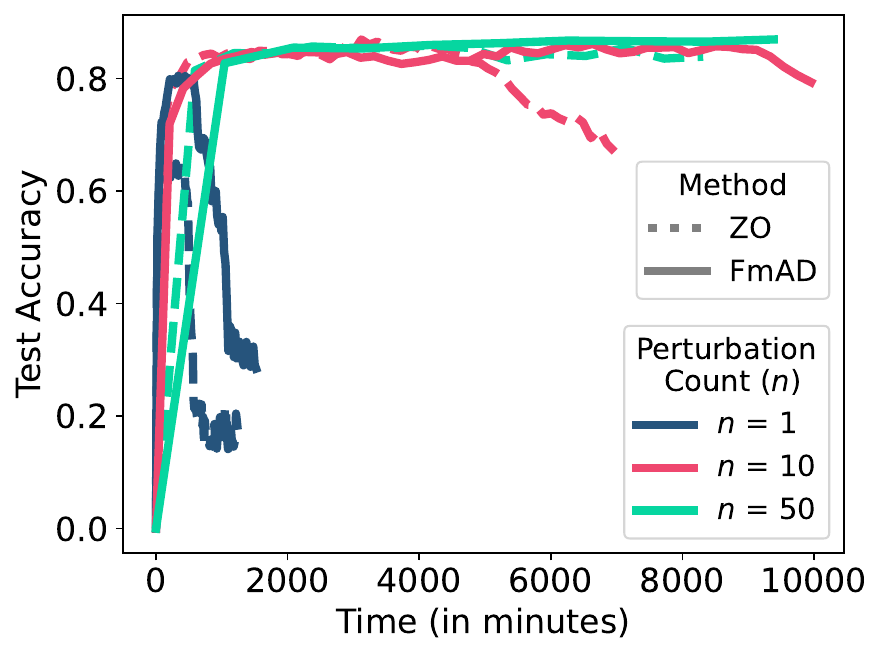}
         \caption{
          Averaging gradients over multiple perturbations per iteration
         }
         \label{fig:multiple-perturbations}
     \end{subfigure}
     \hfill
     \begin{subfigure}[b]{0.4\textwidth}
         \centering
         \includegraphics[width=\textwidth]{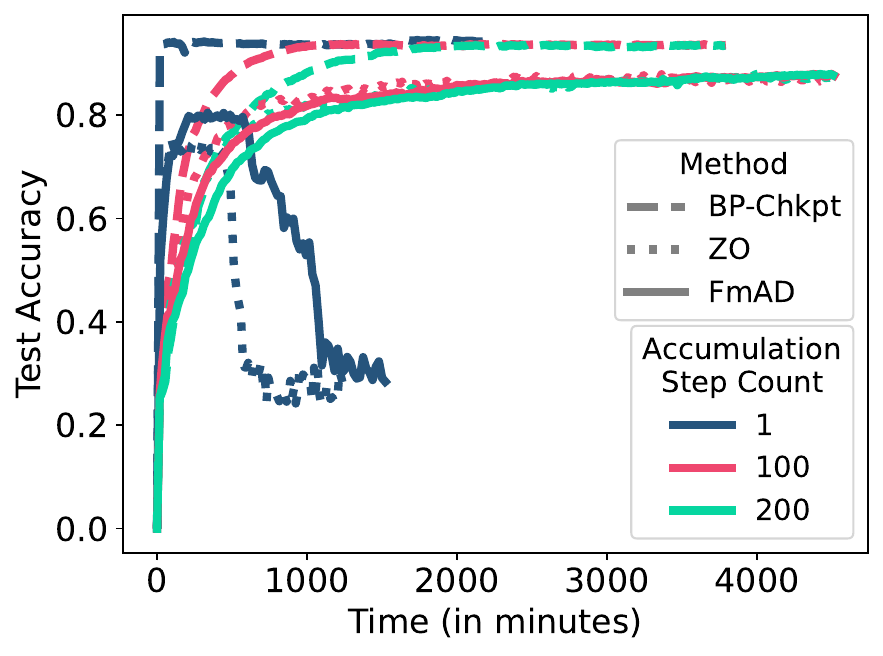}
         \caption{Accumulating and averaging gradients across iterations}
         \label{fig:gradient-accumulation}
     \end{subfigure}
        \caption{
        Experiments (on AGNews with \textsc{Llama} 3.1 8B) using varying perturbation counts (\textsc{-Multiple}) and accumulation steps (\textsc{-Accumulate}) show both strategies reduce gradient noise and improve convergence stability for \fmad and ZO.
        However, \textsc{-Multiple} increases memory and compute costs, while \textsc{-Accumulate} slows convergence.
        As shown in (\emph{right}), \textsc{BP-Checkpointing} achieves the highest, most stable accuracy fastest. \fmad performs moderately but is unstable or slow, and ZO (step size 1) collapses early and fails to match BP's accuracy.
        }
        \label{fig:variance-reduction-methods}
\end{figure}

\textbf{Variance reduction approaches improve the accuracy of \fmad and ZO methods yet fall short of closing the gap with BP methods.} 
Both \fmad and ZO benefit from their \textsc{-Accumulate} and \textsc{-Multiple} variants, which reduce gradient noise by trading off higher compute or memory. 
\textsc{FmAD-Accumulate} improves over \textsc{FmAD-Vanilla} by 1.4--9.8\% across datasets, except on GSM8K (-6.9\%), likely due to its need for smaller batch sizes. 
Similarly, \textsc{ZO-Accumulate} boosts accuracy by 0.2--12.2\%, with an 8.3\% drop on GSM8K. 
\textsc{FmAD-Multiple} improves by 0.6--5.7\%, and \textsc{ZO-Multiple} by 0.2--13.2\%, with only a 0.5\% drop on GSM8K.

To understand the effects of these two variance reduction techniques, we vary the number of perturbations and accumulation steps.
Figure~\ref{fig:multiple-perturbations} shows that increasing perturbation count ($n=10, 50$) yields 5.7--7.7\% (\fmad) and 13.2--14.0\% (ZO) accuracy gains on AGNews, consistent with the observations of \S~\ref{sec:theoretical-analysis}.
Similarly, Figure~\ref{fig:gradient-accumulation} shows that increasing accumulation steps (100, 200) yields 7.5--7.6\% (\fmad) and 12.2--14.0\% (ZO) gains.
These improvements come at the cost of increased convergence time (sequential implementation of \textsc{-Multiple}), memory (parallel implementation of \textsc{-Multiple}), or slower updates (\textsc{-Accumulate}). These trade-offs are discussed in \S~\ref{subsec:wallclock-runtime} and~\S~\ref{subsec:memory-consumption}.

\textbf{Other variance reduction approaches offer limited or inconsistent accuracy improvements for \fmad and ZO.}
\textsc{-Adaptive} often underperforms, with \textsc{FmAD-Adaptive} trailing \textsc{-Vanilla} on BoolQ (-4.3\%) and MultiRC (-3.2\%), likely due to biased updates from gradient-informed perturbation sampling.
\textsc{-Sparse} performs worst overall, lagging \textsc{-Vanilla} by 1.2--10.1\% (\fmad) and 1.9--9.0\% (ZO), as random perturbations of early steps mislead saliency-based parameter selection.
\textsc{-SVRG} improves classification accuracy by 4.0--11.1\%, but failing on GSM8K (-4.2\%) due to homogenized updates that weaken variance correction (see Appendix~\ref{adx:challenges-with-svrg}).
\textsc{MeZO} offers modest gains (1.0--6.9\%) on classification but lacks applicability to generative and vision-language tasks.

\textbf{Accuracy gaps widen as trainable parameters or model size increases.} To further evaluate the impact of trainable parameter count on \fmad and ZO, we conducted additional experiments on medium-sized models (110M–350M parametered BERT and \textsc{RoBERTa}) and large-sized models (OPT 1.3B, 6.7B, with various \textsc{LoRA} ranks, and 13B), 
as detailed in Appendices~\ref{adx:experiments-with-medium-sized-models} and~\ref{subsec:reducing_trainable_parameter_count}. Both \fmad and ZO still exhibit slower convergence and degraded performance compared to BP, with the gap widening as model size increases (especially in the case of BERT and \textsc{RoBERTa}).
Experiments on changing the perturbation variance are presented in Appendix~\ref{adx:changing-variance-of-random-pert-sampling}.

\subsection{Comparison on Wallclock Convergence Time}
\label{subsec:wallclock-runtime}

Convergence time determines how quickly a trained model becomes feasible for practical use.
We contextualize our analysis using Figure~\ref{fig:variance-reduction-methods}, which illustrates the time-to-accuracy curve of the \textsc{-Vanilla} methods.
Figure~\ref{fig:loss-curves} in Appendix~\ref{adx:experimental_variance} includes results on the remaining datasets.

\textbf{Compared to \textsc{ZO} and \textsc{FmAD}, \textsc{BP-Checkpointing} achieves the fastest convergence speed and highest accuracy.} 
Figure~\ref{fig:gradient-accumulation} (AGNews, batch size 40) compares test accuracy against wall-clock time. Since \textsc{BP-Vanilla} runs out of memory at this batch size, we instead report a smaller batch size (8) for a fair runtime comparison between \textsc{BP-Vanilla} and \textsc{BP-Checkpointing}. At batch size 8, \textsc{BP-Vanilla} requires 804.4s/iter, while \textsc{BP-Checkpointing} takes 936.3s/iter ($\sim$1.2$\times$ slower per iteration). Despite this overhead, \textsc{BP-Checkpointing} still outperforms \fmad by $\sim$1.2$\times$ per iteration and achieves 4.5--27.6\% higher accuracy.
\begin{figure}%{r}{0.5\textwidth}
    \centering
    \includegraphics[width=\linewidth]{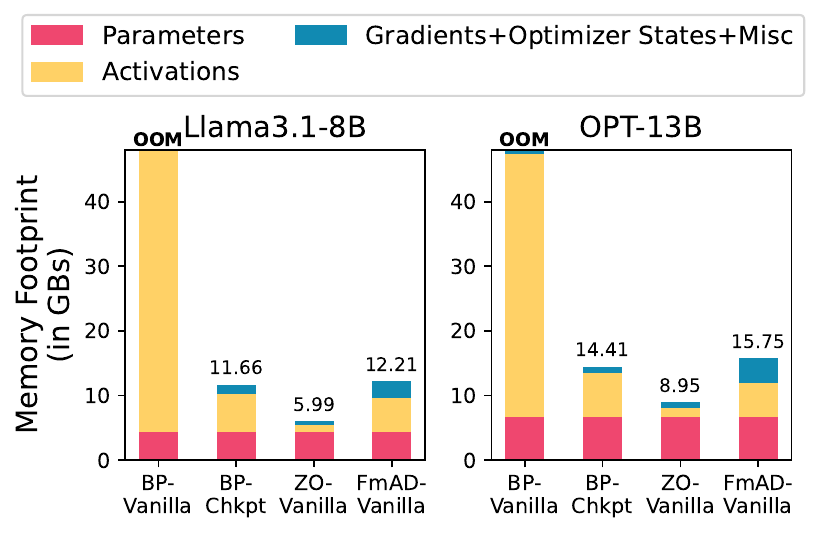}
    \caption{
    Breakdown of memory consumption of training (\emph{left}) \textsc{Llama 3.1} (8B) and (\emph{right}) \textsc{OPT} (13B) models on AGNews dataset on one L40s GPU.
    Although \textsc{BP-Checkpointing} is 1.6--1.9$\times$ takes more memory than ZO, it takes far fewer iterations to achieve 4.5--31.1\% higher accuracy (as shown in Figure~\ref{fig:gradient-accumulation}). 
    }
    \label{fig:memory-cost-llama-opt}
\end{figure}
At batch size 40, where memory is the limiting factor, \textsc{BP-Checkpointing} converges reliably with 1112.8s/iter. In comparison, \fmad requires 1286.5s/iter, and ZO is the fastest at 726.7s/iter ($\sim$1.5$\times$ faster than \textsc{BP-Checkpointing}). However, this runtime advantage does not translate to accuracy: \textsc{BP-Checkpointing} reaches $\sim$94\% accuracy, while \fmad and ZO fall short due to slower convergence and instability. Specifically, ZO suffers from approximation errors in gradient estimation, leading to accuracy degradation of 5.6--31.1\% relative to BP-based methods.

In terms of overall time-to-accuracy, \textsc{BP-Checkpointing} achieves convergence 21.4--34.8\% faster than alternatives. 
The gap with \fmad arises from its computational inefficiency: unlike BP, which reuses downstream gradients with a single matrix multiplication per layer, \fmad requires two matrix multiplications per layer for \jvp evaluation (Eq.~\ref{eq:fmad-one-layer-computations}).

\textbf{Variance reduction improves convergence, but often slows down convergence time.}
As shown in Figure~\ref{fig:multiple-perturbations}, \textsc{-Multiple} variants (e.g., with $n = 10$, $50$) yield smoother training and higher accuracy than their $n = 1$ counterparts.
However, these gains come with a proportional increase in convergence time for sequential implementations as the runtime scales linearly with the number of perturbations. 
\textsc{-Accumulate} variants (Figure~\ref{fig:gradient-accumulation}) improve accuracy without increasing per-iteration cost, as they amortize single-step estimates over multiple updates. However, the delay in parameter updates slows down overall convergence: with an accumulation window of 200, training is 14.8$\times$ and 10.9$\times$ slower for \fmad and ZO, respectively, than when trained without accumulation.

\subsection{Comparison on Memory Consumption}
\label{subsec:memory-consumption}

\textbf{Memory savings from \fmad and ZO come at the cost of accuracy and convergence speed.}
Figure~\ref{fig:memory-cost-llama-opt} shows that both \fmad and ZO reduce memory usage relative to \textsc{BP-Vanilla}, which runs out of memory (OOM) due to storing all activations. 
By contrast, \fmad and ZO store only the previous layer's activation, yielding a lower memory footprint. 
However, as seen in Figure~\ref{fig:gradient-accumulation}, these savings lead to significantly longer training times and degraded model performance.
Meanwhile, \textsc{BP-Checkpointing} uses 0.6--1.3GB less memory than \fmad, while delivering substantially faster convergence and 4.5–31.1\% higher accuracy.
Further, \fmad consumes 3.3–4.3$\times$ more activation memory than ZO. 
This overhead stems from the need to simultaneously store previous layer's intermediate activations for both the primary forward pass and the additional \jvp computation.

\textbf{Variance reduction strategies introduce memory-accuracy trade-offs.}
The \textsc{-Multiple} variants improve gradient quality by evaluating multiple perturbations per step, but parallel implementations require linearly more memory. 
For instance, if one forward pass needs 1.26GB (ZO) or 5.41GB (\fmad) for activations, using $n$ perturbations inflates this to 1.26$n$GB or 5.41$n$GB, respectively. 
On the other hand, \textsc{-Accumulate} amortizes these computations over time and introduces no memory overhead, though at the cost of slower convergence.

Appendix~\ref{adx:more-dataset-model-combinations} provides additional dataset–model results and the corresponding runtime and compute-to-convergence comparisons.

\subsection{Comparison on Computation Cost}
\label{subsec:computation-cost}

Computation cost in terms of FLOPS directly impacts energy consumption and determines whether training large models is feasible under given resources. 
Table~\ref{tbl:computation_cost} reports both per-iteration cost and total cost until convergence. (Table~\ref{tbl:theoretical-bounds} summarized the theoretical bounds.)

\textbf{\fmad and ZO methods  reduce per-iteration compute costs but incur significantly higher total compute due to slow convergence.}
\textsc{ZO-Vanilla} incurs a relatively low per-iteration cost of 288.7 TFLOPs, approximately 0.7$\times$ the cost of \textsc{BP-Checkpointing}, because it only requires two forward passes per gradient estimate. 
However, this advantage is misleading: 
due to slow convergence, its total computation cost \emph{until convergence} is 3.8$\times$ higher than that of \textsc{BP-Checkpointing}.
\textsc{FmAD-Vanilla} shows a per-iteration cost nearly identical to \textsc{BP-Checkpointing}, but its convergence is hindered by  gradient estimates with high variance, leading to 3.2$\times$ higher total compute costs. 

 \begin{table}%{r}{0.6\textwidth}
\centering
\footnotesize
\tabcolsep=0.03cm
\caption{
Computation cost per iteration and to convergence (lower is better) for \textsc{Llama 3.1} (8B) on AGNews.
\textsc{BP-Checkpointing} remains by far the most compute-efficient, whereas perturbation-based methods (ZO, \fmad), even their \textsc{-Accumulate} variants, incur an order-of-magnitude more TFLOPs to reach convergence.
}
\begin{tabular}{lccc}
\toprule
Method & \begin{tabular}[c]{@{}c@{}} TFLOPs \\  per Iter. ($\downarrow$)\end{tabular} & 
\begin{tabular}[c]{@{}c@{}}TFLOPs until \\  Convergence ($\downarrow$)\end{tabular}
& \begin{tabular}[c]{@{}c@{}}\# Iter. until\\ Convergence \end{tabular} \\
\midrule
\textsc{BP-Checkpointing} & \phantom{0}434.4 & \phantom{00}65.2 $\times 10^4$ & \phantom{0}1.5$\times 10^3$ \\ 
\midrule
\textsc{ZO-Vanilla}       & \phantom{0}288.7 & \phantom{0}251.2 $\times 10^4$ & \phantom{0}8.7 $\times 10^3$ \\
\textsc{ZO-Multiple}      &           2886.8 &           2425.0 $\times 10^4$ & \phantom{0}8.4 $\times 10^3$ \\
\textsc{ZO-Accumulate}    & \phantom{0}288.7 &           2165.1 $\times 10^4$ &           75.0 $\times 10^3$ \\ 
\midrule
\textsc{FmAD-Vanilla}     & \phantom{0}432.0 & \phantom{0}207.4 $\times 10^4$ & \phantom{0}4.8 $\times 10^3$ \\
\textsc{FmAD-Multiple}    &           4320.3 &           4147.5 $\times 10^4$ & \phantom{0}9.6 $\times 10^3$\\
\bottomrule
\end{tabular}
\label{tbl:computation_cost}
% \vspace{-0.5cm}
\end{table}

\textbf{Multiple perturbations per iteration improves accuracy but linearly increases cost.}
In \textsc{ZO-Multiple}, using 10 perturbations per iteration leads to a 9.7$\times$ increase in compute, showcasing the linear relationship between the number of perturbations and cost. 
In contrast, \textsc{ZO-Accumulate}, which accumulates gradients across iterations without increasing perturbation count, maintains similar cost to \textsc{ZO-Vanilla} but still suffers from slow convergence.
Similarly, for \fmad, when we increase the number of perturbations by 10$\times$ to reduce gradient variance and improve accuracy, the cost increases by 20$\times$ that of BP, as each \jvp involves two matrix multiplications.

\subsection{Algorithmic vs. Implementation Effects}
\label{subsec:algo-vs-impl-effects}

\begin{table}[]
\caption{Controlled comparison of efficacy, memory, and time-to-convergence for BP, BP-Checkpointing, and \fmad on a small MLP under identical optimization settings. 
Even after isolating implementation overhead, \fmad is intrinsically less efficient, with higher MSE and slower convergence than backprop variants.}
\scriptsize
\setlength{\tabcolsep}{2.3pt}
\label{tab:algo-vs-impl-effects}
\begin{tabular}{lccccc}
\toprule
Method & \begin{tabular}[c]{@{}c@{}}Train\\ MSE (↓)\end{tabular} & \begin{tabular}[c]{@{}c@{}}Val\\ MSE (↓)\end{tabular} & \begin{tabular}[c]{@{}c@{}}Iter Time\\ (ms, ↓) \end{tabular} & \begin{tabular}[c]{@{}c@{}}Memory\\ (MB) (↓)\end{tabular} & \begin{tabular}[c]{@{}c@{}}Time to\\Converge (s, ↓) \end{tabular} \\
\midrule
\textsc{BP (PyTorch)} & \phantom{0}0.11 & \phantom{0}0.63 & 2.46 & 243 & \phantom{0}8.70 \\
\begin{tabular}[c]{@{}l@{}}\textsc{BP-Checkpointing}\\\textsc{\phantom{0}(PyTorch)}\end{tabular} & \phantom{0}0.11 & \phantom{0}0.63 & 9.80 & 184 & 35.24 \\
\textsc{FmAD (PyTorch)} & 16.98 & 60.17 & 5.54 & 203 & 98.22 \\
\textsc{FmAD (Custom)} & 16.98 & 60.17 & 3.99 & 195 & 78.54 \\
\bottomrule
\end{tabular}
\end{table}
To separate algorithmic limitations from framework-dependent overhead, we reimplemented \fmad with a fully custom \texttt{jvp} propagation, reducing its per-iteration runtime from 5.54 ms to 3.99 ms.
The full comparison is summarized in Table~\ref{tab:algo-vs-impl-effects}.
Despite this improvement, \fmad remains substantially less efficient than backpropagation, exhibiting 2.2–9$\times$ slower convergence and significantly higher error (95$\times$ higher validation MSE at convergence).

We evaluate all methods on a controlled small MLP setup (25,348 parameters; \textsc{AdamW} with lr=1e-3; synthetic linear regression; train/val split 4096/512; MSE metric), where the custom \fmad variant replaces \texttt{torch.autograd.forward\_ad} with explicit layer-wise \texttt{jvp} propagation. 
This isolates algorithmic behavior from PyTorch’s automatic differentiation overhead.

\textbf{\fmad's inefficiency is primarily algorithmic rather than implementation-driven.}
Across methods, BP achieves the lowest error and fastest convergence, while \textsc{BP-Checkpointing} introduces higher per-step cost but still converges more efficiently overall. 
In contrast, both \fmad variants converge to substantially worse solutions, with 160$\times$ higher training MSE and 95$\times$ higher validation MSE after 200 steps. 
At convergence, BP (with or without checkpointing) reaches 0.0424 validation MSE, whereas \fmad remains at 1.75, corresponding to a 9$\times$ longer runtime (78.5s vs. 8.7s for BP). 
Even \textsc{BP-Checkpointing}, despite its higher per-iteration cost, remains 2.2–3$\times$ faster due to requiring fewer optimization steps.

Overall, these results highlight that \textsc{BP-Checkpointing} offers a more favorable compute–memory–accuracy trade-off under identical optimization settings.

\subsection{{Failure Mode Analysis}}
\label{subsec:failure-mode-analysis}
Here, we analyze why variance reduction methods and adaptive optimizers sometimes fail to make \fmad and ZO converge reliably. 

\textbf{Cascading \jvp Amplification with Adaptive Optimizers.}
A key failure mode in \fmad arises with adaptive optimizers, such as \textsc{AdamW}, triggering cascading amplification of Jacobian-vector products (\jvp).  
On GSM8K, \jvp magnitudes remain stable under \textsc{SGD} within $[-50,50]$, but spike $8\text{--}10\times$ under \textsc{AdamW} (Figure~\ref{fig:unstable-jvp-optimizer}).  
These spikes produce large gradient updates, inflating weights and further amplifying \jvp values, a positive feedback loop that can cause divergence or noisy updates.  

\textbf{Gradient Variance and Magnitude Explains Performance Drops.} 
Effective gradient variance under \textsc{AdamW} is 4--6$\times$ higher than SGD, with peaks of 200--400 in hidden layers of the LLaMA-7B subset.  
This instability correlates with 2--5\% lower final accuracy vs.\ BP with checkpointing, and some runs yield \texttt{NaN} gradients.  
Spikes typically appear after 50--100 iterations, indicating accumulation from the rolling-average mechanism in adaptive optimizers.  
This behavior is consistent with our theoretical analysis: both ZO and \fmad exhibit variance that scales with $\mathcal{O}((d+1)/n)$, making the optimization particularly sensitive in high-dimensional settings with few perturbation samples. 
Under adaptive optimizers, these noisy gradient estimates interact with momentum and variance normalization terms, causing rare high-magnitude \jvp realizations to persist across multiple updates rather than dissipate after a single step. 
As training progresses, this accumulation increases effective gradient magnitudes and destabilizes the optimization trajectory.

\textbf{Non-Adaptive SGD Maintains Stability.} 
In contrast, \textsc{SGD} keeps \jvp bounded and gradients closely track backpropagation, producing stable convergence (Figures~\ref{fig:stable-gradients},~\ref{fig:unstable-gradients}).  
These results highlight a critical interaction between optimizer choice and \fmad stability: adaptive optimizers can introduce harmful gradient artifacts in \fmad and ZO methods.  
Further details, including additional datasets, layer-wise analyses, and variance-reduction strategies, are provided in Appendix~\ref{adx:failure-mode-analysis}.

\textbf{\texttt{jvp} Amplification in \textsc{Adam} Creates a Compounding Instability Loop.}
Corollary~\ref{cor:jvp_amplification_adam} shows a feedback loop between \texttt{jvp}-induced variability and \textsc{Adam}’s adaptive normalization. 
As \texttt{jvp} magnitudes increase, the effective update scaling $\psi_t$ grows, which amplifies gradient noise rather than averaging it out. 
This amplification degrades the stability term in the convergence bound, tightening the step-size condition. 
Consequently, larger \texttt{jvp} values lead to larger parameter updates, which further increase \texttt{jvp} ranges through growing weight norms. 
The effect becomes particularly severe in high-dimensional settings, where the variance scales with $\mathcal{O}((d+1)/n)$ and optimization becomes increasingly sensitive to perturbation noise. 
Unlike \textsc{SGD}, \textsc{Adam} maintains rolling first- and second-moment statistics, allowing rare high-magnitude \texttt{jvp} realizations to persist across many iterations. 
Our empirical results on GSM8K reflect this behavior, with \texttt{jvp} spikes often preceding divergence or \texttt{NaN} updates. 
Overall, this creates a compounding instability effect that worsens convergence guarantees in practice.

\section{Conclusion}
\label{sec:conclusion}

While forward-mode AD (\fmad) and zero-order (ZO) optimization have been proposed as memory-efficient alternatives to backpropagation (BP), prior work lacked comparison with checkpointed BP and unified theoretical bounds for LLM training.
Our analysis closes these gaps, revealing that \fmad and ZO incur higher computational cost, slower convergence, and greater sensitivity to model dimensionality and perturbation budgets.
Even with enhancements like variance reduction, they remain less efficient and robust than BP with activation checkpointing.
Empirical results on large LLMs and vision-language models confirm that checkpointed BP consistently outperforms \fmad and ZO across accuracy, convergence speed, and compute cost, at comparable memory usage.
These findings reaffirm checkpointed BP as the most practical strategy for memory-constrained LLM training and clarify the limitations of \fmad and ZO.

\clearpage

\section*{Acknowledgements}
\label{sec:acknowledgements }

This material is based upon work supported by the National Science Foundation under Grant No. CCF-2210243, CNS-2312396, CNS-2338512, IIS-2435822, and CCF-2449995.
Any opinions, findings, and conclusions or recommendations expressed in this material are those of the author(s) and do not necessarily reflect the views of the National Science Foundation.

\section*{Impact Statement}
\label{sec:impact-statement}
Training deep learning models already carries a high environmental cost due to significant energy consumption.
Our study shows that forward-mode AD and zero-order optimization, despite saving memory in some cases, require much longer training times and compute compared to backpropagation with checkpointing. 
This inefficiency leads to greater carbon emissions overall.
Therefore, we show that optimizing for true computational efficiency (time-to-convergence and compute; along with memory consumption) is crucial for reducing the environmental footprint of large-scale training.

We also acknowledge that misinterpreting our results could lead to the premature dismissal of forward-mode AD or zero-order methods altogether. 
While they are not scalable replacements for backpropagation in large-scale training, they may still be uniquely suited for small models, non-differentiable tasks, or privacy-preserving settings where explicit gradients are inaccessible. 
Careful contextual understanding is necessary when applying our conclusions.

\bibliography{bibliography01, bibliography02}
\bibliographystyle{icml2026}

%%%%%%%%%%%%%%%%%%%%%%%%%%%%%%%%%%%%%%%%%%%%%%%%%%%%%%%%%%%%%%%%%%%%%%%%%%%%%%%
%%%%%%%%%%%%%%%%%%%%%%%%%%%%%%%%%%%%%%%%%%%%%%%%%%%%%%%%%%%%%%%%%%%%%%%%%%%%%%%
% APPENDIX
%%%%%%%%%%%%%%%%%%%%%%%%%%%%%%%%%%%%%%%%%%%%%%%%%%%%%%%%%%%%%%%%%%%%%%%%%%%%%%%
%%%%%%%%%%%%%%%%%%%%%%%%%%%%%%%%%%%%%%%%%%%%%%%%%%%%%%%%%%%%%%%%%%%%%%%%%%%%%%%
\newpage
\appendix
\onecolumn

\section{Related Work}
\label{sec:related-work}
Here we review recent works on forward-mode AD and zero-order optimization methods; along with a discussion on various methods to refine the memory and time efficiency of backpropagation. 

\subsection{Forward-mode AD}
The application of forward-mode automatic differentiation (\fmad) for training deep neural networks was first introduced in Forward-Gradient Descent (\textsc{Fgd})~\citep{baydin2022gradients}, building on an earlier survey on automatic differentiation~\citep{baydin2017autodiff}. 
\textsc{Fgd} demonstrated \fmad on a small-scale three-layer fully connected model and a four-layer convolutional network, claiming that \fmad can outperform backpropagation (BP) in speed and loss reduction per unit time. 
However, these claims remain unverifiable, as the implementation was never made publicly available, and subsequent independent evaluations~\citep{orobix2023fgd} have found these results difficult to reproduce.

Beyond this initial demonstration, more recent efforts have attempted to improve \fmad's efficiency.
\emph{Can Forward Gradients Match Backpropagation?}~\citep{fournier2023canforwardgrad} seeks to enhance \fmad by generating more structured perturbations rather than relying on random sampling. 
This approach introduces local losses computed via small auxiliary networks to inform perturbation choices. 
However, training these auxiliary networks significantly increases memory consumption and computational overhead, undermining \fmad's intended efficiency advantage.

Other studies have focused on extending \fmad beyond first-order gradients. 
\emph{Second-order \fmad}~\citep{cobb2024secondorderforwardmodeautomaticdifferentiation} provides a formal framework for computing second-order gradients with \fmad, demonstrating improved optimization performance. 
However, this comes at a substantial computational cost, and experiments remain limited to small-scale benchmarks (e.g., a CNN with only 431K parameters), leaving open the question of whether second-order \fmad can scale competitively against BP. 
Similarly, Taylor-mode Auto Differentiation~\citep{bettencourt2019taylormode} generalizes \fmad to compute higher-order gradients, yet the memory and time-to-convergence trade-offs compared to BP remain unexplored.

Several other works have proposed variations of \fmad without fundamentally addressing its inefficiencies. 
\emph{Randomized Forward Gradient-based GD}~\citep{shukla2023randomized} provides a convergence analysis of \fmad using random perturbations but offers no new insights into its computational efficiency. 
\textsc{Projected-FG}~\citep{rostami2024projectedforwardgradientguidedfrankwolfe} applies \fmad to memory-efficient Frank-Wolfe optimization but evaluates only small models, making its conclusions inapplicable to large-scale deep learning. 
\emph{Beyond Backpropagation}~\citep{flugel2024beyond} investigates the use of multiple perturbations per iteration to improve forward-gradient computation but fails to identify why \fmad remains inferior to BP in practice.

A more recent large-scale application of \fmad appears in \textsc{Spry}~\citep{panchal2024thinking}, which employs \fmad for fine-tuning large models (ranging from 100K to 13B parameters) in a federated learning setting. 
By restricting each client to a small subset of weights, \textsc{Spry} circumvents \fmad's poor performance in high-dimensional perturbations. 
However, even in this setting, \fmad exhibits slower convergence and higher variance than BP, further reinforcing its fundamental limitations.

For applications besides LLM training or finetuning, biological plausibility~\citep{schmidthieber2023interpretinglearningbiologicalneural, xiao2024onlinepseudozerothordertrainingneuromorphic} has been proposed as a motivating factor for exploring alternative gradient estimation techniques. 
\fmad avoids the backward signal transport required by backpropagation and has therefore been considered more biologically plausible. 
Though \fmad still relies on first-order derivatives and engineered automatic differentiation, which limits its direct applicability to biological systems.

While prior works have demonstrated narrow successes of \fmad in specialized scenarios, none have systematically analyzed its fundamental computational constraints.
Besides, the comparison of \fmad against a strong baseline of \textsc{BP-Checkpointing} remains uncharted.
Unlike these related studies, our work provides a principled theoretical and empirical investigation into the scalability bottlenecks of \fmad, explicitly comparing its memory and time complexity against \textsc{BP-Checkpointing}. 
We also uncover failure modes of \fmad in deep networks, offering new insights into why it cannot consistently surpass BP in terms of both time-to-convergence and efficiency.

\subsection{Zero-order Optimization}
Zero-order (ZO) optimization has received significant attention, particularly in settings where first-order gradient information is unavailable or impractical to compute. 
Unlike \fmad, which has seen limited large-scale adoption, ZO methods have been actively explored in deep learning due to their applicability in scenarios such as adversarial attacks, black-box optimization, and gradient-free fine-tuning.
Similar to \fmad, we also note that none of the works discussed below have made a comparison of their ZO-based variant against \textsc{BP-Checkpointing}, an aspect which is fleshed out in this work.

\textsc{MeZO}~\citep{malladi2023mezo} and its extension \textsc{MeZO-SVRG}~\citep{gautam2024mezosvrg} introduced memory-efficient ZO optimization strategies that regenerate random perturbations instead of storing them, effectively reducing memory overhead. 
These methods have demonstrated practical advantages in fine-tuning large language models (LLMs) for classification tasks without requiring explicit backpropagation. 
While they address memory constraints, they do not provide insights into the fundamental efficiency trade-offs between ZO and BP in terms of time-to-convergence, memory consumption, and attained accuracy; which are the key concerns of our work.
{A closely related line of work is \textsc{HiZOO}~\citep{zhao2025hizoo}, which proposes a forward-only second-order ZO optimizer that uses Hessian-informed perturbations to accelerate MeZO-style fine-tuning. 
While \textsc{HiZOO} successfully demonstrates reduced activation-memory usage relative to \textsc{MeZO}, its evaluation focuses primarily on memory rather than wall-clock convergence time or total compute cost; metrics that are central to our analysis.
Moreover, the algorithm introduces additional second-order computations (via Hessian-related estimators), whose overhead is not thoroughly quantified.}

Expanding on these efforts, \textsc{DeepZero}~\citep{chen2024deepzero} proposed a ZO deep learning framework capable of training deep neural networks from scratch. 
By leveraging coordinate-wise gradient estimation (CGE) over randomized vector-wise estimation, \textsc{DeepZero} achieves improved accuracy and computational efficiency.
Additionally, the introduction of sparsity-induced training, feature reuse, and forward parallelization brings ZO training closer to first-order methods, achieving state-of-the-art results on ResNet-20 trained on CIFAR-10. 
However, despite these advancements, ZO remains fundamentally limited by high variance and inefficient gradient estimation, resulting in slower convergence compared to BP, which is an issue we empirically validate in our benchmarks.

Other works, such as \textsc{DZOVR}~\citep{chen2023zeroth} and \textsc{ZO-SVRG}~\citep{liu2018zosvrg}, have attempted to improve ZO efficiency by incorporating Stochastic Variance Reduced Gradients (SVRG)~\citep{johnson2013svrg}.
Similarly, research on ZO methods for non-convex and non-smooth optimization~\citep{liu2024zerothorder, kornowski2024algorithm, balasubramanian2018zerothorder} has provided valuable theoretical insights. 
However, none of these studies systematically compare ZO to BP in terms of memory consumption, execution time, and scalability, leaving open the question of whether ZO can ever be a viable alternative. 
Our work explicitly addresses this gap by benchmarking these methods against BP and highlighting their structural inefficiencies.

Further, \textsc{ZO-AdaMM}~\citep{chen2019zo} integrates an adaptive optimizer (\textsc{AdaMM}) into ZO, demonstrating improved stability.
However, even with adaptive optimization, ZO struggles to match the convergence speed of BP, as shown in their experiments on a small-scale CNN. 
Additionally, work on ZO optimization in high-dimensional settings~\citep{wang2018stoachasticzerothorder} has focused primarily on convergence properties rather than the computational and memory efficiency bottlenecks that limit ZO's practical scalability.

\emph{Revisiting ZO}~\citep{zhang2024revisitingzo} benchmarks the performance of large language models trained using BP, \fmad, and ZO optimization. 
Our work differs in several key ways:
(a)~We include comparisons against a backpropagation with checkpointing baseline, offering new insights into the memory-efficiency trade-offs among gradient computation methods.
(b)~Unlike \emph{Revisiting ZO}, our study evaluates both time-to-convergence and overall computational cost, which are critical for understanding practical scalability.
(c)~We also provide an in-depth failure mode analysis, focusing on the behavior of Jacobian-vector products and their influence on model updates, an aspect unexplored in their work.

{We note that like ZO with LLMs, ZO for biological systems~\citep{schmidthieber2023interpretinglearningbiologicalneural} would face scalability and convergence challenges when applied to high-dimensional models. 
Our study does not aim to contest the conceptual motivations behind these techniques; rather, we show their practical limitations, in terms of computational cost and optimization performance, for large-scale models like LLMs.}

\subsection{Optimizations on Backpropagation}
Backpropagation (BP) remains the dominant method for training deep neural networks due to its computational efficiency and well-optimized implementations.
However, standard BP incurs high memory costs, as it requires storing intermediate activations for the entire computational graph during the forward pass. 
This limitation has motivated extensive research into memory-efficient variants of BP that aim to reduce memory consumption without significantly compromising training speed.

Checkpointing-based methods, such as \textsc{Reversible Residual Networks}~\citep{gomez2017reversibleresidual} and \textsc{Activation Checkpointing}~\citep{chen2016activationcheckpointing}, trade memory for recomputation by strategically discarding and later recomputing activations.
These techniques have proven effective in reducing memory overhead, but they introduce additional computational costs.
More recent approaches, such as \textsc{Efficient Rematerialization}~\citep{gruslys2016memory} and \textsc{Dynamic Programming-based Activation Offloading}~\citep{beaumont2021rematerializationandoffloading}, attempt to optimize checkpointing strategies to minimize recomputation overhead. 
Despite these advances, BP with checkpointing still follows the same fundamental backpropagation framework and benefits from computation reuse -- an efficiency advantage that \fmad and ZO methods lack.

\section{Datasets}
\label{adx:datasets}
In this section, we provide detailed descriptions of the datasets used in our experiments. 
For each dataset, we outline its origin, licensing, the version we have used, and task-specific characteristics, including the number of samples, sequence lengths and relevant domain or classification details. 

\paragraph{AGNews.}
The AG News dataset~\citep{zhang2015agnews} is derived from a corpus of 496,835 labeled news articles collected from over 2,000 web-based news sources published between 2004 and 2005. 
For this work, we use a widely adopted, cleaned, and balanced subset comprising 120,000 training samples and 7,600 test samples, evenly distributed across four categories: 
World, 
Sports, 
Business, 
and Science/Technology. 
We divide the test data into half to create validation and test splits.
The dataset is primarily used for topic classification, which is also the focus of our study. 
The maximum sequence length for our experiments is set to 350 tokens during training.
It is released under the Creative Commons CC0 1.0 Universal license, placing it in the public domain. 
We obtained the dataset via the Hugging Face Datasets library~\citep{hf2022agnews}.

\paragraph{BoolQ.}
The Boolean Questions (BoolQ) dataset~\citep{clark2019boolq} is a reading comprehension benchmark consisting of naturally occurring yes/no questions. 
Each instance includes a question, a passage (typically a paragraph from Wikipedia), and a binary answer (``yes'' or ``no'') derived from the passage content. 
Unlike synthetic question-generation benchmarks, BoolQ features real user queries collected from Google search logs, making the task more reflective of real-world comprehension. 
The dataset contains approximately 9,427 question-passage training pairs, and 3,270 validation pairs.
We divide the validation data into half to create the validation and test data splits for this work.
The maximum sequence length for our experiments is set to 1200 tokens during training.
BoolQ is released under the Creative Commons Share-Alike 3.0, which allows for flexible use, modification, and redistribution with appropriate attribution. 
Once again, Hugging Face Datasets was used to access BoolQ~\citep{hf2022boolq}.

\paragraph{MultiRC.}
The Multi-Sentence Reading Comprehension (MultiRC)~\citep{khashabi2018multirc} dataset is a benchmark corpus designed to evaluate machine reading comprehension over short paragraphs. 
Each example consists of a paragraph followed by one or more questions, with corresponding candidate answers that must be inferred from the text. 
In our setup, we frame the task as a binary classification problem, determining whether a given question-answer pair is correct or incorrect based on the paragraph content.
The dataset contains approximately 6,000 multi-sentence questions drawn from over 800 distinct paragraphs. 
The maximum sequence length for our experiments is set to 1500 tokens during training.
MultiRC is released under the MIT License, permitting broad use and redistribution with attribution.
We accessed the dataset through Hugging Face~\citep{hf2022multirc}.

\paragraph{GSM8K.}
The Grade School Math 8K (GSM8K) dataset~\citep{cobbe2021gsm8k} is a high-quality benchmark for evaluating arithmetic reasoning and problem-solving abilities of language models. 
Each example consists of a single math word problem followed by a detailed, step-by-step answer. 
Designed to emphasize multi-step reasoning, the problems are written in natural language and reflect concepts typically found in grade school (middle school) curricula. 
The dataset contains 7,470 training examples and 1,319 test examples, all manually curated for clarity and correctness.
The maximum sequence length for our experiments is set to 800 tokens during training.
In this work, we use GSM8K as a text-to-text supervised learning task, where the input is the problem statement and the target is the final answer (without the reasoning steps). 
The dataset is publicly available under the MIT License, allowing broad reuse and modification with attribution.
The dataset is available on Hugging Face~\citep{hf2022gsm8k}.

\paragraph{MMLU.}
The Massive Multitask Language Understanding (MMLU)~\citep{hendryckstest2021mmlu} dataset is a comprehensive benchmark designed to assess general knowledge and reasoning ability across a wide range of academic and professional subjects. 
It covers 57 diverse topics, including mathematics, history, law, medicine, and the sciences, with questions derived from standardized exams and expert-written materials. 
Each example is a multiple-choice question with four answer options, requiring both factual knowledge and reasoning skills.
All four answer options are included in the prompt.
The dataset consists of 99.8k training samples, 1.5k validation samples, and 14k test samples.
The maximum sequence length for our experiments is set to 1500 tokens during training.
MMLU is publicly available under the MIT License, allowing free use, modification, and distribution with appropriate credit. 
Its breadth and difficulty make it a challenging benchmark for evaluating finetuned language models.
In line with rest of the datasets, we have used the Hugging Face Datasets version of MMLU~\citep{hf2024mmlu}.

\paragraph{VQAv2.}
The Visual Question Answering v2.0 (VQAv2)~\citep{goyal2019vqav2} dataset is a large-scale benchmark designed to evaluate a model's ability to understand and reason over both visual and textual inputs. 
Each example consists of an image (sourced primarily from the MS COCO dataset~\citep{lin2014mscoco}) paired with a natural language question, and the task is to generate an accurate, typically short (often single-word), answer based on the visual content of the image.

VQAv2 addresses the language bias issues present in its predecessor (VQAv1) by ensuring that each question is associated with multiple images, such that the correct answer varies depending on the visual context. This structure encourages models to genuinely integrate image understanding rather than relying solely on question priors.

The dataset contains 443,757 training questions, 214,354 validation questions, and 447,793 test questions, associated with over 200,000 images. 
Each question has 10 human-provided answers, allowing for nuanced evaluation metrics such as accuracy based on answer consensus~\citep{vqav2}.
The maximum sequence length for our experiments is set to 100 tokens during training.
VQAv2 is distributed under the 2-Clause BSD License, allowing for use and adaptation with attribution.
We access the dataset through the VisualQA website~\citep{vqav2}.

\paragraph{TextVQA.}
The TextVQA (Text-based Visual Question Answering) dataset~\citep{singh2019textvqa} is a vision-language benchmark specifically designed to evaluate a model's ability to read and reason about text within images.
Unlike standard VQA tasks that focus on general object and scene understanding, TextVQA centers on questions where the answer relies on text present in the image itself; such as signs, labels, documents, product packaging, and storefronts.

Each example in the dataset includes an image, a natural language question, and a free-form textual answer. 
To correctly answer a question, models must integrate visual understanding with OCR (Optical Character Recognition) capabilities. 
TextVQA challenges systems to perform multimodal reasoning that spans spatial, linguistic, and visual modalities.

The dataset consists of approximately 28,408 questions associated with 14,987 images, split into: 21,953 training questions; 3,166 validation questions; and 3,289 test questions.
Each question is annotated with 10 answers from human annotators to support consensus-based evaluation metrics.
The maximum sequence length for our experiments is set to 100 tokens during training.
TextVQA is publicly available under the CC BY 4.0 (Creative Commons Attribution 4.0 International License), allowing flexible use, sharing, and adaptation with attribution.
The dataset is available for access on Hugging Face~\citep{hf2022textvqa}.

\section{Baselines and Hyperparameters}
\label{adx:baselines-and-hyperparameters}
\paragraph{\textsc{BP-Vanilla}.}
This baseline~\citep{rumelhart1986learning} uses a standard implementation of the training loop with backpropagation as the gradient computation method, without any modifications or enhancements. 
Due to out-of-memory (OOM) issues encountered with larger batch sizes, most experiments involving \textsc{BP-Vanilla} are conducted using smaller batches. 
Table~\ref{tbl:bp-vanilla-hyperparameters} lists the hyperparameters.

\begin{table}[h]
\footnotesize
\centering
\caption{Hyperparameters related to \textsc{BP-Vanilla}, for all datasets.}
\begin{tabular}{lccccccc}
\toprule
 & AGNews & BoolQ & MultiRC & GSM8K & MMLU & VQAv2 & TextVQA \\
\midrule
Batch Size & 8 & 4 & 8 & 4 & 6 & 6 & 8 \\
Learning Rate & $10^{-3}$ & $10^{-3}$ & $10^{-3}$ & $10^{-5}$ & $10^{-4}$ & $10^{-4}$ & $10^{-4}$ \\
Optimizer & \textsc{AdamW} & \textsc{AdamW} & \textsc{AdamW} & \textsc{AdamW} & \begin{tabular}[c]{@{}l@{}}SGD Nesterov \\ Momentum 0.9\end{tabular} & SGD & \textsc{AdamW} \\
\bottomrule
\end{tabular}
\label{tbl:bp-vanilla-hyperparameters}
\end{table}

\paragraph{\textsc{BP-Checkpointing}.}
\textsc{BP-Checkpointing}~\citep{chen2016activationcheckpointing} is identical to \textsc{BP-Vanilla} with one key difference: 
it employs activation checkpointing (also known as gradient checkpointing) to reduce memory consumption, allowing for larger batch sizes without incurring out-of-memory (OOM) errors.
To ensure a fair comparison, the batch sizes used for \textsc{BP-Checkpointing} match those used for the ZO and \fmad variants.
The hyperparameters are given in Table~\ref{tbl:bp-checkpointing-hyperparameters}.

\begin{table}[h]
\footnotesize
\centering
\caption{Hyperparameters related to \textsc{BP-Checkpointing} and \textsc{BP-Accumulate}, for all datasets.}
\begin{tabular}{lccccccc}
\toprule
 & AGNews & BoolQ & MultiRC & GSM8K & MMLU & VQAv2 & TextVQA \\
\midrule
Batch Size & 40 & 40 & 40 & 6 & 8 & 8 & 8 \\
Learning Rate & $10^{-3}$ & $10^{-3}$ & $10^{-3}$ & $10^{-5}$ & $10^{-4}$ & $10^{-4}$ & $10^{-4}$ \\
Optimizer & \textsc{AdamW} & \textsc{AdamW} & \textsc{AdamW} & \textsc{AdamW} & \begin{tabular}[c]{@{}l@{}}SGD Nesterov \\ Momentum 0.9\end{tabular} & SGD & \textsc{AdamW} \\
\bottomrule
\end{tabular}
\label{tbl:bp-checkpointing-hyperparameters}
\end{table}

\paragraph{\textsc{BP-Accumulate}.}
\textsc{BP-Accumulate} follows the same training procedure as \textsc{BP-Checkpointing}, but incorporates gradient accumulation to simulate larger effective batch sizes without exceeding memory constraints. 
Instead of updating model weights after every mini-batch, gradients are accumulated over multiple smaller batches and the update is performed after a fixed number of steps. 
At the end of the accumulation period, the summed gradients are averaged by dividing them by the number of accumulation steps.
The hyperparameters are same as those of \textsc{BP-Checkpointing} (see Table~\ref{tbl:bp-checkpointing-hyperparameters}), with accumulation step count being 100 as default. 

\paragraph{\textsc{ZO-Vanilla}.}
\textsc{ZO-Vanilla}~\citep{chen2019zo} implements a standard zero-order optimization approach, which estimates gradients using only function evaluations according to Equation~\ref{eq:zo-one-layer-computations}, without requiring access to the model's internal, first-order gradients. 
Specifically, it perturbs the model parameters along randomly sampled directions and uses finite differences to approximate the gradient. 
We have used the memory-efficient perturbation trick of \textsc{MeZO} for all the \textsc{ZO-} variants, which includes storing the random seed and regenerating perturbations for forward pass evaluations, instead of persisting entire perturbations in the memory.
For fair comparison, we use the same batch sizes as in \textsc{BP-Checkpointing} and \fmad baselines.
The hyperparameters are given in Table~\ref{tbl:zo-vanilla-hyperparameters}.

\begin{table}[ht]
\footnotesize
\centering
\caption{Hyperparameters related to \textsc{ZO-Vanilla}, for all datasets.}
\begin{tabular}{lccccccc}
\toprule
 & AGNews & BoolQ & MultiRC & GSM8K & MMLU & VQAv2 & TextVQA \\
 \midrule
Batch Size & 40 & 40 & 40 & 6 & 8 & 8 & 8 \\
Learning Rate & $10^{-4}$ & $10^{-3}$ & $10^{-3}$ & $10^{-5}$ & $10^{-5}$ & $10^{-4}$ & $10^{-4}$ \\
Optimizer & \textsc{AdamW} & \textsc{AdamW} & \textsc{SGD} & \begin{tabular}[c]{@{}l@{}}SGD \\Nesterov \\ Mmtm 0.9\end{tabular} & \begin{tabular}[c]{@{}l@{}}SGD \\Nesterov \\ Mmtm 0.9\end{tabular} & \textsc{AdamW} & \textsc{SGD} \\
\begin{tabular}[c]{@{}l@{}}Perturbation\\ Step Size\end{tabular} & $10^{-3}$ & $10^{-2}$ & $10^{-2}$ & $10^{-3}$ & $10^{-4}$ & $10^{-3}$ & $10^{-3}$ \\
\bottomrule
\end{tabular}
\label{tbl:zo-vanilla-hyperparameters}
\end{table}

\paragraph{\textsc{ZO-Accumulate}.}
\textsc{ZO-Accumulate} extends the \textsc{ZO-Vanilla} baseline by incorporating gradient accumulation to simulate larger effective batch sizes without exceeding memory constraints. 
Instead of estimating and applying a parameter update after each mini-batch, gradient approximations (based on finite differences) are accumulated over multiple steps and averaged before updating the model. 
This approach results in improved stability due to averaging out the noisy gradient estimates. 
The hyperparameters are same as with \textsc{ZO-Vanilla}, given in Table~\ref{tbl:zo-vanilla-hyperparameters}, with default accumulation window of 100. 

\paragraph{\textsc{ZO-Multiple}.}
\textsc{ZO-Multiple} (also shown in~\citep{panchal2024thinking, feng2024baffle, xu2024fwdllm}) builds on the \textsc{ZO-Vanilla} method by using multiple random perturbation directions per iteration, to improve the accuracy of the gradient estimate.
Instead of relying on a single direction, this variant samples several perturbations and averages the resulting finite-difference approximations, leading to a lower-variance and more stable update. 
However, this approach increases the number of function evaluations per step. 
The hyperparameters are same as with \textsc{ZO-Vanilla}, given in Table~\ref{tbl:zo-vanilla-hyperparameters}, with default perturbation count per iteration of 10. 

\paragraph{\textsc{ZO-Adaptive}.}
\textsc{ZO-Adaptive} enhances zero-order optimization by incorporating an adaptive perturbation strategy that aligns gradient estimates more closely with the true gradient direction over time. 
The optimization proceeds in two phases. 
In the \emph{calibration phase} (typically the first iteration), multiple perturbation directions are sampled, and the one with the highest positive projected gradient is selected.
This direction is assumed to have the smallest angle with the true gradient. 
This calibrated perturbation is then used to compute an initial gradient estimate. 
In the \emph{adaptive phase} (subsequent iterations), new perturbations are sampled based on the previously estimated gradient, and a rolling average is maintained between the new perturbation and the historical gradient direction. 
This mechanism biases the search toward more promising directions while still allowing for exploratory variation.
The hyperparameters are same as those of \textsc{ZO-Vanilla}, with the inclusion of sampling 4 perturbations during the calibration phase.

\paragraph{\textsc{ZO-SVRG}.}
\textsc{ZO-SVRG}~\citep{liu2018zosvrg} applies the principles of Stochastic Variance Reduced Gradient (SVRG)~\citep{johnson2013svrg} to the zero-order optimization setting, aiming to improve convergence speed and stability by reducing the variance inherent in gradient estimates. 
The method alternates between two types of updates: 
full gradient estimation at a reference point (called a snapshot) and subsequent inner-loop updates that correct noisy estimates using control variates. 
In the zero-order context, both the snapshot gradient and the inner-loop updates are computed using finite-difference approximations along random perturbations. 
The variance reduction comes from reusing the snapshot gradient to correct each inner-step estimate. 
Besides the hyperparameters shown in Table~\ref{tbl:zo-vanilla-hyperparameters}, we use interval of 5 epochs to compute full gradients.

\paragraph{\textsc{ZO-Sparse}.}
\textsc{ZO-Sparse}~\citep{guo2025zosparse} introduces sparsity into zero-order optimization by restricting gradient estimation and updates to only the top 1\% of model parameters, selected based on their magnitude at each iteration.
Unlike structured approaches such as LoRA, this method dynamically identifies and perturbs the most significant weights, those likely to contribute most to loss reduction. Hence, \textsc{ZO-Sparse} focuses the optimization on a small, adaptive subset of parameters. 
This sparsity constraint reduces the dimensionality of the optimization problem, leading to fewer function evaluations.
The hyperparameters are exactly the same as those of Table~\ref{tbl:zo-vanilla-hyperparameters}.

\paragraph{\textsc{MeZO}.}
\textsc{MeZO}~\citep{malladi2023mezo} builds on \textsc{ZO-Vanilla}, but with a key modification tailored for classification tasks using language models. 
Instead of relying on a separate classifier head, \textsc{MeZO} employs the language modeling (LM) head and masks out logits corresponding to vocabulary tokens that are not class labels. 
This approach is presented in the prompt-based fine-tuning strategy introduced by~\citet{gao2021promptfinetuning}.
\textsc{MeZO} integrates this prompting technique with zero-order optimization, enabling effective gradient-free fine-tuning of large language models, although it is limited to the classification tasks.
We use the same hyperparameters as \textsc{ZO-Vanilla} (see Table~\ref{tbl:zo-vanilla-hyperparameters}).

\paragraph{\textsc{FmAD-Vanilla}.}
\textsc{FmAD-Vanilla} implements the standard forward-mode automatic differentiation~\citep{baydin2017autodiff, baydin2022gradients} approach for computing gradients, more details are in \S~\ref{sec:background}.
In this baseline, we use a straightforward implementation of forward-mode AD without any memory-saving strategies or structural optimizations. 
The hyperparameters used for \textsc{FmAD-Vanilla} are summarized in Table~\ref{tbl:fmad-vanilla-hyperparameter}.
Additionally, the variance of the Gaussian distribution used for perturbation sampling is fixed at 1 across all datasets.

\begin{table}[h]
\footnotesize
\centering
\caption{Hyperparameters related to \textsc{FmAD-Vanilla}, for all datasets.}
\begin{tabular}{lccccccc}
\toprule
 & AGNews & BoolQ & MultiRC & GSM8K & MMLU & VQAv2 & TextVQA \\
 \midrule
Batch Size & 40 & 40 & 40 & 6 & 8 & 8 & 8 \\
Learning Rate & $10^{-3}$ & $10^{-4}$ & $10^{-4}$ & $10^{-5}$ & $10^{-5}$ & $10^{-4}$ & $10^{-4}$ \\
Optimizer & \textsc{AdamW} & SGD & \textsc{AdamW} & \begin{tabular}[c]{@{}l@{}}SGD \\ Nesterov \\ Mmtm 0.9\end{tabular} & \begin{tabular}[c]{@{}l@{}}SGD \\ Nesterov \\ Mmtm 0.9\end{tabular} & SGD & \textsc{SGD} \\
\bottomrule
\end{tabular}
\label{tbl:fmad-vanilla-hyperparameter}
\end{table}

\paragraph{\textsc{FmAD-Accumulate}.}
\textsc{FmAD-Accumulate} extends the standard forward-mode automatic differentiation by incorporating gradient accumulation to simulate larger batch sizes without increasing memory consumption. 
The same accumulation strategy is used in corresponding \textsc{BP-Accumulate} and ZO baselines to maintain fairness in comparison.
The hyperparameters are given in Table~\ref{tbl:fmad-vanilla-hyperparameter}, with the addition of accumulation window of 100.

\paragraph{\textsc{FmAD-Multiple}.}
\textsc{FmAD-Multiple} enhances the basic forward-mode AD approach by using multiple perturbation directions per update to improve the stability and accuracy of gradient estimates. 
The setup closely mirrors that of \textsc{ZO-Multiple}, with hyperparameters listed in Table~\ref{tbl:fmad-vanilla-hyperparameter}. 
The only addition is the use of 10 perturbation count per iteration.

\paragraph{\textsc{FmAD-Adaptive}.}
\textsc{FmAD-Adaptive} mirrors the two-phase procedure described in \textsc{ZO-Adaptive}, including the calibration phase for selecting an initial perturbation direction and the adaptive phase that updates this direction using a rolling average of past gradients. 
For full details, we refer the reader to the \textsc{ZO-Adaptive} description. 
All hyperparameters remain consistent with Table~\ref{tbl:fmad-vanilla-hyperparameter}, with calibration phase including 4 perturbations just like \textsc{ZO-Adaptive}.

\paragraph{\textsc{FmAD-SVRG}.}
\textsc{FmAD-SVRG} adopts the same stochastic variance-reduced gradient (SVRG) framework used in \textsc{ZO-SVRG}, but applies it within the forward-mode AD setting. 
It alternates between full-gradient computation on a reference batch and variance-reduced updates on mini-batches, thereby reducing the noise in gradient estimates while maintaining computational efficiency. 
For details on the SVRG formulation, we refer the reader to the description of \textsc{ZO-SVRG}. 
Hyperparameters are in Table~\ref{tbl:fmad-vanilla-hyperparameter}, with full gradients getting computed every 5 epochs (similar to \textsc{ZO-SVRG}).

\paragraph{\textsc{FmAD-Sparse}.}
\textsc{FmAD-Sparse} adopts the same sparsity strategy described in \textsc{ZO-Sparse}, where only the top 1\% of parameters (by magnitude) are selected for gradient updates during each iteration. 
As with the \textsc{ZO-Sparse} variant, this method avoids techniques like LoRA and instead relies on direct selection of high-magnitude weights. 
For complete details on the sparsity mechanism, we refer the reader to the \textsc{ZO-Sparse} description. 
All hyperparameters are in Table~\ref{tbl:fmad-vanilla-hyperparameter}.

\paragraph{A Note on the Theoretical vs.\ Empirical Learning Rate.}
The theoretical convergence bound of ZO (Theorem~\ref{thm:convergence-zero-order}) has the condition of $\eta < \frac{2}{L(1 + \frac{d+1}{n})}$. 
The condition becomes increasingly conservative as $L$ and $d$ scale, which is especially relevant for large models. 
This is a standard limitation of worst-case analysis: 
the bound is derived under minimal assumptions (e.g., global $L$-smoothness, worst-case variance), and thus prioritizes generality over tightness. 
In practice, we start with relatively large learning rates ($10^{-4}$ to $10^{-3}$) to measure the best-case time to convergence for ZO and \fmad.
With adaptive optimizers like \textsc{AdamW}, the learning rate is automatically scaled down during training, often yielding stable and effective performance even when theoretical bounds are violated. 

However, in line with the theory, we observe convergence failures (including NaNs or divergence, see Appendix~\ref{adx:failure-mode-analysis}) when using non-adaptive optimizers such as SGD, especially under large $d/n$ ratios (typically around $10^5$) or for \fmad and ZO methods. 
These failures reinforce that while the theoretical bound is conservative, it qualitatively predicts instability when learning rates are too aggressive relative to dimensionality and batch size (see Appendices~\ref{adx:changing-variance-of-random-pert-sampling} and~\ref{adx:failure-mode-analysis}). 
That said, we do observe (especially in the zero-order case) that overly aggressive learning rates can lead to instability or degraded final performance, in line with the theoretical intuition.
Hence, the theoretical rate serves as a safeguard for convergence analysis rather than a recommended training setting, and that practical hyperparameters typically benefit from empirical tuning beyond what the theory prescribes.
Further discussion is provided in Corollary~\ref{cor:zo-d-over-t}.

\section{Limitations and Future Work}
\label{adx:limitations-future-work}
While the aim of our work was to provide a comprehensive comparison of backpropagation (BP), forward-mode automatic differentiation (\fmad), and zero-order (ZO) optimization strategies, several limitations remain, which can serve as venues for a further exploration. 

First, our experiments focus on deep models, and we did not systematically evaluate backpropagation with checkpointing (\textsc{BP-Checkpointing}) on wider but shallower models. 
In principle, checkpointing may offer less benefit for such architectures. 
However, since wider and shallower models are relatively uncommon in practice, we chose not to extend our evaluations in that direction.
Further, our checkpointing implementation operates at only one granularity (where which activations to checkpoint is not controlled by us) due to current Hugging Face library support, which limits finer control over which activations are saved or recomputed. 
Finer-grained checkpointing could reduce memory usage further and potentially narrow the memory efficiency gap between \textsc{BP-Checkpointing} and ZO methods.
However, this would come at the cost of increased runtime, introducing a different trade-off.
Finally, while we focused on tuning and training \textsc{LoRA} layers, an important future direction would be to extend our comparison framework to full model finetuning. 
Such an extension would allow for a more complete characterization of the trade-offs between memory, time-to-convergence, and accuracy across different gradient computation strategies.

\section{Additional Results}
\label{adx:additional_results}
\subsection{Experimental Variance and Loss Curves}
\label{adx:experimental_variance}

\begin{figure}
     \centering
     \begin{subfigure}[b]{0.32\textwidth}
         \centering
         \includegraphics[width=\textwidth]{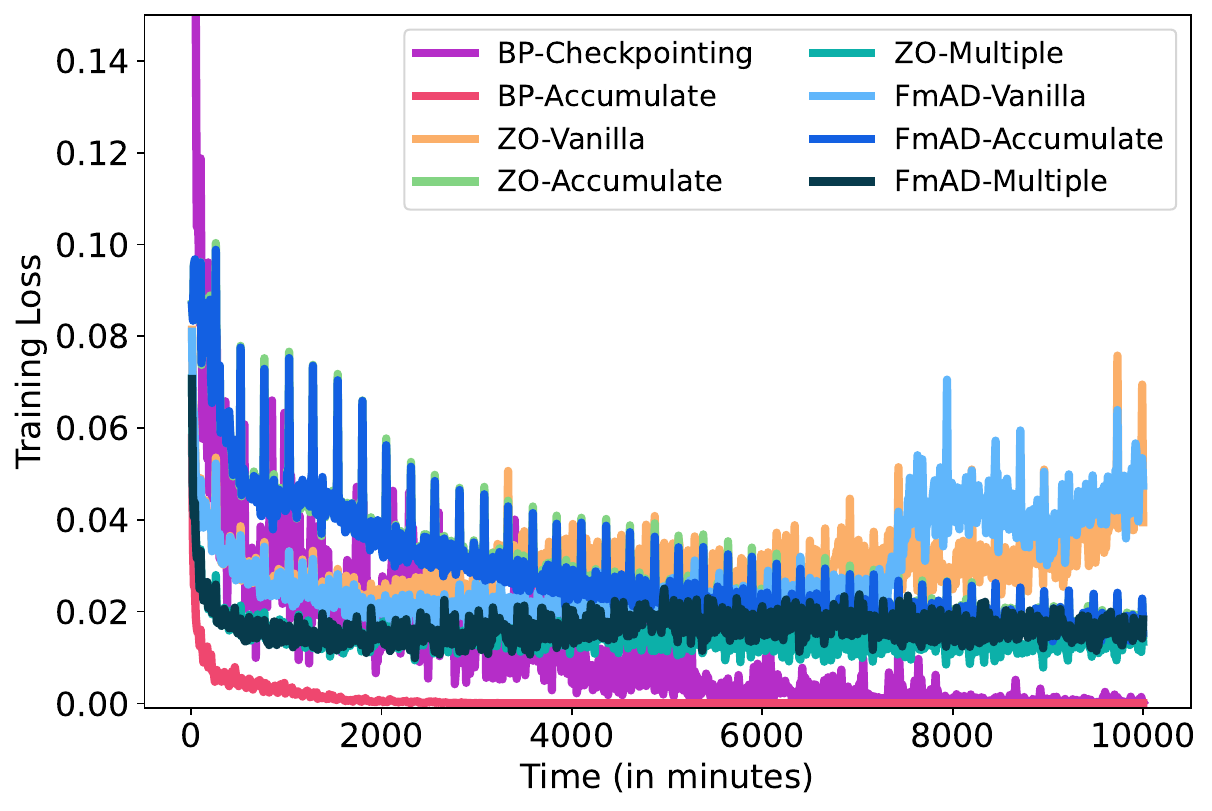}
         \caption{AGNews.}
         \label{fig:agnews-loss}
     \end{subfigure}
     \hfill
     \begin{subfigure}[b]{0.32\textwidth}
         \centering
         \includegraphics[width=\textwidth]{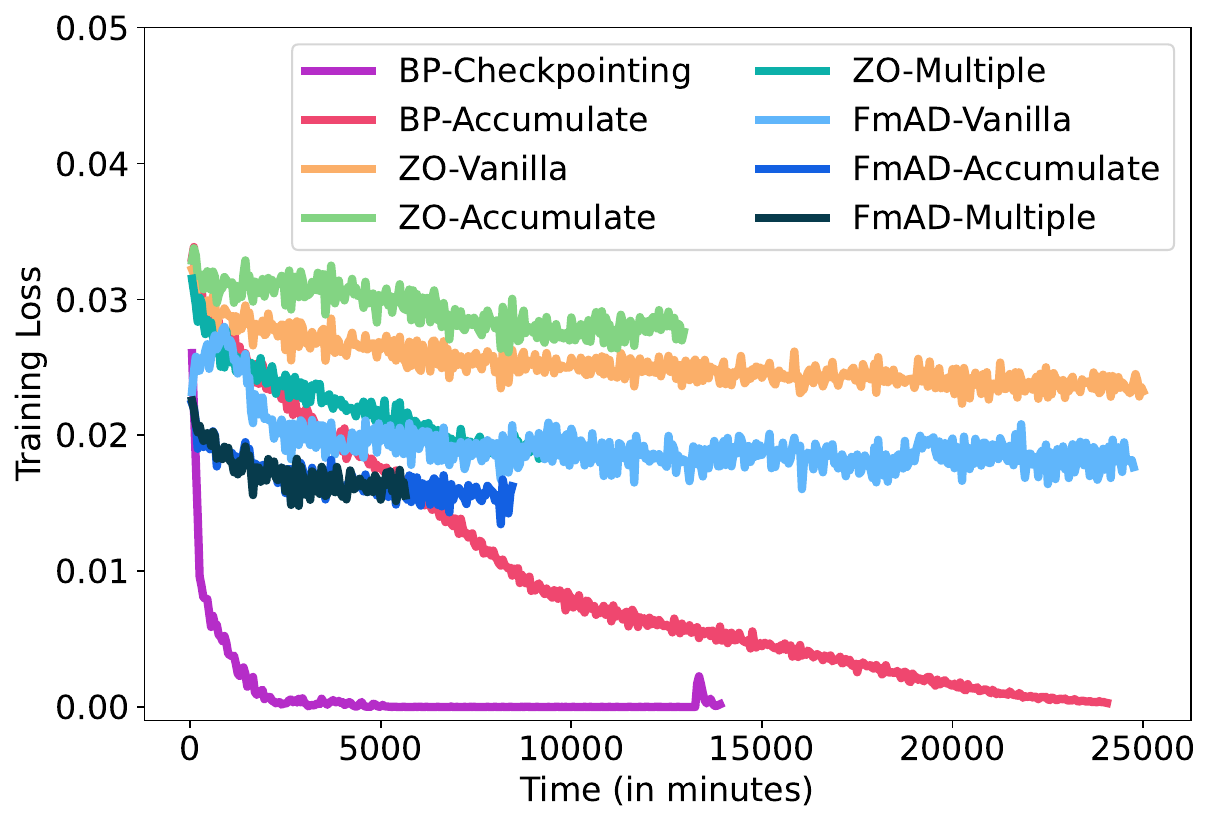}
         \caption{BoolQ.}
         \label{fig:boolq-loss}
     \end{subfigure}
     \hfill
     \begin{subfigure}[b]{0.32\textwidth}
         \centering
         \includegraphics[width=\textwidth]{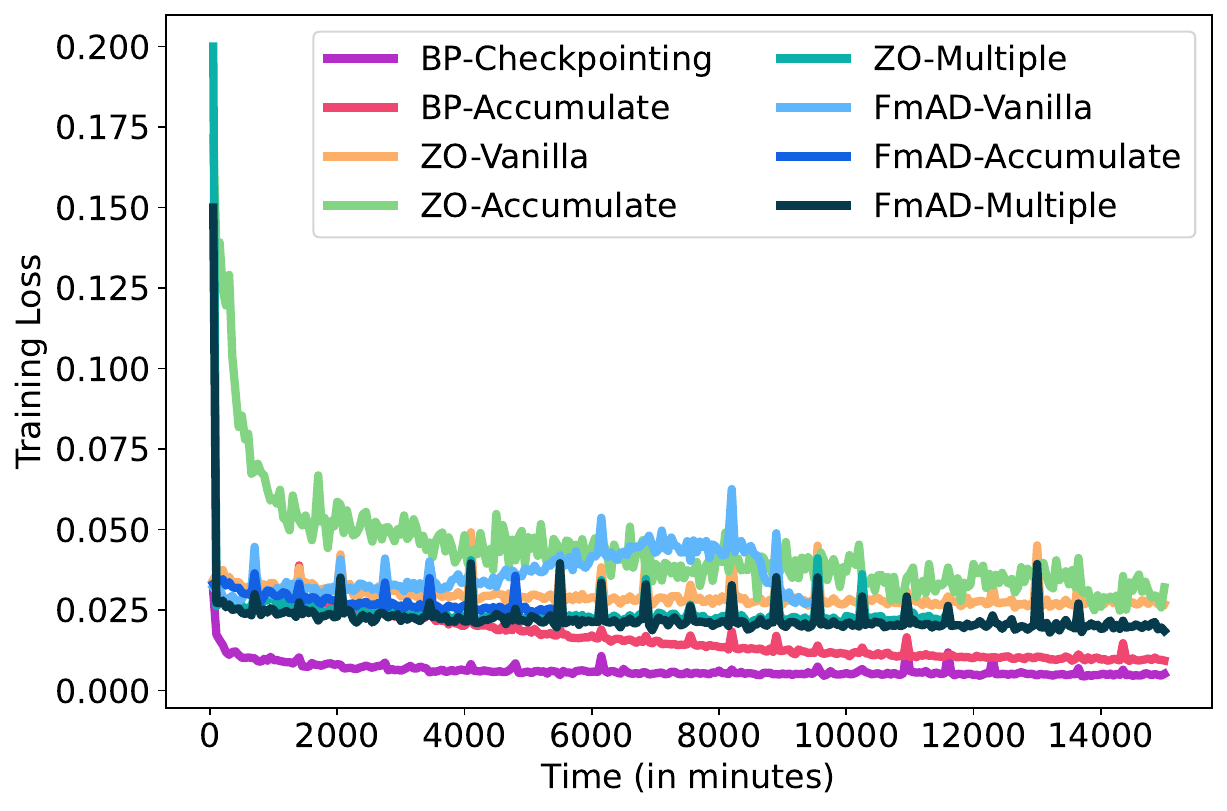}
         \caption{MultiRC.}
         \label{fig:multirc-loss}
     \end{subfigure}
     \hfill
     \begin{subfigure}[b]{0.49\textwidth}
         \centering
         \includegraphics[width=\textwidth]{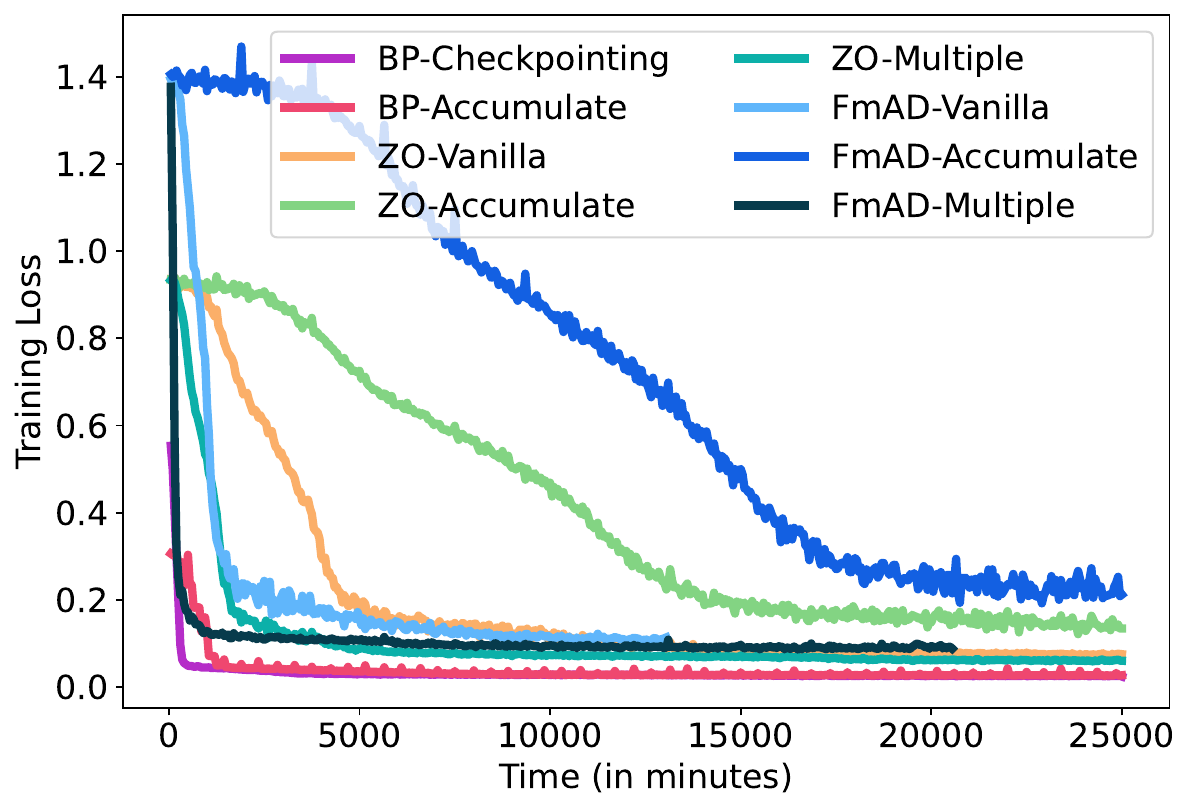}
         \caption{GSM8K.}
         \label{fig:gsm8k-loss}
     \end{subfigure}
     \hfill
     \begin{subfigure}[b]{0.49\textwidth}
         \centering
         \includegraphics[width=\textwidth]{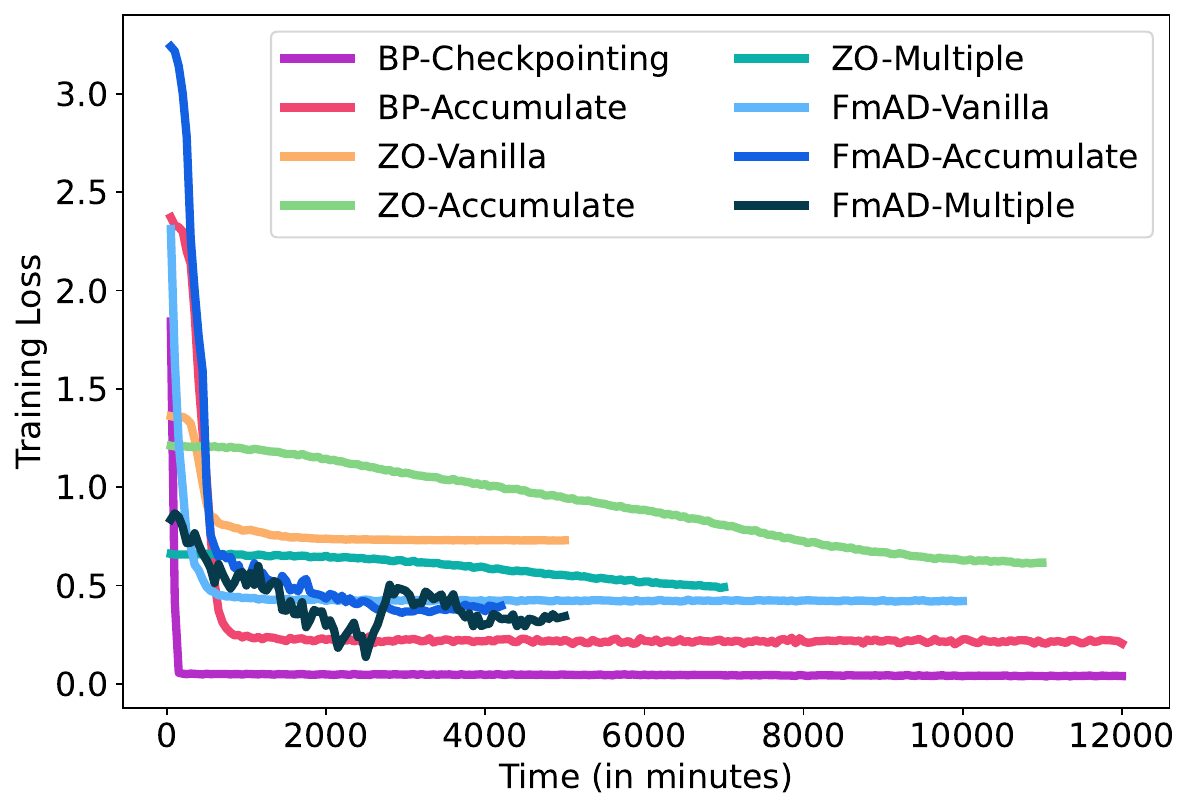}
         \caption{MMLU.}
         \label{fig:mmlu-loss}
     \end{subfigure}
     \hfill
     \begin{subfigure}[b]{0.49\textwidth}
         \centering
         \includegraphics[width=\textwidth]{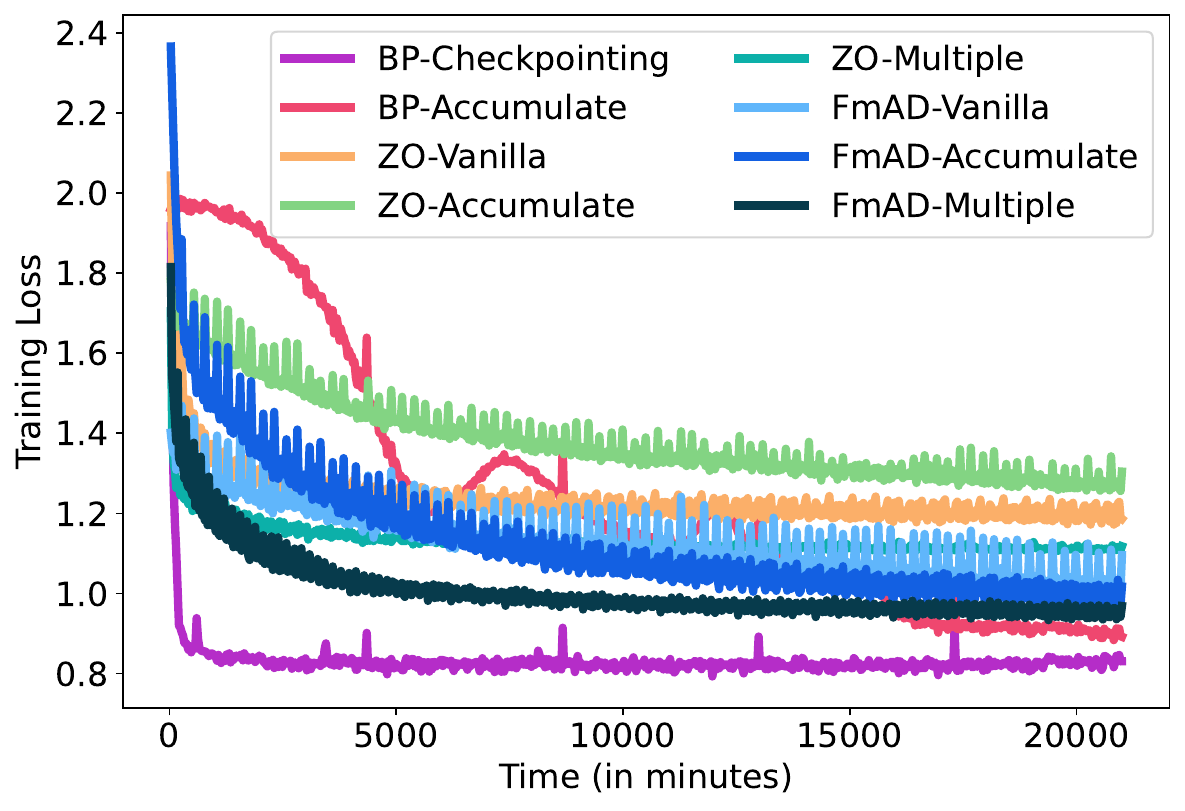}
         \caption{TextVQA.}
         \label{fig:tetxvqa-loss}
     \end{subfigure}
     \hfill
     \begin{subfigure}[b]{0.49\textwidth}
         \centering
         \includegraphics[width=\textwidth]{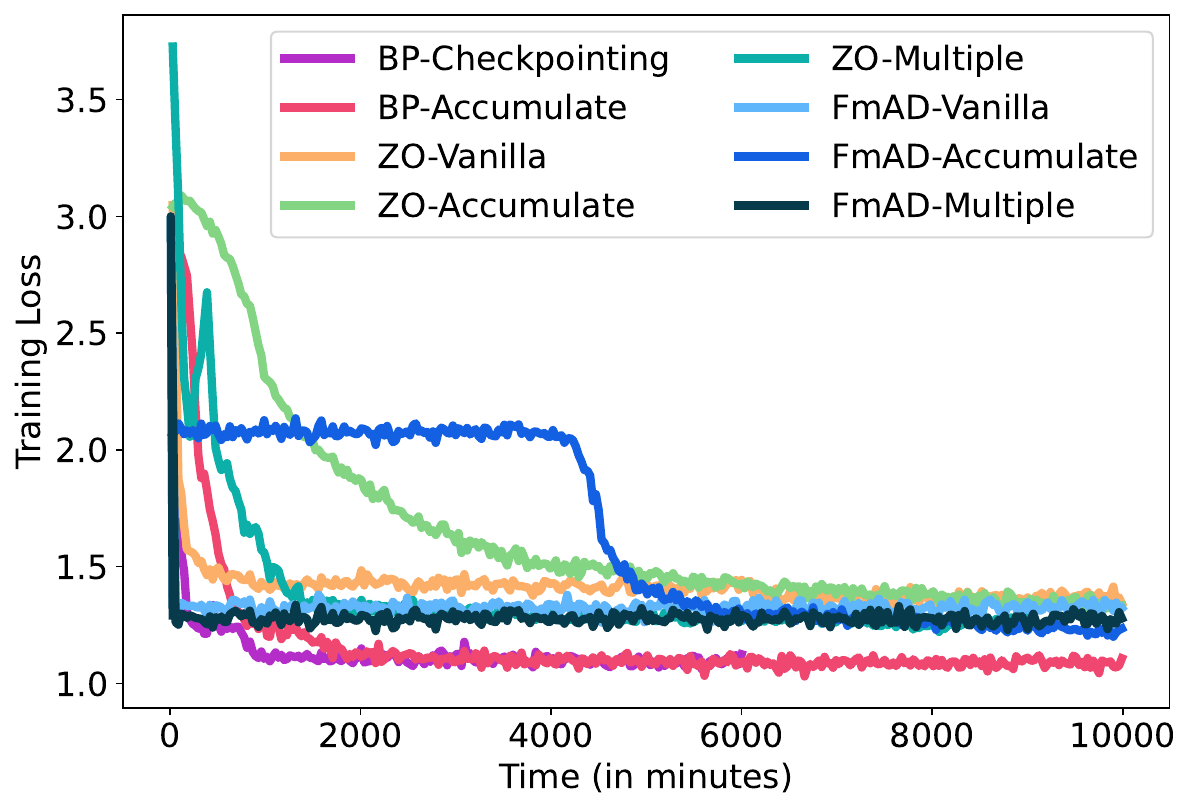}
         \caption{VQAv2.}
         \label{fig:vqav2-loss}
     \end{subfigure}
        \caption{
        Training loss vs.\ training time (in minutes) for \emph{(top)} training \textsc{Llama 3.1} (8B) on three text classification datasets (AGNews, BoolQ, and MultiRC), and \emph{(middle)} two text generation datasets (GSM8K and MMLU).
        \emph{(bottom)} VQAv2 and TextVQA are used to train \textsc{Qwen 2 VL} (7B) on visual question-answering task.
        }
        \label{fig:loss-curves}
        \vspace{-0.7cm}
\end{figure}

Table~\ref{tbl:experimental-variance} shows variance in reported accuracy numbers of Table~\ref{tbl:llama-accuracy}.
For each experiment, we performed three independent runs on seeds 0, 1, and 2.
For each run, we computed the steady-state accuracy (averaged over the final evaluation steps). 
We then reported the mean (in Table~\ref{tbl:llama-accuracy}) and variance (in Table~\ref{tbl:experimental-variance}) computed across these three steady-state accuracies.
\begin{table}[ht]
\centering
\caption{Experimental variance ($\pm$) of test accuracy across three runs with seeds 0, 1, and 2. }
\footnotesize
\begin{tabular}{l|ccccc|cc}
\toprule
\multirow{2}{*}{\diagbox[width=\dimexpr \textwidth/4+1\tabcolsep\relax, height=0.94cm]{Method}{Model + Dataset}} & \multicolumn{5}{c|}{\textsc{Llama 3.1} (8B)} & \multicolumn{2}{c}{\textsc{Qwen} 2 VL (7B)} \\ \cmidrule{2-8} 
 & \multicolumn{1}{c}{AGNews} & \multicolumn{1}{c}{BoolQ} & \multicolumn{1}{c}{MultiRC} & \multicolumn{1}{c}{GSM8K} & \multicolumn{1}{c|}{MMLU} & \multicolumn{1}{c}{VQAv2} & \multicolumn{1}{c}{TextVQA} \\ \midrule
\textsc{BP-Vanilla} & 0.46 & 0.54 & 0.59 & 0.41 & 0.63 & 1.49 & 0.78 \\
\textsc{BP-Checkpointing} & 0.45 & 0.56 & 0.62 & 0.39 & 0.61 & 1.52 & 0.77 \\
\textsc{BP-Accumulate} & 0.67 & 0.78 & 0.84 & 0.79 & 0.69 & 1.71 & 0.98 \\ \midrule
\textsc{ZO-Vanilla} & 0.98 & 0.76 & 0.8 & 0.55 & 0.95 & 1.22 & 0.89 \\
\textsc{ZO-Accumulate} & 0.84 & 0.72 & 0.76 & 0.53 & 0.84 & 1.16 & 0.85 \\
\textsc{ZO-Multiple} & 0.79 & 0.64 & 0.67 & 0.53 & 0.86 & 1.13 & 0.86 \\
\textsc{ZO-Adaptive} & 1.02 & 0.95 & 1.13 & 0.84 & 0.83 & 0.95 & 0.78 \\
\textsc{ZO-SVRG} & 0.94 & 1.03 & 0.92 & 0.82 & 0.46 & 1.02 & 1.13 \\
\textsc{ZO-Sparse} & 0.53 & 0.67 & 0.62 & 0.34 & 1.03 & 0.89 & 0.9 \\
\textsc{MeZO} & 0.86 & 0.73 & 0.73 & --- & --- & --- & --- \\ \midrule
\textsc{FmAD-Vanilla} & 0.81 & 0.72 & 0.64 & 0.73 & 0.86 & 1.34 & 0.92 \\
\textsc{FmAD-Accumulate} & 0.69 & 0.73 & 0.80 & 0.62 & 0.78 & 0.91 & 0.95 \\
\textsc{FmAD-Multiple} & 0.85 & 0.77 & 0.89 & 1.04 & 0.96 & 0.74 & 0.83 \\
\textsc{FmAD-Adaptive} & 1.63 & 1.25 & 1.34 & 0.95 & 1.52 & 1.11 & 1.31 \\
\textsc{FmAD-SVRG} & 1.42 & 0.96 & 0.89 & 1.02 & 1.44 & 1.05 & 1.29 \\
\textsc{FmAD-Sparse} & 0.93 & 0.75 & 1.10 & 0.54 & 0.67 & 1.24 & 0.93 \\ \bottomrule
\end{tabular}
\label{tbl:experimental-variance}
\end{table}
Furthermore, Figure~\ref{fig:loss-curves} illustrates the training loss curves with respect to the training time, highlighting the convergence behavior.
We have only showed the best-performing baselines to maintain clarity.

\begin{figure}[!h]
     \centering
     \begin{subfigure}[b]{0.32\textwidth}
         \centering
         \includegraphics[width=\textwidth]{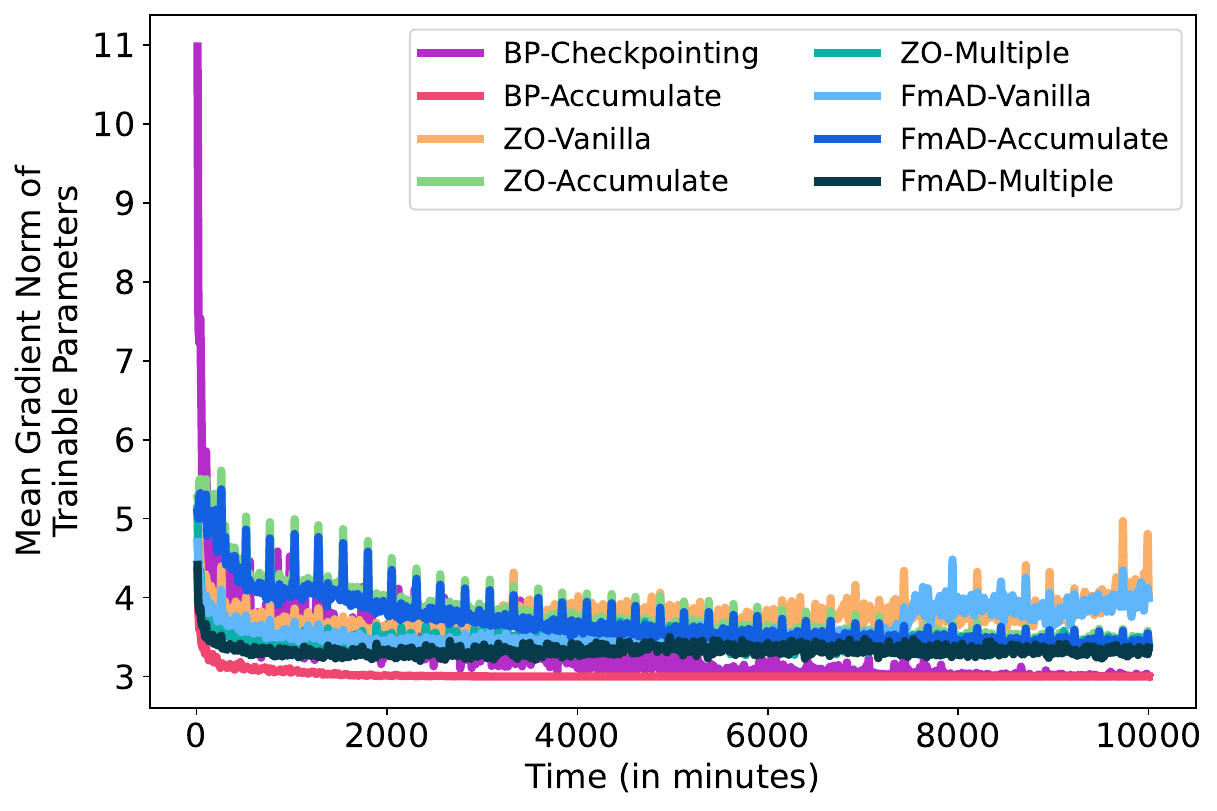}
         \caption{AGNews}
         \label{fig:agnews-grad-norm}
     \end{subfigure}
     \hfill
     \begin{subfigure}[b]{0.32\textwidth}
         \centering
         \includegraphics[width=\textwidth]{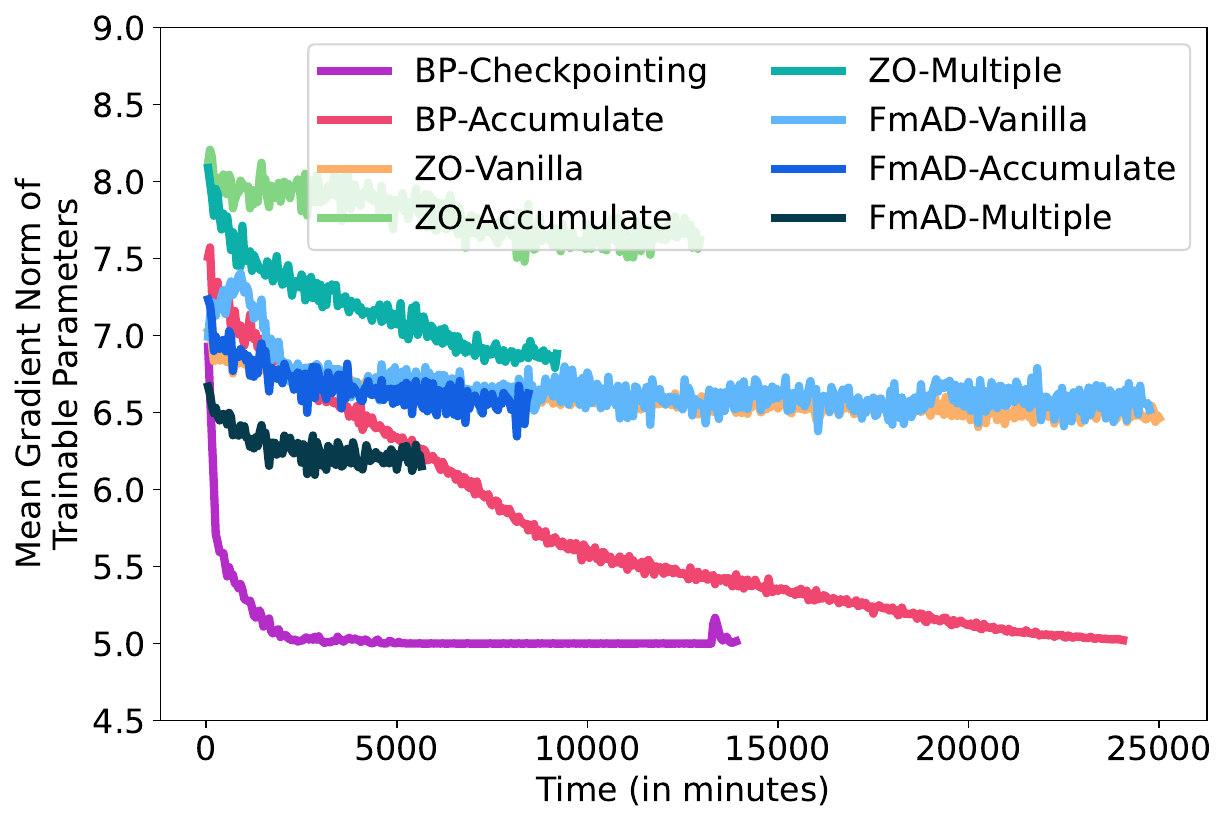}
         \caption{BoolQ}
         \label{fig:boolq-grad-norm}
     \end{subfigure}
     \hfill
     \begin{subfigure}[b]{0.32\textwidth}
         \centering
         \includegraphics[width=\textwidth]{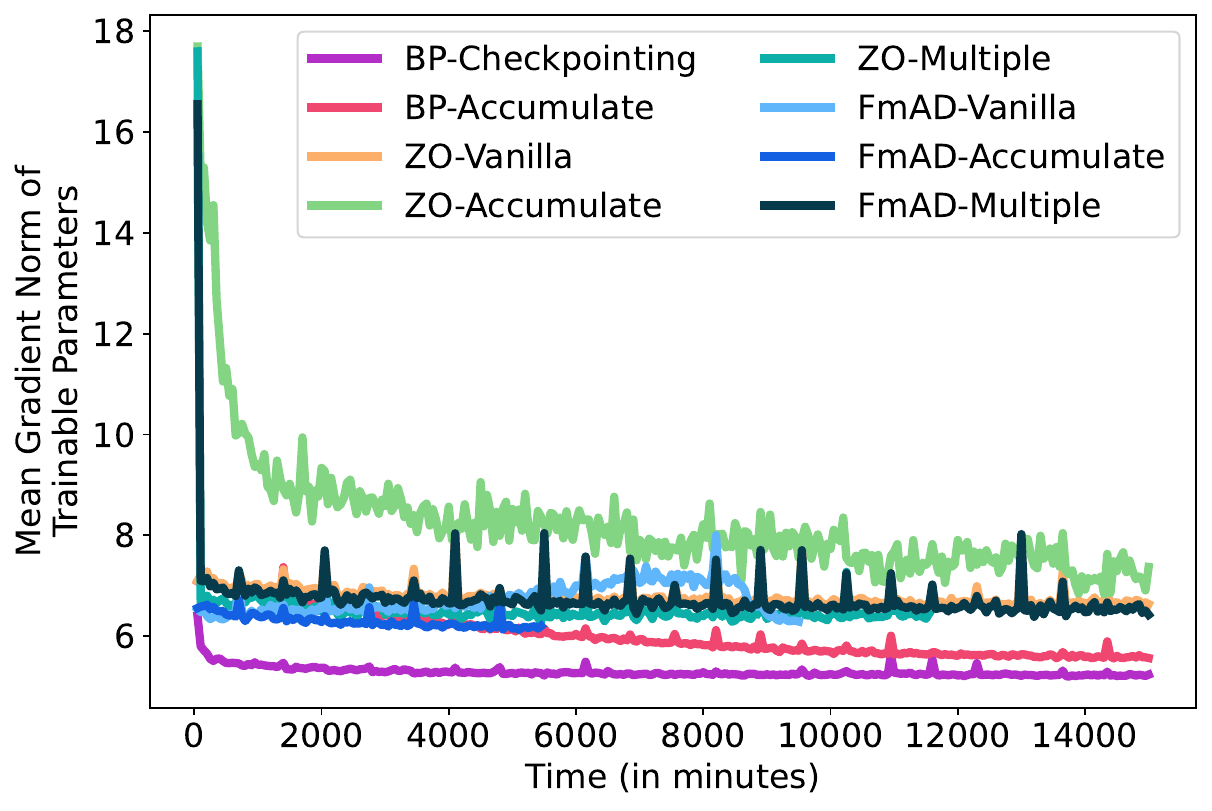}
         \caption{MultiRC}
         \label{fig:multirc-grad-norm}
     \end{subfigure}
     \hfill
     \begin{subfigure}[b]{0.49\textwidth}
         \centering
         \includegraphics[width=\textwidth]{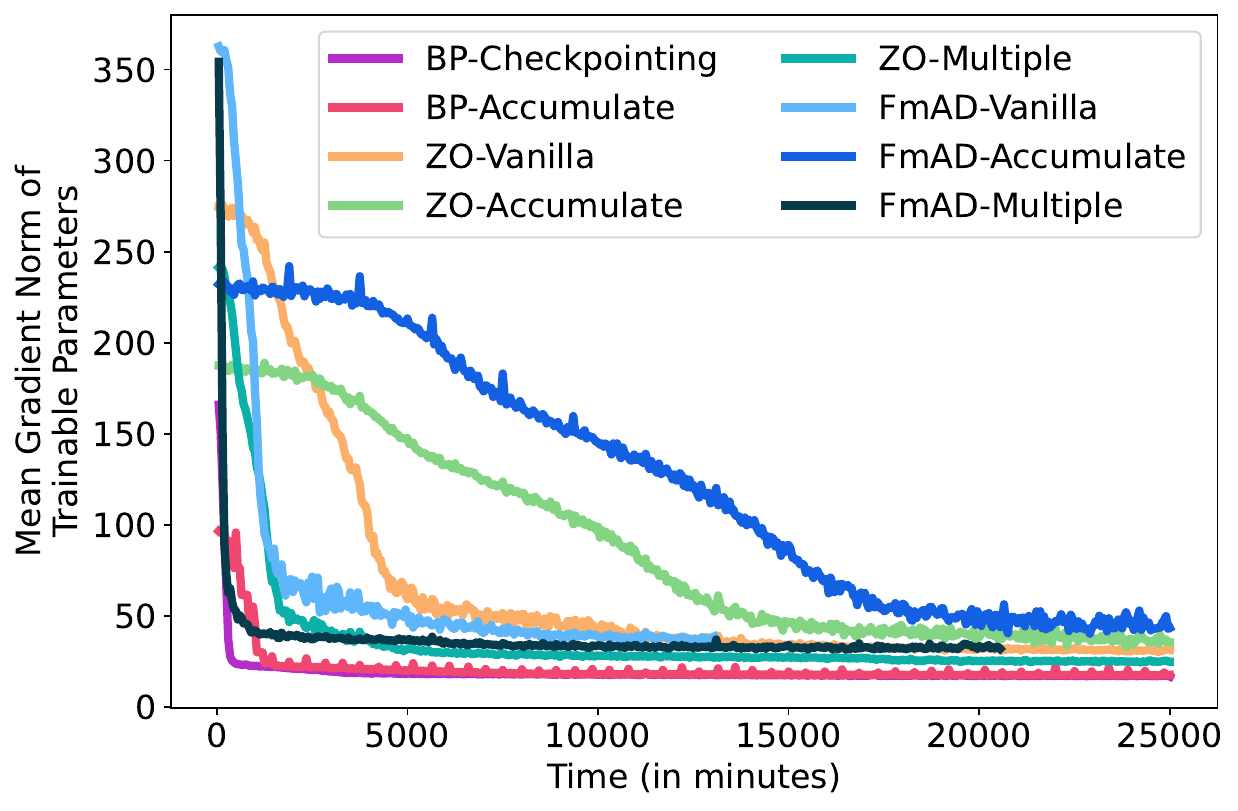}
         \caption{GSM8K}
         \label{fig:gsm8k-grad-norm}
     \end{subfigure}
     \hfill
     \begin{subfigure}[b]{0.49\textwidth}
         \centering
         \includegraphics[width=\textwidth]{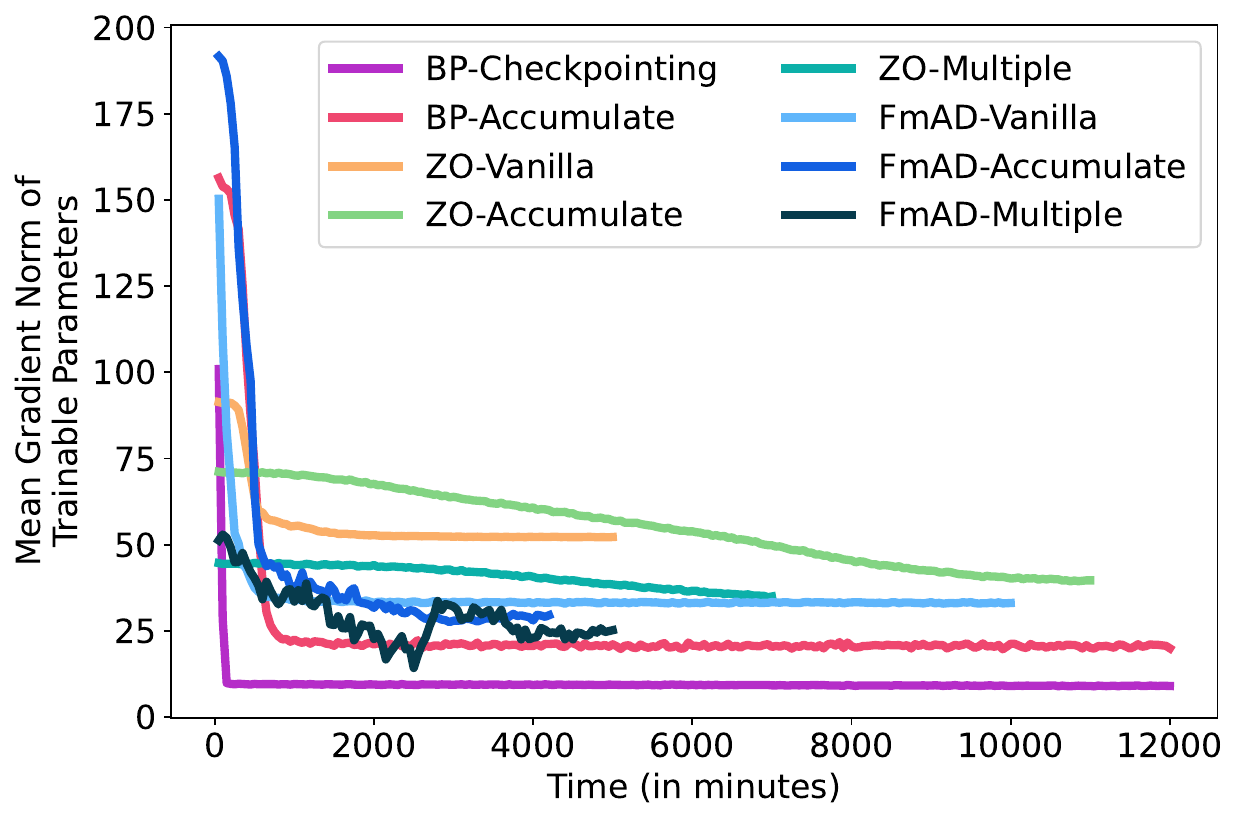}
         \caption{MMLU}
         \label{fig:mmlu-grad-norm}
     \end{subfigure}
     \hfill
     \begin{subfigure}[b]{0.49\textwidth}
         \centering
         \includegraphics[width=\textwidth]{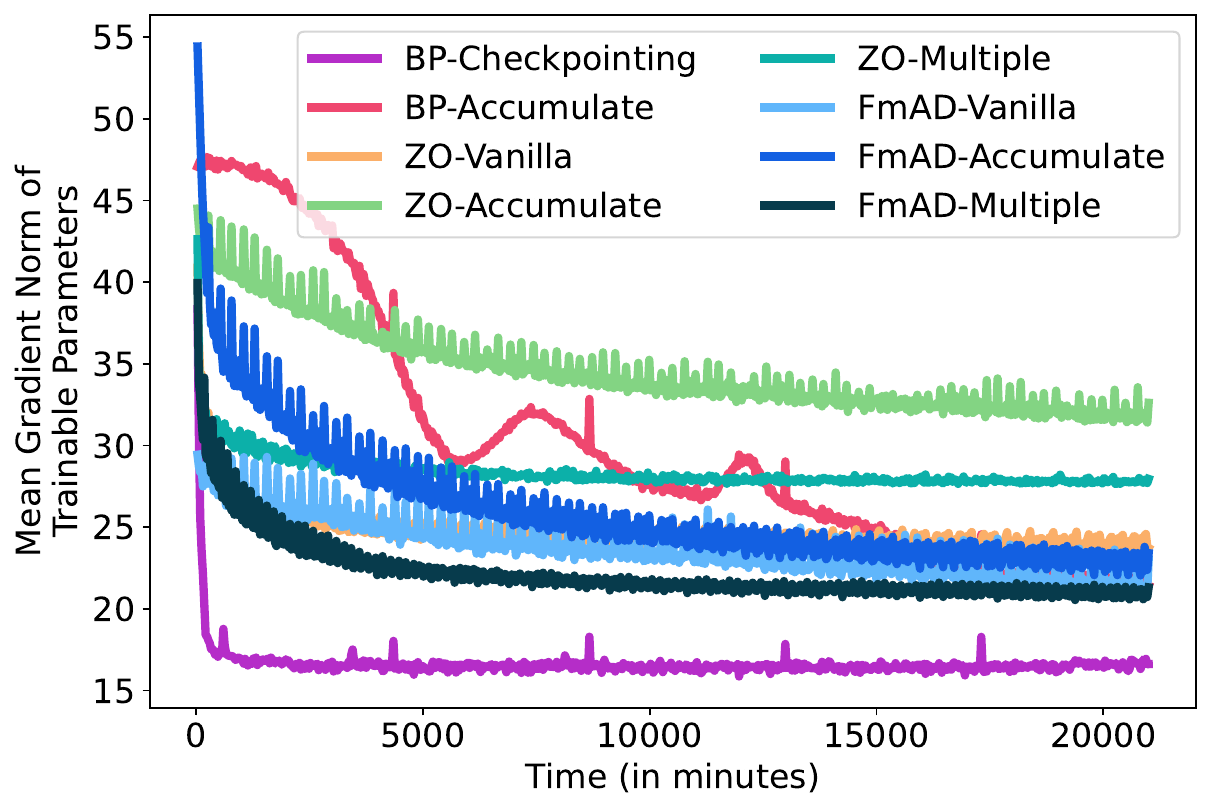}
         \caption{TextVQA}
         \label{fig:tetxvqa-grad-norm}
     \end{subfigure}
     \hfill
     \begin{subfigure}[b]{0.49\textwidth}
         \centering
         \includegraphics[width=\textwidth]{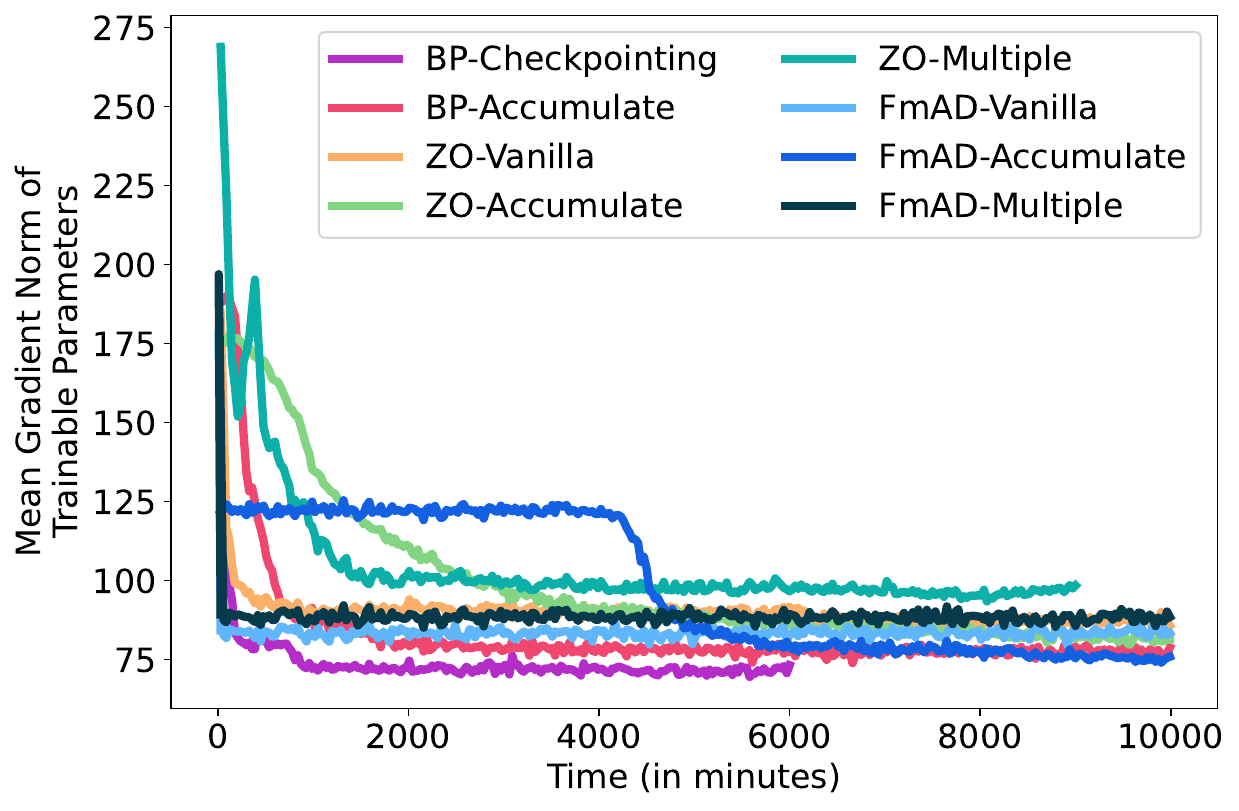}
         \caption{VQAv2}
         \label{fig:vqav2-grad-norm}
     \end{subfigure}
        \caption{
        Gradient norm vs.\ training time (in minutes) for \emph{(top)} training \textsc{Llama 3.1} (8B) on three text classification datasets (AGNews, BoolQ, and MultiRC), and \emph{(middle)} two text generation datasets (GSM8K and MMLU).
        \emph{(bottom)} VQAv2 and TextVQA are used to train \textsc{Qwen 2 VL} (7B) on visual question-answering task.
        }
        \label{fig:grad-norm-curves}
\end{figure}

 {Figure~\ref{fig:grad-norm-curves} reports the mean gradient norm across all trainable parameters for all the datasets on \textsc{Llama 3.1 (8B)} model. 
These curves closely mirror the loss trajectories reported in Figure~\ref{fig:loss-curves}, exhibiting similar convergence tendencies across all methods.
Specifically, \textsc{BP-Checkpointing} shows the steepest and most stable decay in gradient norm, aligning with its superior convergence behavior in loss and accuracy.
This strengthens the consistency between the theoretical observations of \S~\ref{sec:theoretical-analysis} (which centers on the gradient norm) and our empirical findings of \S~\ref{sec:empirical-evaluation}.}

\subsection{More Dataset-Model Combinations}
\label{adx:more-dataset-model-combinations}

Tables~\ref{tab:gsm8k-opt13b} and~\ref{tab:vqav2-qwen2-7b} show runtime and compute-to-convergence results on \textsc{GSM8K-OPT13B} and \textsc{VQAv2-Qwen7B} dataset-model combinations. 
We observe a similar trend: \textsc{BP-Checkpointing} requires the least total compute to reach convergence, even though ZO and \fmad variants have lower per-iteration costs, because they need more iterations to converge.

\begin{table}[!h]
\caption{Compute-to-convergence and wall-clock comparisons for GSM8K (Free-form Math Reasoning Dataset) on OPT (13B) model.}
\centering
\footnotesize
\label{tab:gsm8k-opt13b}
\begin{tabular}{lcccc}
\toprule
\textbf{Method}  & \begin{tabular}[c]{@{}c@{}}\textbf{TFLOPs}\\ \textbf{per Iter.} ($\downarrow$)\end{tabular} & \begin{tabular}[c]{@{}c@{}}\textbf{TFLOPs until}\\ \textbf{Convergence} ($\downarrow$) ($\times 10^4$)\end{tabular} & \begin{tabular}[c]{@{}c@{}}\# \textbf{Iter. until}\\ \textbf{Convergence} ( $\times 10^3$)\end{tabular} & \begin{tabular}[c]{@{}c@{}}\textbf{Runtime in}\\ \textbf{Minutes} ($\downarrow$)\end{tabular} \\
\midrule
\textsc{BP-Checkpointing} & \phantom{0}2458.9 & \phantom{00}542.1 & \phantom{00}2.20 & \phantom{00}250 \\
\textsc{ZO-Vanilla} & \phantom{0}1301.2 & \phantom{0}2886.7 & \phantom{0}22.18 & \phantom{0}3050 \\
\textsc{ZO-Multiple} & 13011.4 & 29115.3 & \phantom{0}22.38 & \phantom{0}4240 \\
\textsc{ZO-Accumulate} & \phantom{0}1301.2 & 24014.4 & 184.55 & 14930 \\
\textsc{FmAD-Vanilla} & \phantom{0}2146.4 & \phantom{0}2429.7 & \phantom{0}11.32 & \phantom{0}2130 \\
\textsc{FmAD-Multiple} & 21465.6 & 25637.0 & \phantom{0}11.94 & \phantom{0}3510 \\
\textsc{FmAD-Accumulate} & \phantom{0}2146.4 & 21592.0 & 100.60 & 11440 \\
\bottomrule
\end{tabular}
\end{table}

\begin{table}[!h]
\caption{Compute-to-convergence and wall-clock comparisons for \textsc{VQAv2} (Visual Question Answering) on \textsc{Qwen} 2 VL (7B) model.}
\centering
\footnotesize
\label{tab:vqav2-qwen2-7b}
\begin{tabular}{lcccc}
\toprule
\textbf{Method}  & \begin{tabular}[c]{@{}c@{}}\textbf{TFLOPs}\\ \textbf{per Iter.} ($\downarrow$)\end{tabular} & \begin{tabular}[c]{@{}c@{}}\textbf{TFLOPs until}\\ \textbf{Convergence} ($\downarrow$) ($\times 10^4$)\end{tabular} & \begin{tabular}[c]{@{}c@{}}\# \textbf{Iter. until}\\ \textbf{Convergence} ( $\times 10^3$)\end{tabular} & \begin{tabular}[c]{@{}c@{}}\textbf{Runtime in}\\ \textbf{Minutes} ($\downarrow$)\end{tabular} \\
\midrule
\textsc{BP-Checkpointing} & \phantom{0}620.5 & \phantom{0}130.3 & \phantom{0}2.1 & 1240 \\
\textsc{ZO-Vanilla} & \phantom{0}412.4 & \phantom{0}445.4 & 10.8 & \phantom{0}840 \\
\textsc{ZO-Multiple} & 4124.0 & 4289.0 & 10.4 & 2250 \\
\textsc{ZO-Accumulate} & \phantom{0}412.4 & 3629.1 & 88.1 & 6780 \\
\textsc{FmAD-Vanilla} & \phantom{0}617.0 & \phantom{0}376.4 & \phantom{0}6.1 & 1030 \\
\textsc{FmAD-Multiple} & 6170.0 & 6910.4 & 11.2 & 2150 \\
\textsc{FmAD-Accumulate} & \phantom{0}617.0 & 3418.9 & 55.4 & 6290 \\
\bottomrule
\end{tabular}
\end{table}

\clearpage
\subsection{Experiments with Medium-sized Models}
\label{adx:experiments-with-medium-sized-models}
The goal of these experiments was to investigate whether forward-mode automatic differentiation (\fmad) and zero-order (ZO) optimization could perform competitively when applied to medium-sized models, specifically BERT Base (110M), BERT Large (340M), \textsc{RoBERTa} Base (125M), and \textsc{RoBERTa} Large (350M). 
While \fmad and ZO have shown some promise on very small-scale problems in prior work~\citep{chen2019zo, cobb2024secondorderforwardmodeautomaticdifferentiation, rostami2024projectedforwardgradientguidedfrankwolfe}, it remained an open question whether the convergence speed could scale reasonably with model sizes.

\begin{figure}[h]
    \centering
    \begin{subfigure}[b]{0.49\textwidth}
        \centering
        \includegraphics[width=\linewidth]{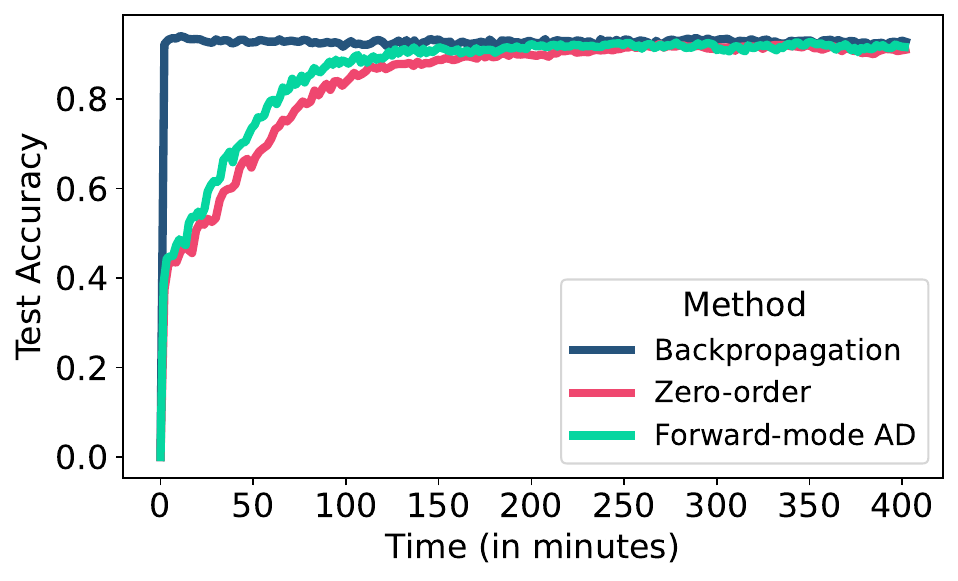}
        \caption{BERT Base (110M)}
        \label{fig:bert-base-runtime}
    \end{subfigure}\hfill
    \begin{subfigure}[b]{0.49\textwidth}
        \centering
        \includegraphics[width=\linewidth]{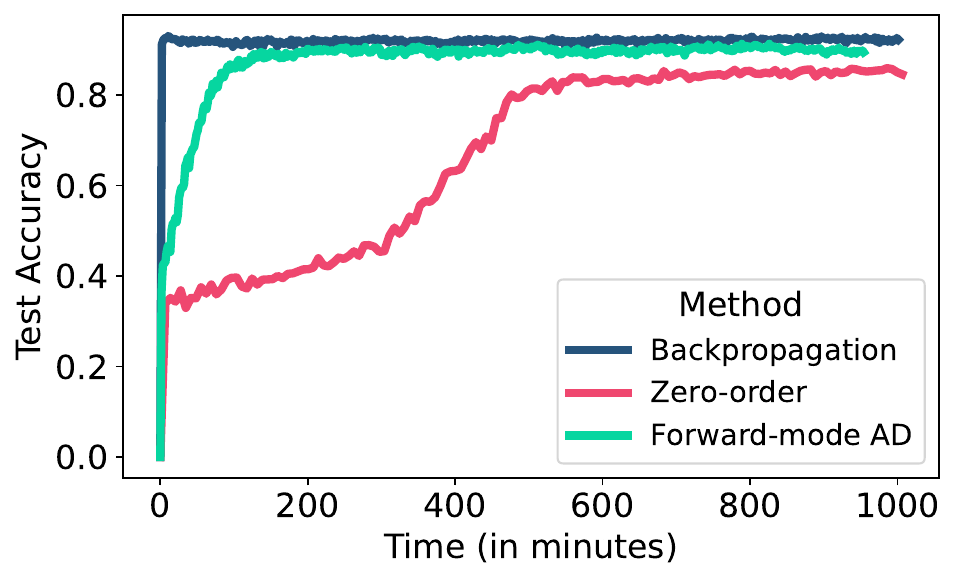}
        \caption{BERT Large (340M)}
        \label{bert-large-runtime}
    \end{subfigure}
    \hfill
    \begin{subfigure}[b]{0.49\textwidth}
        \centering
        \includegraphics[width=\linewidth]{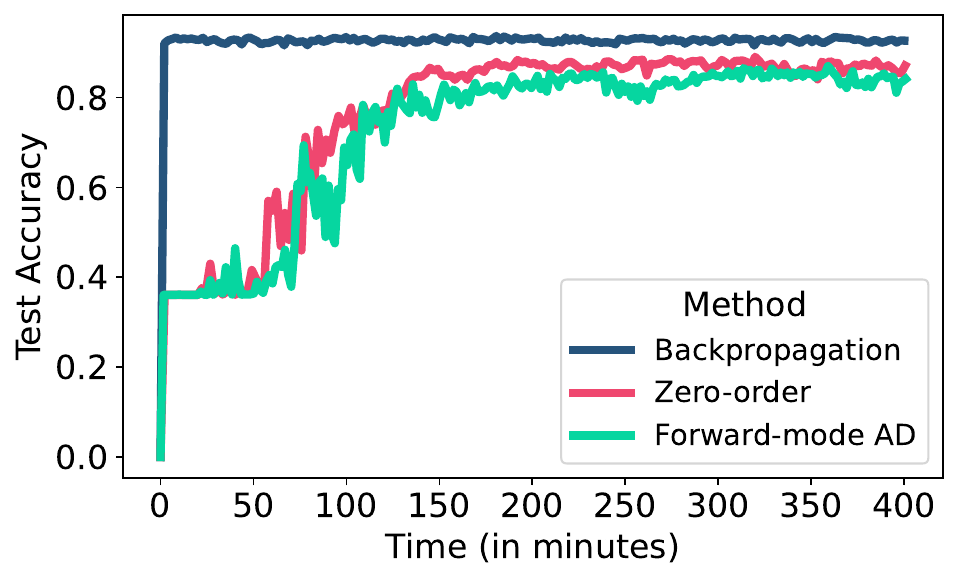}
        \caption{\textsc{RoBERTa} Base (125M)}
        \label{roberta-base-runtime}
    \end{subfigure}
    \hfill
    \begin{subfigure}[b]{0.49\textwidth}
        \centering
        \includegraphics[width=\linewidth]{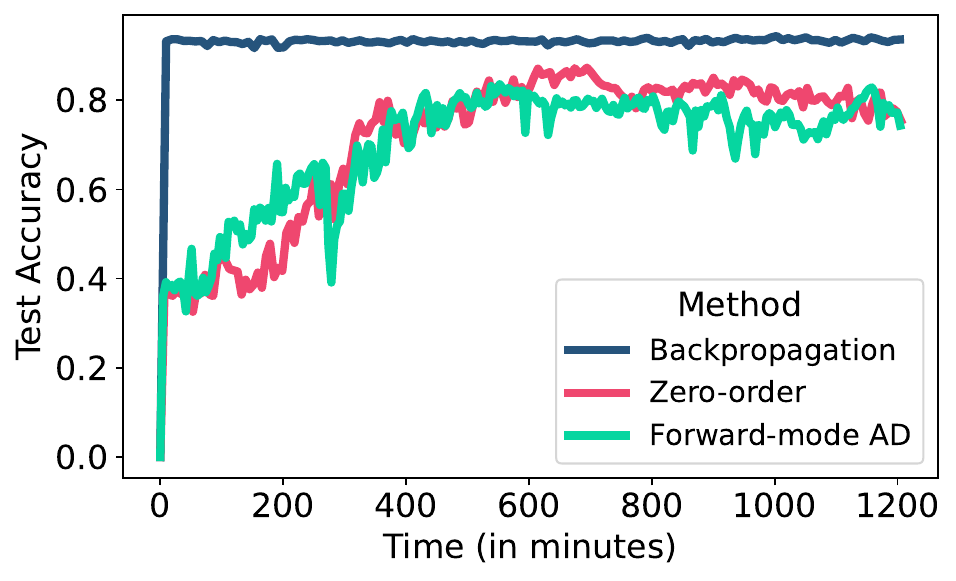}
        \caption{\textsc{RoBERTa} Large (350M)}
        \label{roberta-large-runtime}
    \end{subfigure}
    \caption{Accuracy versus training time comparison across Backpropagation (with checkpointing), Zero-order (ZO), and Forward-mode AD (\fmad) on BERT (Base and Large) and \textsc{RoBERTa} (Base and Large).
    Even at a smaller scale of trainable parameter count, ZO and \fmad either fail to reach to the accuracy of backpropagation (in case of \textsc{RoBERTa}), or takes longer to reach to the desired accuracy (in case of BERT).}
    \label{fig:bert-roberta-runtimes}
\end{figure}

Figure~\ref{fig:bert-roberta-runtimes} highlights a clear and consistent trend: 
backpropagation (with checkpointing) achieves superior convergence speed and final test accuracy, even for medium-sized models, compared to \fmad and ZO methods. 
Even for BERT Base (110M weights), \fmad and ZO lag significantly behind backpropagation in terms of convergence rate. 
While \fmad and ZO eventually approach a comparable final accuracy (with a gap of 0.74--1.66\%) for BERT Base, they require substantially more training time to do so.

As we scale to larger models, BERT Large and \textsc{RoBERTa} variants, the performance of \fmad and ZO deteriorates further. 
Both methods experience slower convergence, greater instability, and often plateau at lower final accuracies (with a drop of 1.19--6.71\% for BERT Large, 6.76--7.62\% for \textsc{RoBERTa} Base, 9.33--12.98\% for \textsc{RoBERTa} Large) despite extensive training. 
ZO, in particular, struggles to reach acceptable performance, while \fmad shows increasingly volatile learning curves.

In summary, our experiments confirm that \fmad and ZO are fundamentally limited in their ability to compete with backpropagation in realistic settings. 
Their inefficiency becomes increasingly pronounced as we evaluate accuracy, along side memory consumption and time-to-convergence.

\subsection{Changing Variance of Random Perturbation Sampling}
\label{adx:changing-variance-of-random-pert-sampling}

\begin{wrapfigure}{r}{0.4\textwidth}
    \centering
    \vspace{-0.3cm}
    \includegraphics[width=\linewidth]{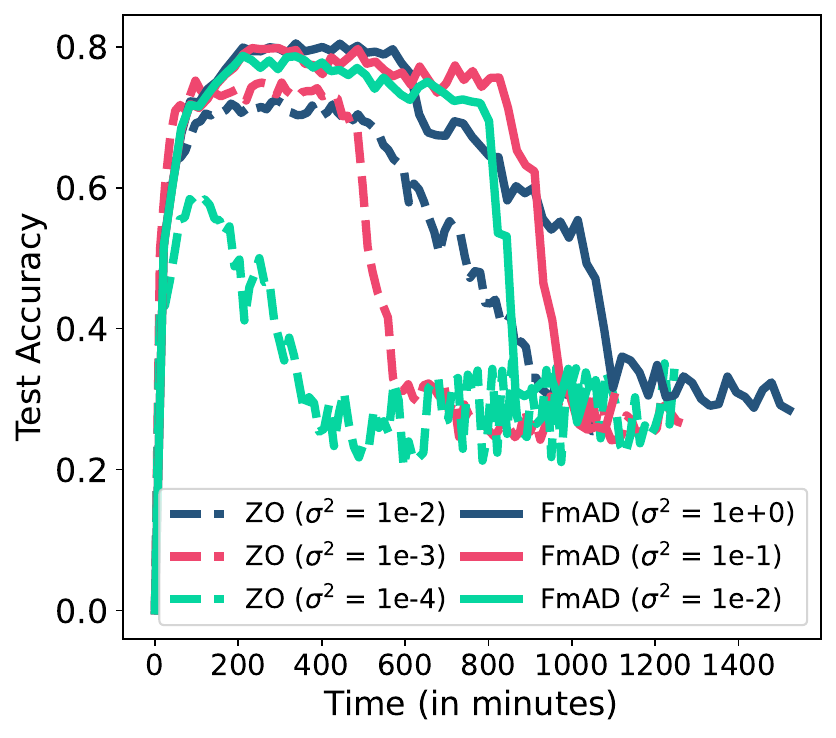}
    \caption{Changing variance $\sigma^2$ of random sampling of perturbations.
    Directly reducing randomness variance does not lead to reduced noise in the gradients.}
    \label{fig:reducing-perturbation-variance}
    \vspace{-0.7cm}
\end{wrapfigure}

We examine the effect of variance $\sigma^2$ of random perturbations which are sampled from Gaussian distribution $\cN(0, \sigma^2)$ on the accuracy performance of \fmad and ZO.
Figure~\ref{fig:reducing-perturbation-variance} presents test accuracy over time for different values of $\sigma^2$, ranging from 1 to $10^{-2}$ for \fmad and from $10^{-2}$ to $10^{-4}$ for ZO.\\
The results reveal a strong sensitivity to the choice of variance: 
small variances reduce the diversity of perturbations, while large variances introduce excessive noise in high-dimensions, destabilizing training. 
Both \fmad and ZO achieve their best performance at intermediate values, $\sigma^2 = 1$ for \fmad and $\sigma^2 = 10^{-3}$ for {ZO}, which balance signal strength and noise.
For {ZO}, reducing the variance from $10^{-3}$ to $10^{-4}$ results in a sharp accuracy drop of 13.75\%. 
In contrast, \fmad shows a more gradual decline of 0.94\% and 2.27\% as $\sigma^2$ decreases from $1$ to $10^{-1}$ and $10^{-2}$, respectively. 
The lower optimal variance for {ZO} arises from its gradient estimator, which includes an explicit division by the perturbation variance to scale the update magnitude (Equation~\ref{eq:zo-one-layer-computations}).
These findings suggest that simply reducing variance of the distribution from which perturbations are sampled does not result in better gradient estimates, nor does it improve convergence.

\subsection{Increasing Trainable Parameter Count}
\label{subsec:reducing_trainable_parameter_count}

\begin{table}[t]
    \centering
    \begin{minipage}{0.49\textwidth}
        \centering
        \footnotesize
        \setlength{\tabcolsep}{4pt}
        \caption{Accuracy of BP, ZO, and \fmad under model size scaling: 
        Only \textsc{BP-Checkpointing} (abbreviated as \textsc{BP-Chkpt}) maintains high accuracy as model size increases.}
        \begin{tabular}{l|c|c}
        \toprule
        \begin{tabular}[c]{@{}l@{}}OPT Variants\end{tabular} & \begin{tabular}[c]{@{}c@{}}Variant Size\end{tabular} & Accuracy ($\uparrow$) \\
        \midrule
        \multirow{3}{*}{\textsc{BP-Chkpt}} & \phantom{0}1.3B & 94.08 \\
         & \phantom{0}6.7B & 94.35 \\
         & 13.0B & 94.51 \\
         \midrule
        \multirow{3}{*}{\textsc{ZO-Vanilla}} & \phantom{0}1.3B & 73.16 \\
         & \phantom{0}6.7B & 65.75 \\
         & 13.0B & 71.00 \\
         \midrule
        \multirow{3}{*}{\textsc{FmAD-Vanilla}} & \phantom{0}1.3B & 88.28 \\
         & \phantom{0}6.7B & 87.50 \\
         & 13.0B & 77.07 \\ 
         \bottomrule
        \end{tabular}
        \label{tbl:opt-accuracy-size-variants}
    \end{minipage}
    \hfill
    \begin{minipage}{0.49\textwidth}
        \centering
        \setlength{\tabcolsep}{4pt}
        \footnotesize
        \caption{Accuracy as the \textsc{LoRA} rank increases for OPT 6.7B: 
        \textsc{BP-Checkpointing} remains robust, while \fmad becomes unstable and ZO shows minimal gains.}
        \begin{tabular}{l|c|c}
        \toprule
        OPT 6.7B & \begin{tabular}[c]{@{}c@{}}\textsc{LoRA} Rank\end{tabular} & Accuracy ($\uparrow$) \\
        \midrule
        \multirow{3}{*}{\textsc{BP-Chkpt}} & \phantom{0}1 & 94.35 \\
         & 16 &  88.44 \\
         & 32 &  85.54 \\
         \midrule
        \multirow{3}{*}{\textsc{ZO-Vanilla}} & \phantom{0}1 & 65.75 \\
         & 16 & 68.07 \\
         & 32 & 68.97 \\
         \midrule
        \multirow{3}{*}{\textsc{FmAD-Vanilla}} & \phantom{0}1  & 87.50 \\
         & 16 & \jvp = \texttt{NaN} \\
         & 32 & \jvp = \texttt{NaN} \\
         \bottomrule
        \end{tabular}
        \label{tbl:opt-accuracy-lora-variants}
    \end{minipage}
\end{table}

\begin{table}[!h]
\caption{Accuracy comparison across LoRA ranks (1 to 128) and with full fine-tuning for \textsc{BP-Checkpointing}, \textsc{ZO-Vanilla}, and \textsc{FmAD-Vanilla} on AGNews with OPT-13B. 
\textsc{BP-Checkpointing} remains robust across ranks, while \fmad/ZO degrade with increasing rank (becoming unstable or diverging at higher capacity) supporting that BP-free methods fail even in favorable low-rank regimes and do not scale to full fine-tuning.}
\footnotesize
\centering
\label{tab:higher-rank-lora}
\begin{tabular}{lll}
\toprule
\textbf{Method} & \textbf{Rank} & \textbf{Accuracy (\%, $\uparrow$)} \\
\midrule
\textsc{BP-Checkpointing} & \multirow{3}{*}{1} & 94.51 \\
\textsc{ZO-Vanilla} &  & 71.00 \\
\textsc{FmAD-Vanilla} &  & 77.07 \\
\midrule
\textsc{BP-Checkpointing} & \multirow{3}{*}{32} & 87.95 \\
\textsc{ZO-Vanilla} &  & 68.19 \\
\textsc{FmAD-Vanilla} &  & 75.09 \\
\midrule
\textsc{BP-Checkpointing} & \multirow{3}{*}{64} & 88.59 \\
\textsc{ZO-Vanilla} &  & 65.94 \\
\textsc{FmAD-Vanilla} &  & \texttt{jvp} = \texttt{NaN} \\
\midrule
\textsc{BP-Checkpointing} & \multirow{3}{*}{128} & 86.47 \\
\textsc{ZO-Vanilla} &  & 64.12 \\
\textsc{FmAD-Vanilla} &  & \texttt{jvp} = \texttt{NaN} \\
\midrule
\textsc{BP-Checkpointing} & \multirow{3}{*}{Full fine-tuning} & 96.68 \\
\textsc{ZO-Vanilla} &  & 32.37 \\
\textsc{FmAD-Vanilla} &  & \texttt{jvp} = \texttt{NaN}\\
\bottomrule
\end{tabular}
\end{table}

\begin{figure}
    \centering
    \includegraphics[width=0.45\linewidth]{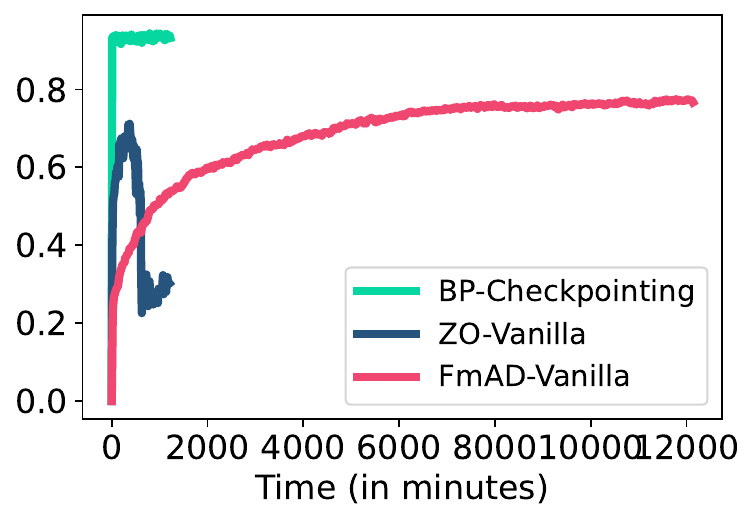}
    \caption{Comparison of convergence time among \textsc{BP-Checkpointing}, \textsc{FmAD-Vanilla}, and \textsc{ZO-Vanilla} with OPT(13B) on AGNews dataset.}
    \label{fig:opt-convergence-time}
\end{figure}

We investigate how increasing the number of trainable parameters affects performance under BP, ZO and \fmad. 
Tables~\ref{tbl:opt-accuracy-size-variants} and~\ref{tbl:opt-accuracy-lora-variants} present results across varying model sizes and \textsc{LoRA} ranks, respectively.
Further comparison of convergence time is available in Figure~\ref{fig:opt-convergence-time}.

In Table~\ref{tbl:opt-accuracy-size-variants}, we evaluate \textsc{BP-Checkpointing}, \textsc{ZO-Vanilla}, and \textsc{FmAD-Vanilla} on OPT model variants of size 1.3B, 6.7B, and 13B. 
As the model size increases, \textsc{BP-Checkpointing} consistently maintains high accuracy of $\sim$94\%. 
In contrast, ZO and \fmad exhibit noticeable drops in accuracy at larger model scales. 
Notably, \fmad achieves 88.28\% accuracy on the 1.3B model but declines to 77.07\% on the 13B model, showing degradation from scaling the count of trainable parameters.
This result are consistent with our theoretical findings of convergence error bounded by the trainable parameter count (\S~\ref{sec:theoretical-analysis}). 
Table~\ref{tbl:opt-accuracy-lora-variants} explores accuracy as a function of \textsc{LoRA} rank for OPT 6.7B. 
While \textsc{BP-Checkpointing} degrades gracefully as rank increases (likely due to overfitting), \fmad becomes unstable and fails to converge beyond rank 1, yielding \texttt{NaN} outputs for higher ranks. 
\fmad's instability at higher \textsc{LoRA} ranks is due to inherent instability of perturbation-based gradient estimations, which we discuss in Appendix~\ref{adx:failure-mode-analysis}. 
ZO, while stable, shows limited improvement with increased rank, reaching only 68.97\% accuracy at rank 32.

For the main text, we chose $r=1$ as a diagnostic tool because it is the most favorable regime for \fmad/ZO. 
These methods suffer from a curse of dimensionality where gradient variance and compute costs scale with the number of trainable parameters ($d$). 
By showing that \textsc{BP-Checkpointing} is superior even at $r=1$, we demonstrate that \fmad/ZO fail even in their most ideal setting. 
If BP-free methods cannot compete in low-rank PEFT (where $d$ is small), they are mathematically precluded from being competitive in full fine-tuning. Furthermore, the original LoRA paper~\cite{hu2022lora} (Table 6) also shows that gains often saturate beyond rank$=1$ on GPT3 for WikiSQL and MNLI datasets. 

Nonetheless, we perform a larger LoRA sweep and full finetuning on AGNews with OPT-13B (Table~\ref{tab:higher-rank-lora}). 
We observe reduction of accuracy from increasing rank.
While ZO sees 39\% accuracy drop from $r=1$ to full finetuning setting, while \fmad becomes unstable (\texttt{NaN}s) at higher ranks, further supporting our claim.

\subsection{Failure Mode Analysis}
\label{adx:failure-mode-analysis}
In order to understand why variance reduction methods or adaptive optimizers sometimes fail to make \fmad and ZO converge, or converge at a suboptimal accuracy; we present failure mode analysis with different optimizers and SVRG.

\subsubsection{Challenges with Optimizer Choice}

Here we discuss a distinct failure mode of \fmad which has been frequently observed in our preliminary experiments:
the computed Jacobian-vector products (\jvp) abruptly spike in magnitude. 
These sudden surges lead to disproportionately large gradient updates, destabilizing training and hindering convergence. 
A similar failure mode has been observed in zero-order (ZO) methods, where the projected gradients, mathematically equivalent to \fmad's \jvp values, exhibit comparable instability.
\begin{figure}[ht]
  \centering
  \begin{minipage}[b]{0.35\textwidth}
    \centering
    \includegraphics[width=\linewidth]{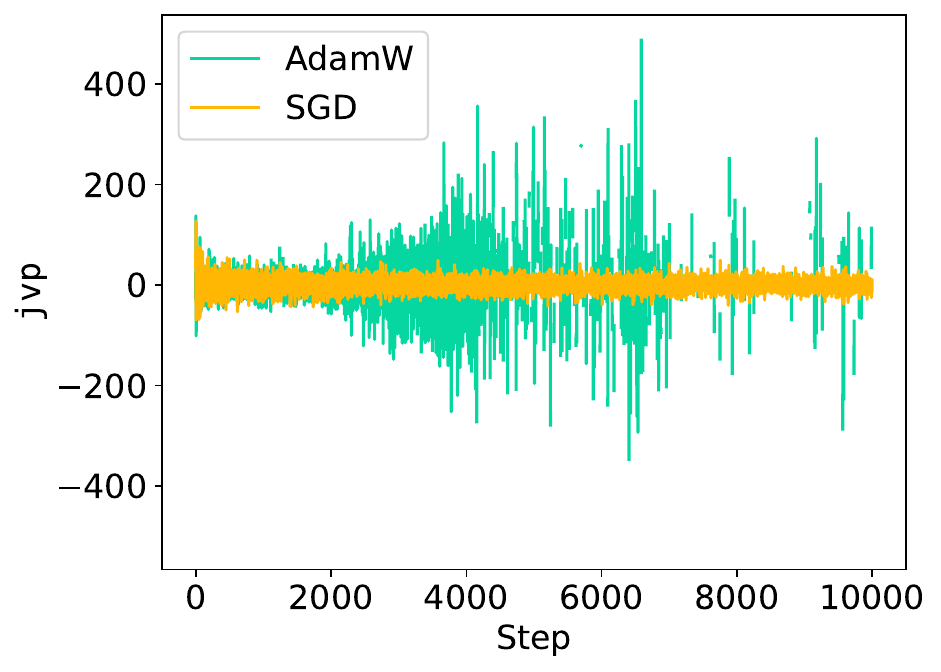}
    \caption{Effect of \textsc{AdamW} and \textsc{SGD} optimizers on \jvp values in \fmad on GSM8K dataset.}
    \label{fig:unstable-jvp-optimizer}
  \end{minipage}
  \hfill
  \begin{minipage}[b]{0.62\textwidth}
    \centering
    \begin{subfigure}[b]{0.49\textwidth}
         \centering
         \includegraphics[width=\textwidth]{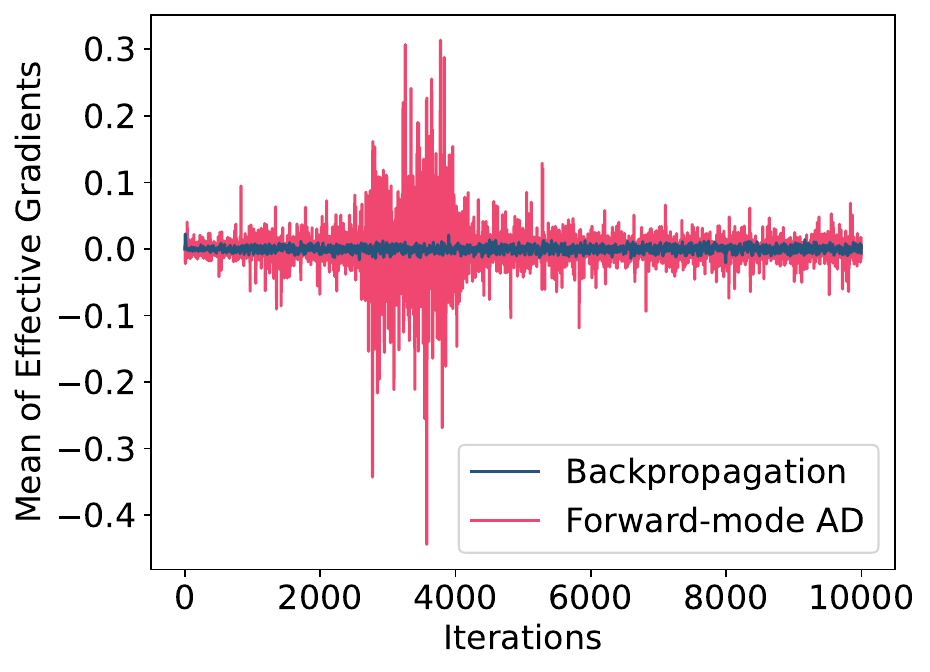}
         \caption{\textsc{AdamW} optimizer.}
         \label{fig:stable-gradients}
     \end{subfigure}
     \hfill
    \begin{subfigure}[b]{0.49\textwidth}
         \centering
         \includegraphics[width=\textwidth]{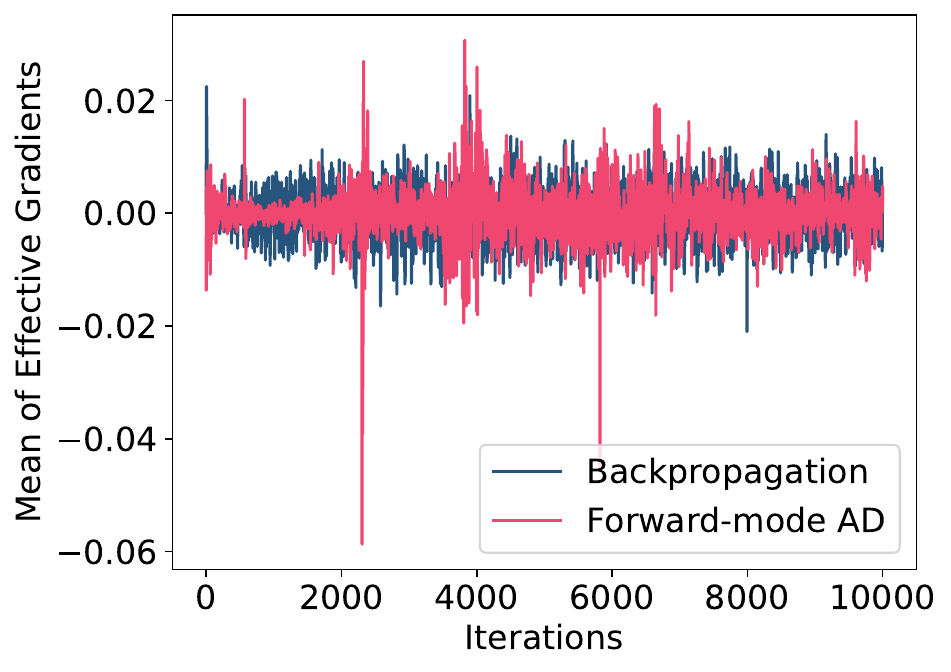}
         \caption{\textsc{SGD} optimizer.}
         \label{fig:unstable-gradients}
     \end{subfigure}
    \caption{Mean of effective gradients of Backpropagation and Forward-mode AD with \textsc{AdamW} and SGD optimizers on GSM8K dataset.}
    \label{fig:effective-gradients}
  \end{minipage}
\end{figure}

Figure~\ref{fig:unstable-jvp-optimizer} illustrates the impact of optimizer choice, specifically \textsc{AdamW} (adaptive) versus \textsc{SGD} (non-adaptive), on \jvp values in \fmad. 
Under \textsc{SGD}, \jvp values remain bounded within a stable range of $[-50, 50]$ for the case of GSM8K dataset. 
However, with \textsc{AdamW}, these values exhibit a gradual increase followed by sharp spikes for certain datasets including GSM8K.
In some cases, the spikes reach 8–10$\times$ higher magnitudes than the stable baseline observed with \textsc{SGD}.

Figures~\ref{fig:stable-gradients} and~\ref{fig:unstable-gradients} further illustrate the implications of these spikes.
Under \textsc{AdamW}, the effective gradient magnitudes produced by \fmad exhibit substantially higher variance than those from backpropagation, indicating instability and less reliable gradient directions. 
These inflated updates also increase weight magnitudes, which in turn amplify subsequent \jvp evaluations, since these depend on both the current weights and their perturbations.
This positive feedback loop can lead to divergence and, eventually, \texttt{NaN} values in \jvp computations, as observed in several \fmad runs in Table~\ref{tbl:opt-accuracy-lora-variants}. 
Even when divergence does not occur, the resulting gradient updates can be excessively noisy or of high magnitude, leading to suboptimal convergence.
In contrast, under \textsc{SGD}, the effective gradients computed by \fmad closely mirror those from backpropagation across most iterations, with stable behavior and no evidence of runaway magnitudes.
We posit that this cascading rise in magnitude for the case of \textsc{AdamW} is due to its adaptive nature, where a rolling average of historical and current gradients is computed each iteration, leading to amplification of higher magnitude gradients.
In contrast, the impact of \jvp spikes is diminished with non-adaptive SGD since the previous iteration's gradients would have limited effect (to only one iteration's gradient updates). 

This stark contrast highlights a critical interaction between optimizer choice and the numerical stability of \fmad. 
While \textsc{AdamW} is widely favored for its adaptive learning rates and regularization capabilities, its use with \fmad (and by extension ZO) can introduce harmful gradient artifacts which are driven by uncontrolled \jvp amplification.
These results underscore the specific vulnerabilities in gradient estimation methods and point to a need for further study into stabilizing \fmad and ZO for more reliable deployment in large-scale training regimes.

\subsubsection{Challenges with SVRG}
\label{adx:challenges-with-svrg}
In this section, we discuss a failure mode of SVRG observed in the context of text generation tasks. 
While SVRG improves performance for both \textsc{ZO-Vanilla} and \textsc{FmAD-Vanilla} baselines by 4.04--11.13 and 1.97--3.86, respectively, in many settings, it leads to performance degradation in certain sequence modeling tasks like GSM8K.
\begin{figure}
     \centering
     \begin{subfigure}[b]{0.49\textwidth}
         \centering
         \includegraphics[width=\textwidth]{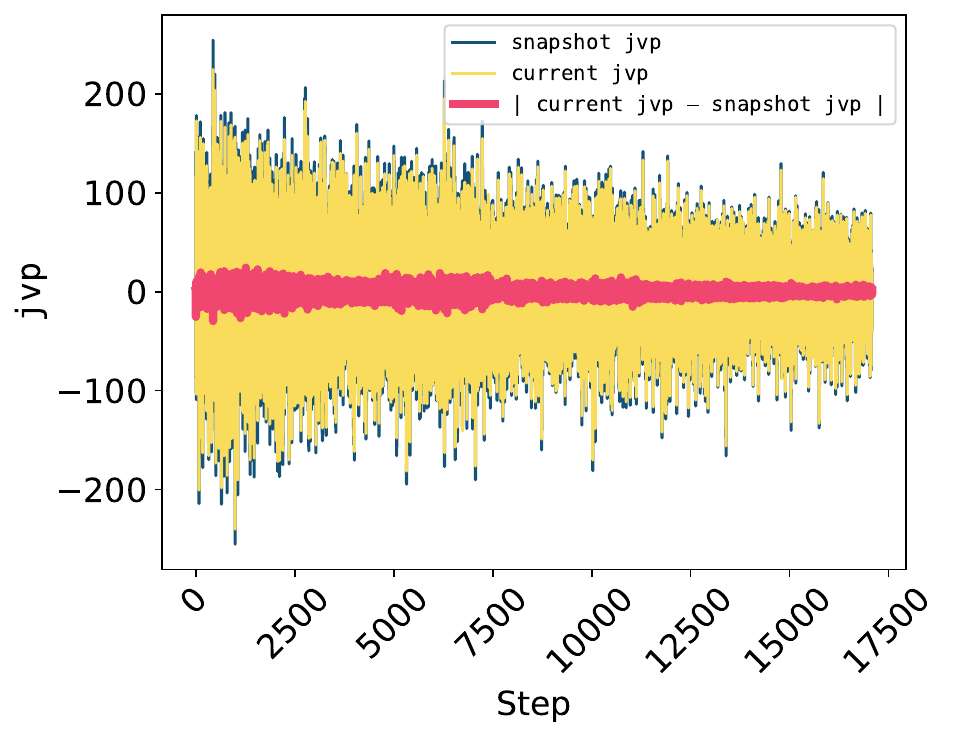}
         \caption{The current iteration's \jvp, \jvp computed based on the snapshot weights, and the difference between these two values.}
         \label{fig:svrg-jvp}
     \end{subfigure}
     \hfill
     \begin{subfigure}[b]{0.49\textwidth}
         \centering
         \includegraphics[width=\textwidth]{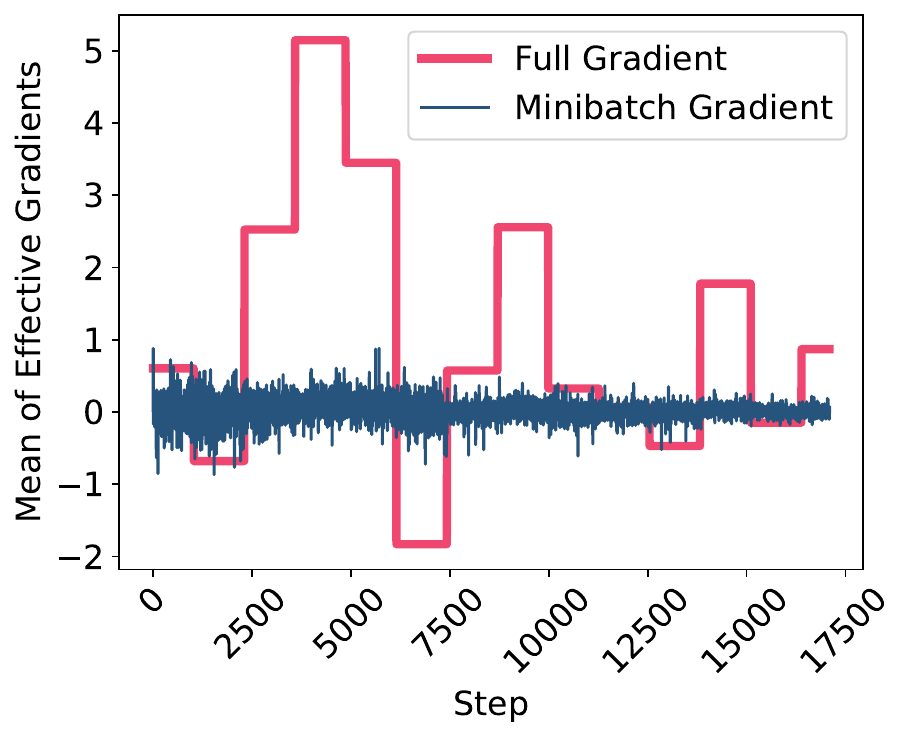}
         \caption{Magnitude of the mean effective full gradients (computed at specific epoch intervals) and the current iteration's mini-batch gradient.}
         \label{fig:svrg-gradients}
     \end{subfigure}
        \caption{Impact of incorporating SVRG into \fmad on (a) \jvp values and the mean of effective full-batch and (b) mini-batch gradients, evaluated on the GSM8K dataset.}
        \label{fig:svrg-jvp-gradients}
\end{figure}
Figure~\ref{fig:svrg-jvp-gradients} illustrates the behavior of \jvp values and the corresponding gradients when SVRG is applied to \fmad on the GSM8K dataset. 
In Figure~\ref{fig:svrg-jvp}, we observe that the difference between the \jvp computed on the current model weights and the one computed on the snapshot weights is minimal.
Consequently, the control variate, the difference between mini-batch gradients at current and snapshot weights, has little impact relative to the magnitude of the full gradient.

This hypothesis is supported by Figure~\ref{fig:svrg-gradients}, which shows that the mean of the effective full gradients remains consistently large, while the mini-batch gradient magnitudes are significantly smaller. 
Because the full gradients are updated only at periodic intervals (every 5 epochs in our case), their inflated magnitude dominates the update direction across multiple steps.
This inflation stems from the accumulation of large \jvp values during the summation of per-batch gradients, occasionally resulting in outlier gradients with extremely high norms.
As a result, the SVRG mechanism fails to provide meaningful variance reduction and instead perpetuates overly large updates, ultimately degrading model performance.

A similar performance degradation was observed in \textsc{ZO-SVRG}~\citep{liu2018zosvrg}, albeit on a smaller model with approximately 852K parameters. 
However, that work does not address the scalability challenges of SVRG-based methods in the context of zeroth-order optimization.

\subsubsection{{Improving Stability via Multiple-Perturbation and Accumulated-Gradient}}
\label{adx:analysis-on-accumulate-multiple}

We further extend our analysis of \texttt{jvp} magnitudes and mean gradient values to the variance-reducing baselines \textsc{-Multiple} (which samples multiple perturbations per iteration and averages the resulting gradients) and \textsc{-Accumulate} (which uses a single perturbation per iteration but accumulates gradients over several steps before applying an update). 

\begin{figure}
    \begin{subfigure}[b]{0.49\textwidth}
         \centering
         \includegraphics[width=\textwidth]{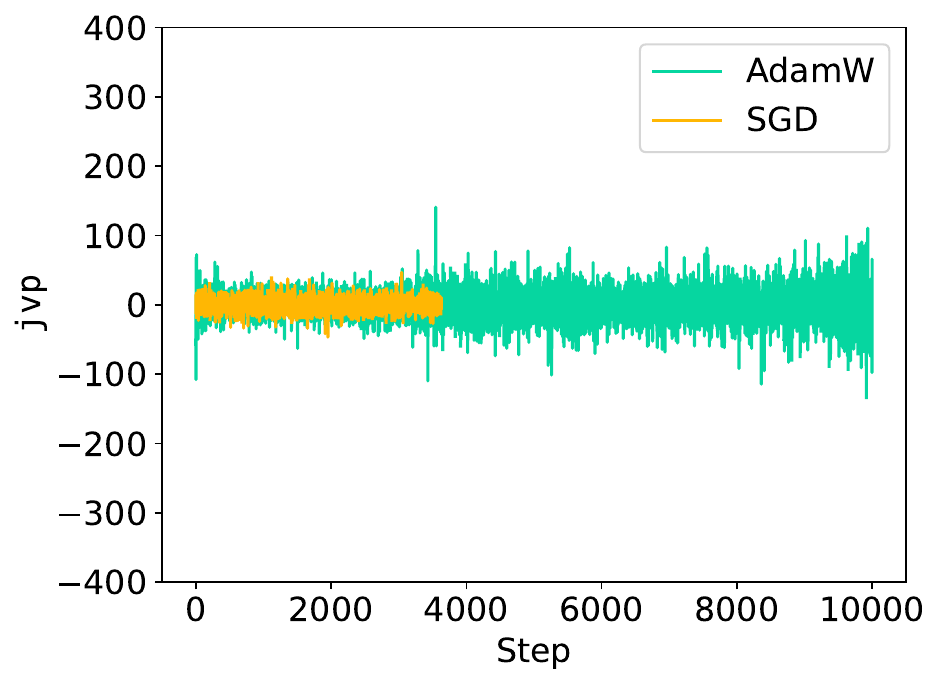}
         \caption{\textsc{-Multiple} with $n=10$.}
         \label{fig:multiple-stable-jvp}
     \end{subfigure}
     \hfill
    \begin{subfigure}[b]{0.49\textwidth}
         \centering
         \includegraphics[width=\textwidth]{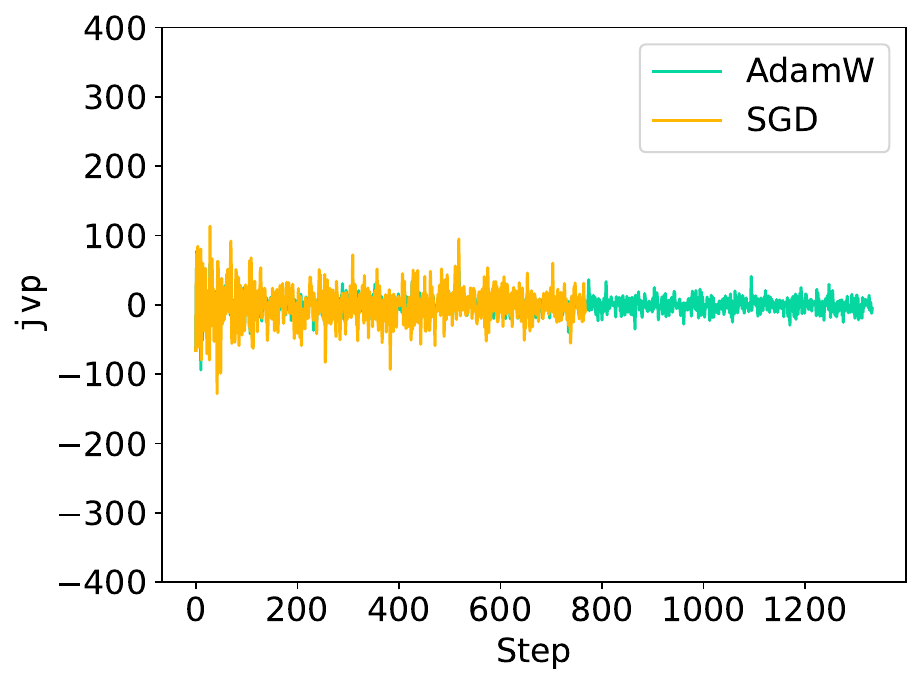}
         \caption{\textsc{-Accumulate} with Step Count=100.}
         \label{fig:accumulate-unstable-jvp}
     \end{subfigure}
    \caption{Effect of \textsc{AdamW} and ~\textsc{SGD} optimizers on \texttt{jvp} values in \textsc{FmAD-Multiple} and \textsc{FmAD-Accumulate} on \textsc{GSM8K} dataset.}
    \label{fig:multiple-accumulate-unstable-jvp}
\end{figure}

Figure~\ref{fig:multiple-accumulate-unstable-jvp} reports the corresponding \texttt{jvp} trajectories.
Note that the \textsc{SGD} baseline contains fewer plotted steps because it converged substantially earlier than the other configurations, and the experiment was therefore terminated once convergence was reached.
In contrast to the instability observed for \textsc{FmAD-Vanilla} in Figure~\ref{fig:unstable-jvp-optimizer}, both baselines exhibit stable \texttt{jvp} magnitudes even under \textsc{AdamW}. 
This stability, in turn, yields lower error and more reliable convergence.
This behavior can be attributed to the inherent variance-reduction mechanisms in these baselines. 
In \textsc{-Multiple}, averaging \emph{multiple} jvp-induced gradient estimates suppresses the high-variance noise that otherwise interacts negatively with \textsc{AdamW}'s adaptive accumulators.
Similarly, \textsc{-Accumulate} delays updates and aggregates gradient signals across several steps, effectively smoothing out perturbation-induced fluctuations before the optimizer sees them. 
In both cases, the optimizer receives a more stable and lower-variance gradient stream, preventing the cascading amplification effects that cause jvp spikes in \textsc{FmAD-Vanilla}.
As a result, these variance-reduction strategies mitigate the optimizer–noise interaction responsible for divergence, leading to substantially more stable training dynamics.

\begin{figure}
    \begin{subfigure}[b]{0.49\textwidth}
         \centering
         \includegraphics[width=\textwidth]{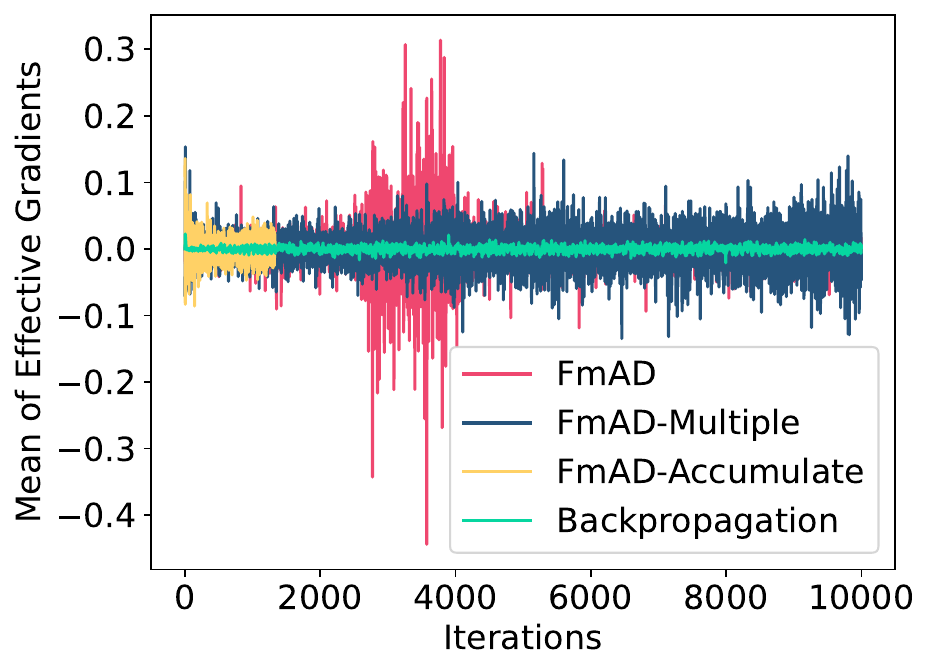}
         \caption{\textsc{AdamW} optimizer.}
         \label{fig:adamw-stable-gradients}
     \end{subfigure}
     \hfill
    \begin{subfigure}[b]{0.49\textwidth}
         \centering
         \includegraphics[width=\textwidth]{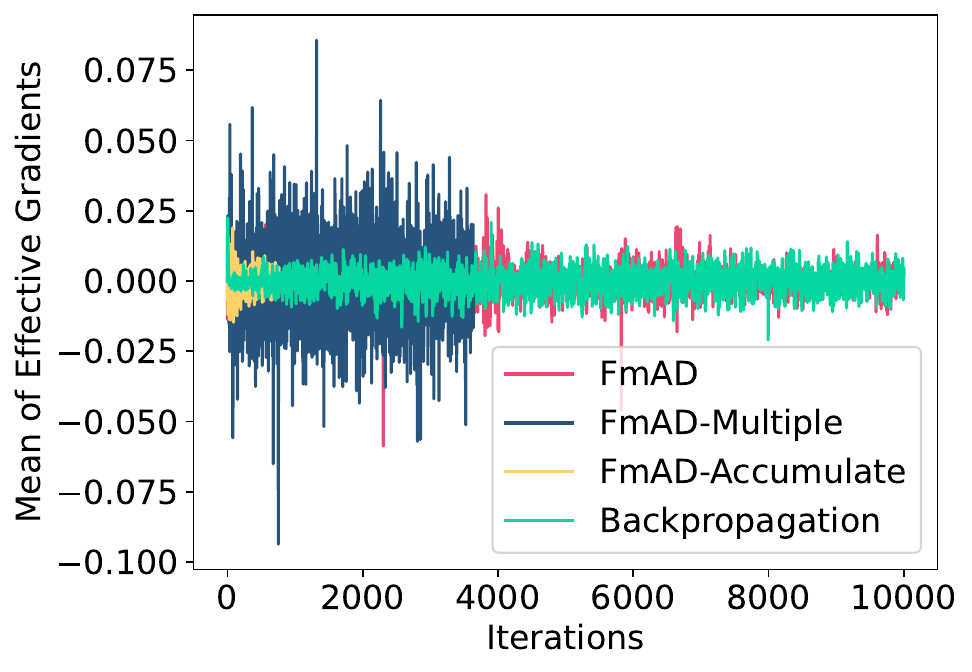}
         \caption{\textsc{SGD} optimizer.}
         \label{fig:sgd-unstable-gradients}
     \end{subfigure}
    \caption{Mean of effective gradients of Backpropagation, \textsc{FmAD}, \textsc{FmAD-Multiple} ($n=10$), and \textsc{FmAD-Accumulate} (Step Count=100) with \textsc{AdamW} and \textsc{SGD} optimizers on \textsc{GSM8K} dataset.}
    \label{fig:multiple-accumulate-unstable-gradients}
\end{figure}

Figure~\ref{fig:multiple-accumulate-unstable-gradients} reports the mean gradient magnitudes. 
The curves for Backpropagation and \textsc{FmAD} are identical to those shown previously in Figure~\ref{fig:effective-gradients}. 
In addition, we include the results for \textsc{FmAD-Multiple} and \textsc{FmAD-Accumulate}.
Unlike the pronounced gradient-magnitude spikes observed in \textsc{FmAD}, both \textsc{-Multiple} and \textsc{-Accumulate} exhibit markedly steadier behavior under both optimizers \textsc{AdamW} and \textsc{SGD}. 
Notably, \textsc{Accumulate} displays the greatest stability. 
This is expected: accumulating gradients over several iterations before applying an update effectively averages out the perturbation-induced noise and prevents high-variance signals from being directly fed into the optimizer's adaptive state. 
As a result, \textsc{AdamW} receives smoother, lower-variance updates, which suppresses the positive feedback loop responsible for the divergence in \textsc{FmAD}.
In contrast, \textsc{Multiple} exhibits a slight upward drift near the end of training when used with \textsc{AdamW}. 
This behavior is consistent with the fact that, although multiple perturbations are averaged per iteration, the optimizer still processes an update at every step; thus, residual noise (especially as weights grow in magnitude) can accumulate in the adaptive moments and produce a mild increase in gradient scale. 
Nevertheless, this increase remains small relative to the uncontrolled spikes in \textsc{FmAD-Vanilla}, confirming that perturbation-level averaging substantially reduces variance.
Finally, note that \textsc{Accumulate} has fewer points plotted because it performs fewer parameter-update steps; gradients are accumulated locally and applied only periodically, resulting in a lower number of optimizer interactions reflected in the visualization. 

\subsection{{Effect of Perturbation Distributions and Normalization Strategies}}
\label{adx:distribution-normalization-strategies}

We additionally experimented with  perturbation sampling strategies:
(a)~Sampling from a normal distribution and using the perturbations as-is (unnormalized),
(b)~Sampling from a normal distribution and normalizing the perturbations,
(c)~Sampling from a uniform distribution and using the perturbations as-is (unnormalized), and
(d)~Sampling from a uniform distribution and normalizing the perturbations.

Our findings are as follows.
Normalization consistently reduces accuracy for both the normal and uniform variants.
\begin{wrapfigure}{r}{0.4\textwidth}
    \centering
    \includegraphics[width=\linewidth]{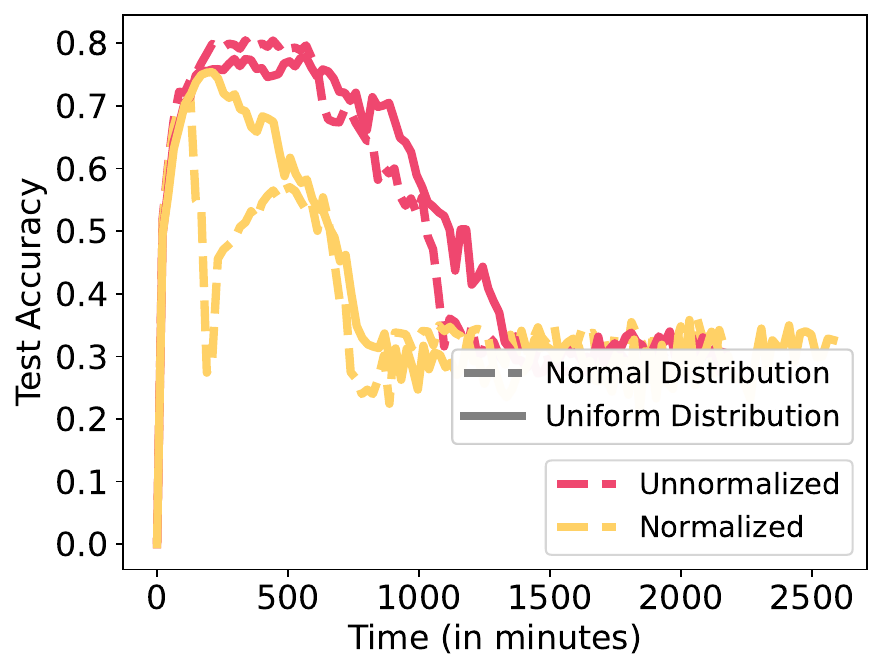}
    \caption{Finetuning \textsc{Llama 3.1} (8B) on the AGNews dataset using \fmad, comparing perturbations drawn from normal vs. uniform distributions, with both normalized and unnormalized variants.}
  \end{wrapfigure}
This degradation arises because normalization forces every perturbation to have identical magnitude, eliminating natural variability in scale that carries useful information for estimating the local curvature of the loss landscape.
By projecting all perturbations onto a fixed-radius hypersphere, the method reduces the effective signal-to-noise ratio of the \texttt{jvp} estimate and prevents larger, informative perturbations (particularly in high-curvature regions) from contributing to learning.
As a result, the gradients become less expressive and exhibit higher relative variance, leading to poorer optimization.

Among the unnormalized variants, sampling from a normal distribution yields the strongest performance, with the unnormalized uniform distribution performing comparably closely, before both resulting in overfitting.
The slight advantage of the normal distribution can be attributed to its heavier tails, which naturally introduce a broader range of perturbation magnitudes.
This diversity more closely mimics the statistical structure of true gradients in large neural networks, allowing the estimator to explore directions of both small and moderately large curvature.
In contrast, the unnormalized uniform distribution produces perturbations bounded within a fixed interval, limiting the range of effective step sizes and resulting in marginally less efficient gradient estimation.

\subsection{{Comparison against SignZO}}
\label{adx:comparison-against-signzo}

Table~\ref{tab:signzo} shows a comparison of accuracy, memory usage, compute cost, and convergence time for \textsc{BP-Checkpointing}, ZO, \textsc{SignZO}, \textsc{ZO-Accumulate}, and \textsc{ZO-Multiple} when finetuning \textsc{Llama-3.1} (8B) on AGNews.

\begin{table}[]
\centering
\caption{Performance, memory, and efficiency trade-offs across \textsc{BP-Checkpointing}, ZO baselines, and \textsc{SignZO} for finetuning \textsc{Llama-3.1} (8B) on AGNews.}
\begin{tabular}{lrcrr}
\toprule
 & {Accuracy} & {\begin{tabular}[c]{@{}c@{}}Memory \\ Consumption \\ (in GB)\end{tabular}} & {\begin{tabular}[c]{@{}l@{}}Compute Cost \\ (in FLOPs)\end{tabular}} & {\begin{tabular}[c]{@{}l@{}}Wallclock Convergence \\ Time (in seconds)\end{tabular}} \\
 \midrule
\textsc{BP-Checkpointing} & 93.8\% & 11.66 & 65.2 x $10^4$ & 16,691 \\
ZO & 73.6\% & \phantom{0}5.99 & 251.2 x $10^4$ & 21,074 \\
\textsc{SignZO} & 82.6\% & \phantom{0}5.99 & 251.9 x $10^4$ & 56,892 \\
\textsc{ZO-Accumulate} & 85.8\% & \phantom{0}5.99 & 2165.1 x $10^4$ & 181,510 \\
\textsc{ZO-Multiple} & 86.7\% & \phantom{0}5.99 & 2425 x $10^4$ & 201,747 \\
\bottomrule
\end{tabular}
\label{tab:signzo}
\end{table}

\begin{wrapfigure}{r}{0.4\textwidth}
  \begin{center}
    \includegraphics[width=0.4\textwidth]{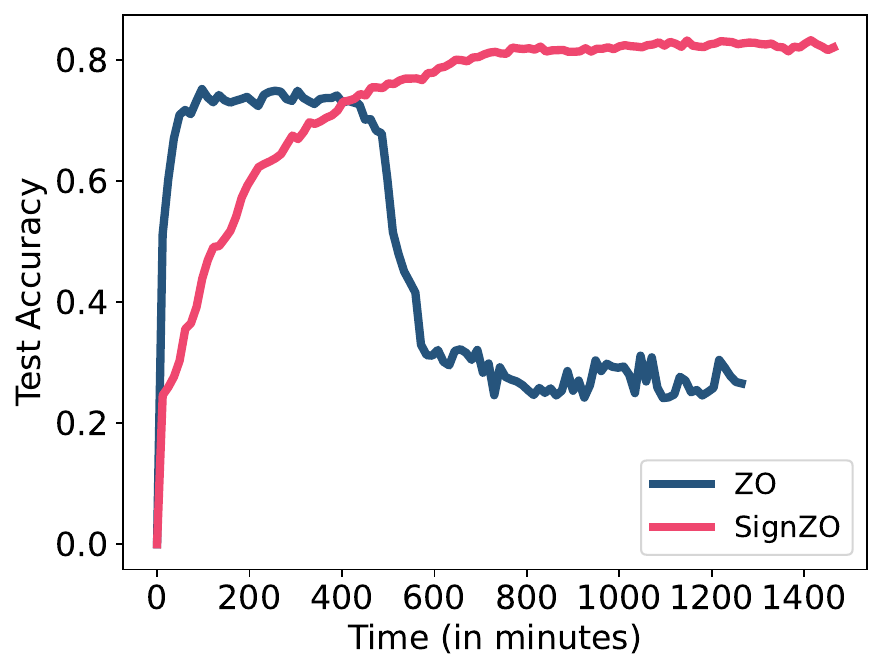}
  \end{center}
  \caption{SignZO against ZO for training \textsc{Llama 3.1} (8B) on AGNews dataset.}
  \label{fig:signzo}
\end{wrapfigure}

\textbf{Accuracy Comparison.} 
Although \textsc{SignZO} improves stability relative to vanilla ZO (as reflected in its smoother learning trajectory in the Figure~\ref{fig:signzo}) its overall accuracy performance remains significantly below the backpropagation baseline. 
On AGNews with \textsc{Llama3.1} (8B), \textsc{SignZO} reaches 82.6\% accuracy, which is a noticeable improvement over the 73.6\% achieved by standard ZO but still far from the 93.8\% obtained via backpropagation.
This gap indicates that the sign-based estimator, while stabilizing the update direction, does not provide sufficient gradient resolution to match the fidelity of true gradients.
\textsc{SignZO} also underperforms compared to the variance-reducing methods (\textsc{ZO-Multiple} and \textsc{ZO-Accumulate}). \textsc{ZO-Multiple} and \textsc{ZO-Accumulate} reach 86–87\% accuracy, and although they require larger FLOPs and longer runtimes, they converge to higher-quality solutions. 

\textbf{Memory, Computation cost, and Convergence Time Comparison.} 
\textsc{SignZO} matches the memory footprint of other ZO baselines (5.99 GB) and maintains similar FLOP-level compute costs. 
However, its wall-clock convergence time is substantially longer ($\approx$56.9k seconds), more than 2.7$\times$ slower than ZO and 3.4$\times$ slower than BP-Checkpointing. 
The longer convergence time stems from the fact that stabilizing noisy ZO directions via sign compression requires more optimization steps to make meaningful progress.
Although Table~\ref{tab:signzo} reports only wall-clock time, \textsc{SignZO} and ZO have identical per-iteration runtime, the only difference between them is that \textsc{SignZO} applies a sign-compressed update during \texttt{optimizer.step()}, which does not affect iteration cost. 
Consequently, time on the x-axis is effectively proportional to the number of optimization steps, allowing us to conclude that the longer wall-clock time directly reflects the larger number of iterations required for convergence.

Overall, \textsc{SignZO} improves upon naïve ZO in terms of final accuracy (82.6\% vs. 73.6\%), but does so by requiring substantially more computation: although its per-iteration FLOPs are nearly identical to ZO, its wall-clock convergence time is 2.7$\times$ longer (56.9k s vs. 21.1k s). Compared to BP-Checkpointing, \textsc{SignZO} achieves a markedly smaller memory footprint (5.99 GB vs. 11.66 GB), but only by trading off both efficiency and performance, requiring $\approx$3.4$\times$ longer time to converge, $\approx$3.9$\times$ more compute, and yielding 11.2 percentage points lower accuracy. 
Furthermore, while \textsc{SignZO} converges faster than the variance-reduced ZO-Accumulate and ZO-Multiple baselines, those methods achieve higher accuracies (85.8\% and 86.7\%), reinforcing the broader trend observed in our paper: 
stability alone is not sufficient, effective ZO training at scale also requires variance-reduction mechanisms to improve both accuracy and efficiency.

\subsection{{Sensitivity to Perturbation Budget for OPT13B}}
\label{adx:perturbation-budget-sensitivity}

\begin{wrapfigure}{r}{0.4\textwidth}
\vspace{-0.6cm}
  \begin{center}
    \includegraphics[width=0.4\textwidth]{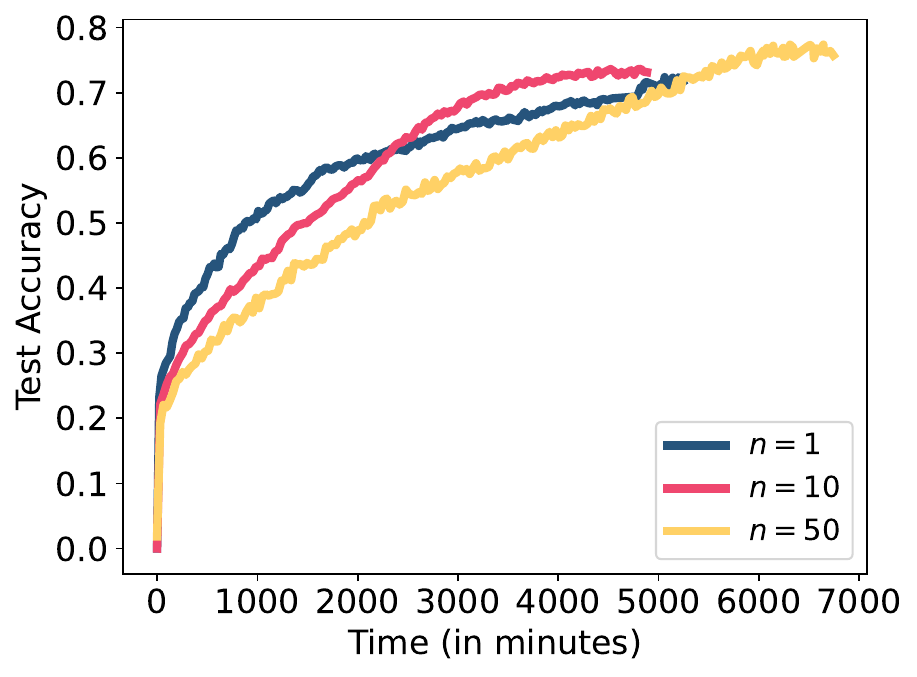}
  \end{center}
  \caption{Training the OPT (13B) model on the AGNews dataset using ZO with varying perturbation budgets $n$. 
  Larger budgets reduce variance in the \texttt{jvp} estimates, improving final accuracy, but sequential application of perturbations under resource constraints increases wallclock convergence time.}
  \label{fig:opt-13b-perturbations}
  \vspace{-0.7cm}
\end{wrapfigure}

Figure~\ref{fig:opt-13b-perturbations} shows the training of the OPT (13B) model on the AGNews dataset using ZO with varying perturbation budgets $n$.
For $n=1$, convergence occurs around 5000 minutes, reaching an accuracy of 71\%. 
With $n=10$, convergence also occurs near 5000 minutes, with a slight improvement in accuracy to 72\%. 
Increasing the budget further to $n=50$ improves the final accuracy to 75.5\%, but convergence is delayed until approximately 6500 minutes.
This behavior can be explained by the trade-off between gradient estimate quality and computational overhead: 
larger perturbation budgets reduce the variance of the \texttt{jvp} estimates, leading to more accurate gradients and higher final accuracy, but in this case, the perturbations are applied sequentially due to hardware limitations, which increases wall-clock time per iteration. Additionally, small increases in $n$ (e.g., from 1 to 10) yield only modest accuracy gains because even a few perturbations are sufficient to capture enough directional information for effective early-stage training. 
Overall, these results highlight a trade-off between perturbation budget, convergence speed, and final accuracy: larger budgets improve the quality of gradient estimates at the cost of increased computation time, particularly when sequential execution is required.

\section{Signal Propagation for Gradient Computation}
\label{adx:signal-propagation}
A key difference between BP and \fmad /ZO methods lies in how they propagate the loss signal to compute weight updates. 
BP computes the derivative of the loss $\cL$ with respect to each weight $w_i$, effectively mapping changes in the loss to precise updates in the parameter space. 
The gradients of the intermediate activations, computed during the backward pass, are also derived from $\delta \cL$, allowing the loss signal to guide every stage of the update. 
This direct path from the loss to the parameters makes BP a \textbf{loss-to-weights} approach, where the signal flows backward through the network in a structured and deterministic way.

In contrast, both \fmad and ZO adopt a \textbf{weights-to-loss} perspective: 
they estimate how perturbations in the weights, $\delta \bw = \bv$; affect the loss, $\delta \cL$. 
The forward-mode Jacobian-vector product (\jvp) and the ZO projected gradient scalar both incorporate the resulting change in the loss, but they do so indirectly. 
Specifically, they multiply $\delta \cL$ by the perturbation direction $\bv$ to approximate weight gradients 
(as detailed in \S~\ref{sec:background}). 
However, in these approaches, the intermediate changes $\delta y_p$ (which influence $\delta \cL$) are driven by the initial perturbations $\delta w_p \sim \mathcal{N}(0, I_d)$; not by the loss. 
As a result, the variance introduced at the input level through the perturbations propagates forward through the network, ultimately contaminating the gradient signal.
This lack of an explicit loss-driven mechanism for shaping activation gradients leads to noisier gradient updates.
Consequently, \fmad and ZO require stricter step size constraints (see Theorems~\ref{thm:convergence-zero-order} and~\ref{thm:convergence-forward-mode-ad}) and exhibit degraded convergence behavior. 

Moreover, both \fmad and ZO optimization methods incur additional noise and estimation error compared to backpropagation.
This noise is not just a side effect, it is an inherent consequence of using random perturbations to estimate gradients. 
In both \fmad and ZO, the injection of perturbations $\delta \bw \sim \mathcal{N}(0, I_d)$ is core to the algorithmic process, and the resulting activation ($\delta y_i$) and loss variations ($\delta \cL$) carry this randomness forward. 
Therefore, the gradient estimates vary depending on the sampled perturbation, making noise a deterministic outcome of the method itself.

In essence, while BP precisely channels loss information to guide weight updates, \fmad and ZO rely on stochastic approximations that make their updates fundamentally noisy and less targeted.

\section{Computational Complexity}
\label{adx:computational-complexity}
In this section, we analyze the computational complexity of different methods used to compute gradients in a neural network setting. 
We begin with a one-layer neural network, providing a detailed breakdown of the computational cost for the forward pass, backpropagation, zero-order optimization, and forward-mode automatic differentiation.
Understanding these complexities is essential for evaluating the efficiency of gradient computation methods, especially in resource-constrained environments.
Empirical computational cost of the gradient computation methods is shown in \S~\ref{subsec:computation-cost}. 

\subsection{Basics}  
In this section, we analyze the computational complexity of a one-layer neural network $f$ with weight matrix $ w \in \bR^{d \times m}$. 
The network takes an input $ x \in \bR^d$ and produces an output $y \in \bR^m$. 
While we focus on a single-layer setting for clarity, the analysis naturally extends to a deep neural network with $L$ layers, each with weight matrix $w_\ell$ for $\ell \in [L]$.  

\paragraph{Forward Pass.} 
Since all three gradient computation methods share the same forward pass, we first establish its computational complexity. 
The forward pass consists of a matrix multiplication $y = xw$, where $x$ has dimensions $1 \times d$ and $w $ has dimensions $d \times m$. 
This results in a computational complexity of $\cO(dm)$.  

\subsection{Backpropagation}  
Backpropagation requires computing the gradient of the loss $\cL$ with respect to the weights, given by  
$$
\frac{\partial \cL}{\partial w} = \frac{\partial \cL}{\partial y} \cdot \frac{\partial y}{\partial w}.
$$  
The first term, $\frac{\partial \cL}{\partial y}$, involves differentiating the loss with respect to the output, which has a computational complexity of $\cO(m)$. 
The second term, $\frac{\partial y}{\partial w}$, follows from the linear transformation $y = xw$, contributing a complexity of $\cO(dm)$. 
The final gradient computation involves the multiplication of a \( 1 \times m \) matrix with an \( m \times d \) matrix, resulting in an additional complexity of $\cO(dm)$.  

Although activation functions introduce constant factors, 3 for the last layer and 5 for intermediate layers, these constants do not affect the asymptotic complexity. Hence, the overall computational complexity of backpropagation remains \( \cO(dm) \).

\subsection{Backpropagation with Checkpointing}
Checkpointing builds on standard backpropagation by trading memory for additional computation. Instead of storing all intermediate activations, only selected layers are checkpointed, and discarded activations are recomputed as needed during the backward pass.  

This recomputation introduces an overhead, resulting in a total compute complexity of $\mathcal{O}(dm \log p)$ for a network with $p$ layers~\cite{griewank2000algo799}.  
Here, the $\log p$ factor reflects the optimal checkpointing schedule, capturing the additional cost of recomputing intermediate activations while still reducing peak memory usage compared to standard backpropagation.  
In this way, checkpointing offers a controlled trade-off between memory efficiency and computational overhead, extending the base $\mathcal{O}(dm)$ cost of standard backpropagation.

\subsection{Zero-Order Optimization}
The zero-order optimization method with central finite differences involves perturbing the weights twice, evaluating at $ (\bw + \epsilon \bv) \text{ and } (\bw - \epsilon \bv), $ 
where \( \bv \in \bR^{d \times m} \) is a randomly sampled perturbation and \( \epsilon \in \bR \) is a small step size. 
The element-wise multiplication \( \epsilon \bv \) incurs a computational cost of \( \cO(dm) \), as does the addition and subtraction with \( w \). 
Since each perturbation requires evaluating the function at the perturbed points, the function evaluations \( f(\bw \pm \epsilon \bv) \) also contribute a complexity of \( \cO(dm) \).  

With \( n \) such perturbations per iteration, the total computational cost sums to \( \cO(ndm) \), where \( n \) is the number of perturbations used in each iteration.  

Compared to the forward pass on the original weights $w$, zero-order adds a constant of 4, which gets absorbed in $\cO(ndm)$.

\subsection{Forward-Mode AD}  
The \jvp (Jacobian-vector product) computation incurs a complexity of $\mathcal{O}(dm)$, as it partially computes $\frac{\partial \mathcal{L}}{\partial w}$. 
The resulting \jvp is then multiplied with the perturbation vector $\bv$ to obtain the weight gradient for $\bw$. 
Since $\bv$ has dimensions $d \times m$, this multiplication also has a computational complexity of $\mathcal{O}(dm)$.  

Repeating this process $n$ times for $n$ perturbations per iteration leads to a total computational cost of $\mathcal{O}(ndm)$.

\section{Proofs of Convergence Bounds}
\label{adx:proofs-convergence}
This section includes the details on upper error bounds of all three gradient computation methods: 
Backpropagation,
Zero-order optimization,
and Forward-mode Auto Differentiation.

\subsection{Basics}
All examples of gradient computation methods are based on a function $f$, which, in the context of machine learning, corresponds to a neural network. 
This function $f$ is composed of nested functions $f_i$, $i\in [p]$; where each function corresponds to an intermediate output (or activation) $y_i = f_i(w_i, y_{i-1})$, generated from the input weights $w_i$ and previous activation $y_{i-1}$. 
$y_0$ is set to $x$, which can be data points in ML. 
We assume that $x$ is fixed, for the ease of exposition.
The input weights are represented by the vector $\bw = {w_1, w_2, \dots, w_p}$, where each $w_{[1, \dots, p]} \in \bR^{[m_1, \dots, m_p]}$. 
The intermediate outputs, or activations, are denoted by $\by = {y_1, \dots, y_p}$.
The final output is $y = y_p = f(\bw, x) \in \bR^n$, where typically $n << m_i$ for all $i \in [p]$. 
The loss function $\cL(y,\hat{y}) \in \bR$ is then computed to measure the difference between the predicted output $y$ and the true target values $\hat{y}$.

With gradient descent, one update to the weights $\bw$ looks like this,
\begin{align}
    \bw_{t+1} \leftarrow \bw_t  - \eta \nabla f (\bw_t), 
    \label{eq:model-update-rule}
\end{align} 
% \kunjal{in the above equation, is it $\nabla f(\bw)$ or $\bE_{(x, y \sim \cD)}[\nabla f(\bw, (x,y))]$?}
where $t$ is the iteration count, and $\eta$ is the learning rate.

The objective is to minimize $f(\bw)$: $\min_{\bw \in \bR^d} f(\bw)$.

\begin{definition}[Optimality Gap]
The optimality gap at iteration $t$ is defined as the difference between the function value at the current iterate $\bw_t$ and the function value at an optimal solution $\bw^*$:
\begin{align}
    \Delta_t = f(\bw_t) - f(\bw^*)
\end{align}
The optimality gap quantifies how far the current function value is from the optimal value. 
In convergence analysis, the goal is to show that this gap decreases over iterations.
\label{eq:optimality-gap}
\end{definition}

\subsection{Assumptions}

\begin{assumption}[Smoothness]
Let $f:\bR^d \rightarrow \bR$ be $L$-smooth, meaning that its gradient is Lipschitz continuous with constant $L>0$. 
That is, for all $\bw, \bw' \in \bR^d$, the function $f$ satisfies,
\begin{align}
    || \nabla f(\bw') - \nabla f(\bw) || \leq L || 
    \bw' - \bw ||.
\end{align}
\label{asm:smoothness}
\end{assumption}

\subsection{Lemmas}
\begin{lemma}[Bias of Gradient Estimate of the Central Finite Difference]
    Let $g(\bv)$ be the gradient estimate obtained using the central finite difference method with a perturbation vector $\bv \sim \mathcal{N}(0, I_d)$. Then, the expectation of the estimator satisfies
    \begin{align}
        \mathbb{E}_{\bv} [g(\bv)] = \nabla f(\bw),
    \end{align}
    implying that the central finite difference gradient estimator is unbiased up to first-order error terms.
    Furthermore, the second moment of the estimator satisfies
    \begin{align}
        \mathbb{E}_{\bv} \left[ \| g(\bv) \|^2 \right] = \| \nabla f(\bw) \|^2 (d + 2) + \mathcal{O}(\epsilon^2) d.
    \end{align}
    \label{lma:bias-central-finite-differences}
\end{lemma}
\begin{proof}
    We start with one evaluation of the central finite difference, for a perturbation $\bv \in \mathbb{R}^d$. The full derivation is given in Theorem~\ref{thm:convergence-zero-order},
    \begin{align}
        g(\bv) = \left( (\bv^\top \nabla f(\bw))\bv + \mathcal{O}(\epsilon)\bv \right).
    \end{align}
    In order to measure the bias of the above gradient estimate, we take expectation with respect to the randomness of $\bv$,
    \begin{align}
        \mathbb{E}_{\bv} \left[ g(\bv) \right] &= \mathbb{E}_{\bv} \left[ (\bv^\top \nabla f(\bw))\bv + \mathcal{O}(\epsilon)\bv \right] \\
        &= \nabla f(\bw) \mathbb{E}_{\bv} \left[ \bv^\top \bv \right] + \mathcal{O}(\epsilon) \mathbb{E}_{\bv} \left[\bv \right] \\
        &= \nabla f(\bw) I_d + \mathcal{O}(\epsilon) \cdot 0 \quad \text{(since $\bv \sim \mathcal{N}(0, I_d)$)} \\
        \therefore \mathbb{E}_{\bv} \left[ g(\bv) \right] &= \nabla f(\bw).
    \end{align}
    This shows that the estimator is unbiased.
    
    Now, we analyze the second moment of the estimator:
    \begin{align}
        \mathbb{E}_{\bv} \left[ \| g(\bv) \|^2 \right] &= \mathbb{E}_{\bv} \left[ \left\| (\bv^\top \nabla f(\bw))\bv + \mathcal{O}(\epsilon)\bv \right\|^2 \right] \\
        &= \mathbb{E}_{\bv} \left[ (\bv^\top \nabla f(\bw))^2 \| \bv \|^2 \right] + \mathcal{O}(\epsilon^2) \mathbb{E}_{\bv} \left[ \| \bv \|^2 \right].
    \end{align}
    Using the known expectation property of Gaussian vectors:
    \begin{align}
        \mathbb{E} \left[ \bv \bv^\top \| \bv \|^2 \right] = (d+2) I_d,
    \end{align}
    we obtain:
    \begin{align}
        \mathbb{E}_{\bv} \left[ \| g(\bv) \|^2 \right] &= \mathbb{E}_{\bv} \left[ \text{Tr}((\bv \bv^\top) \nabla f(\bw) \nabla f(\bw)^\top \bv \bv^\top) \right] + \mathcal{O}(\epsilon^2) d \\
        &= \text{Tr}(\nabla f(\bw) \nabla f(\bw)^\top) (d + 2) + \mathcal{O}(\epsilon^2) d \\
        &= \| \nabla f(\bw) \|^2 (d + 2) + \mathcal{O}(\epsilon^2) d. \label{eq:second-moment-of-central-finite-differences-gradient-estimator}
    \end{align}
\end{proof}

\begin{lemma}[Variance of Gradient Estimate of the Central Finite Difference]
\label{lma:var-central-finite-differences}
Let $\hat{g}(\bv)$ be the central finite difference gradient estimator using $n$ perturbations $\bv_1, \dots, \bv_n \sim \mathcal{N}(0, I_d)$, given by
\begin{align}
    \hat{g}(\bw) = \frac{1}{n} \sum_{i=1}^n g(\bv_i), \quad \text{where} \quad g(\bv_i) = \frac{f(\bw + \epsilon \bv_i) - f(\bw - \epsilon \bv_i)}{2\epsilon} \bv_i.
\end{align}
Assuming the finite-difference step $\epsilon$, the variance of the estimator satisfies:
\begin{align}
    \text{Var}[\hat{g}(\bw)] = \frac{1}{n} \left( \| \nabla f(\bw) \|^2 (d + 1) + \mathcal{O}(\epsilon^2) d \right).
\end{align}
This result shows that the variance of the gradient estimator scales as $\mathcal{O}((d+1)/n)$, which quantifies how the dimension $d$ and the number of samples $n$ influence the estimator's variance.
\end{lemma}
\begin{proof}
    We can derive the variance of the estimator
    \begin{align}
        \hat{g}(\bw) = \frac{1}{n} \sum_{i=1}^n g(\bv_i)\text{, with } g(\bv_i) = \frac{f(\bw + \epsilon \bv_i) - f(\bw - \epsilon \bv_i)}{2\epsilon} \bv_i \nonumber
    \end{align}
    by computing
    \begin{align}
        \text{Var}[\hat{g}(\bw)] = \bE_{\bv}[|| \hat{g}(\bw) ||^2] - || \bE_{\bv}[\hat{g}(\bw)] ||^2. \label{eq:variance-g-hat}
    \end{align}
    For clarity, in the following we assume that the finite‐difference step is chosen so that the bias is negligible (i.e. the estimator is unbiased according to Lemma~\ref{lma:bias-central-finite-differences}, so that $\bE_{\bv}[g(\bv_i)]=\nabla f(\bw)$ and $\bE_{\bv}[\hat{g}(\bw)] = \nabla f(\bw)$. (We can add the higher-order remainder later.) 
    
    We write the second moment (squared norm) of $\hat{g}$ as
    \begin{align}
        \bE_{\bv}[||\hat{g}(\bw)||^2] &= \bE_{\bv}\left[\left|\left| \frac{1}{n} \sum_{i=1}^n g(\bv_i) \right|\right|^2\right] = \frac{1}{n^2} \bE_{\bv} \left[ \sum_{i=1}^n \sum_{j=1}^n g(\bv_i)^\top g(\bv_j) \right].
    \end{align}
    We split the sum into diagonal and off-diagonal parts:
    \begin{align}
        \bE_{\bv}[||\hat{g}(\bw)||^2] &= \frac{1}{n^2} \left( \sum_{i=1}^n  \bE_{\bv}[||g(\bv_i)||^2] + \sum_{i \neq j} \bE [g(\bv_i)^\top g(\bv_j)] \right).
    \end{align}
    Since the $g_i$ are independent,
    \begin{align}
        \bE_{\bv} [g(\bv_i)^\top g(\bv_j)] &= \bE_{\bv}[g(\bv_i)]^\top \bE_{\bv}[g(\bv_j)] = \nabla f(\bw)^\top \nabla f(\bw) = || \nabla f(\bw) ||^2 &\text{for }i\neq j.
    \end{align}
    Thus, 
    \begin{align}
        \bE_{\bv}[||\hat{g}(\bw)||^2] &= \frac{1}{n^2} (n \bE_{\bv} [|| g(\bv) ||^2 ] + n(n-1) || \nabla f(\bw) ||^2).
    \end{align}
    Plugging the above result, along with the result derived from Lemma~\ref{lma:bias-central-finite-differences} on $|| \bE_{\bv} [\hat{g}(\bw)] ||^2 = || \nabla f(\bw) ||^2$, in Equation~\ref{eq:variance-g-hat},
    \begin{align}
        \text{Var}[\hat{g}(\bw)] &= \frac{1}{n^2} \left( n \bE_{\bv} [|| g(\bv) ||^2 ] + n(n-1) || \nabla f(\bw) ||^2 \right) - || \nabla f(\bw) ||^2\\
        &= \frac{1}{n} \left( \bE_{\bv} [|| g(\bv) ||^2 ]  -  || \nabla f(\bw) ||^2\right).
    \end{align}
    The second moment of the estimator was derived in Lemma~\ref{lma:bias-central-finite-differences}, in Equation~\ref{eq:second-moment-of-central-finite-differences-gradient-estimator}. 
    We use that result in the above equation as follows,
    \begin{align}
        \text{Var}[\hat{g}(\bw)] &= \frac{1}{n} \left( \| \nabla f(\bw) \|^2 (d + 2) + \mathcal{O}(\epsilon^2) d - \| \nabla f(\bw) \|^2\right) \\
        &= \frac{1}{n} \left( \| \nabla f(\bw) \|^2 (d + 1) + \mathcal{O}(\epsilon^2) d\right)
    \end{align}
    This leads us to a variance bound that scales as $\frac{d+1}{n}$ times $||\nabla f(\theta)||^2$ (plus a $\cO(\epsilon^2)$ contribution), which exhibits the dependence of variance of the estimator $\hat{g}(\bw)$ on the dimension $d$ and the number of samples $n$.
\end{proof}

\begin{tcolorbox}
The key difference between the above two lemmas and the next two lemmas is that the central finite difference estimator introduces a small $O(\epsilon)$ bias due to numerical approximation, whereas the forward-mode AD estimator is exactly unbiased. 
Additionally, the second moment of the central finite difference estimator includes an extra $O(\epsilon^2)d$ term, which is absent in forward-mode AD, making the latter more precise.
\end{tcolorbox}

\begin{lemma}[Bias of Gradient Estimate of Forward-mode Auto Differentiation]
    Let $g(\bv)$ be the gradient estimate obtained using the central finite difference method with a perturbation vector $\bv \sim \mathcal{N}(0, I_d)$. 
    Then, the expectation of the estimator satisfies
    \begin{align}
        \mathbb{E}_{\bv} [g(\bv)] = \nabla f(\bw),
    \end{align}
    implying that the central finite difference gradient estimator is unbiased.
    Furthermore, the second moment of the estimator satisfies
    \begin{align}
        \mathbb{E}_{\bv} \left[ \| g(\bv) \|^2 \right] = \| \nabla f(\bw) \|^2 (d + 2).
    \end{align}
    \label{lma:bias-forward-mode-ad}
\end{lemma}
\begin{proof}
    We start with one evaluation of forward-mode auto differentiation, for a perturbation $\bv \in \mathbb{R}^d$. 
    The full derivation is given in Theorem~\ref{thm:convergence-forward-mode-ad},
    \begin{align}
        g(\bv) = \left( \bv^\top \nabla f(\bw) \right)\bv.
    \end{align}
    In order to measure the bias of the above gradient estimate, we take expectation with respect to the randomness of $\bv$,
    \begin{align}
        \mathbb{E}_{\bv} \left[ g(\bv) \right] &= \mathbb{E}_{\bv} \left[ (\bv^\top \nabla f(\bw))\bv \right] \\
        &= \nabla f(\bw) \mathbb{E}_{\bv} \left[ \bv^\top \bv \right] = \nabla f(\bw) I_d  \\
        \therefore \mathbb{E}_{\bv} \left[ g(\bv) \right] &= \nabla f(\bw).
    \end{align}
    This shows that the estimator is unbiased.
    
    Now, we analyze the second moment of the estimator:
    \begin{align}
        \mathbb{E}_{\bv} \left[ \| g(\bv) \|^2 \right] &= \mathbb{E}_{\bv} \left[ \left\| (\bv^\top \nabla f(\bw))\bv \right\|^2 \right] = \mathbb{E}_{\bv} \left[ (\bv^\top \nabla f(\bw))^2 \| \bv \|^2 \right].
    \end{align}
    Using the known expectation property of Gaussian vectors:
    \begin{align}
        \mathbb{E} \left[ \bv \bv^\top \| \bv \|^2 \right] = (d+2) I_d,
    \end{align}
    We obtain:
    \begin{align}
        \mathbb{E}_{\bv} \left[ \| g(\bv) \|^2 \right] &= \mathbb{E}_{\bv} \left[ \text{Tr}((\bv \bv^\top) \nabla f(\bw) \nabla f(\bw)^\top \bv \bv^\top) \right] \\
        &= \text{Tr}(\nabla f(\bw) \nabla f(\bw)^\top) (d + 2) = \| \nabla f(\bw) \|^2 (d + 2). \label{eq:second-moment-of-forward-mode-ad-gradient-estimator}
    \end{align}
\end{proof}

\begin{lemma}[Variance of Gradient Estimate of Forward-mode Auto Differentiation]
\label{lma:var-forward-mode-auto-differentiation}
Let $\hat{g}(\bv)$ be the central finite difference gradient estimator using $n$ perturbations $\bv_1, \dots, \bv_n \sim \mathcal{N}(0, I_d)$, given by
\begin{align}
    \hat{g}(\bw) = \frac{1}{n} \sum_{i=1}^n g(\bv_i), \quad \text{where} \quad g(\bv_i) = (\bv_i^\top \nabla f(\bw)) \bv_i.
\end{align}
Assuming the finite-difference step $\epsilon$, the variance of the estimator satisfies:
\begin{align}
    \text{Var}[\hat{g}(\bw)] = \frac{1}{n} \left( \| \nabla f(\bw) \|^2 (d + 1)\right).
\end{align}
This result shows that the variance of the gradient estimator scales as $\mathcal{O}((d+1)/n)$, which quantifies how the dimension $d$ and the number of samples $n$ influence the estimator's variance.
\end{lemma}
\begin{proof}
    We can derive the variance of the estimator
    \begin{align}
        \hat{g}(\bw) = \frac{1}{n} \sum_{i=1}^n g(\bv_i)\text{, with } g(\bv_i) = (\bv_i^\top \nabla f(\bw)) \bv_i \nonumber
    \end{align}
    by computing
    \begin{align}
        \text{Var}[\hat{g}(\bw)] = \bE_{\bv}[|| \hat{g}(\bw) ||^2] - || \bE_{\bv}[\hat{g}(\bw)] ||^2. \label{eq:variance-g-hat-fmad}
    \end{align}
    The estimator is unbiased according to Lemma~\ref{lma:bias-forward-mode-ad}, so that $\bE_{\bv}[g(\bv_i)]=\nabla f(\bw)$ and $\bE_{\bv}[\hat{g}(\bw)] = \nabla f(\bw)$.  
    
    We write the second moment (squared norm) of $\hat{g}$ as
    \begin{align}
        \bE_{\bv}[||\hat{g}(\bw)||^2] &= \bE_{\bv}\left[\left|\left| \frac{1}{n} \sum_{i=1}^n g(\bv_i) \right|\right|^2\right] = \frac{1}{n^2} \bE_{\bv} \left[ \sum_{i=1}^n \sum_{j=1}^n g(\bv_i)^\top g(\bv_j) \right].
    \end{align}
    We split the sum into diagonal and off-diagonal parts:
    \begin{align}
        \bE_{\bv}[||\hat{g}(\bw)||^2] &= \frac{1}{n^2} \left( \sum_{i=1}^n  \bE_{\bv}[||g(\bv_i)||^2] + \sum_{i \neq j} \bE [g(\bv_i)^\top g(\bv_j)] \right).
    \end{align}
    Since the $g_i$ are independent,
    \begin{align}
        \bE_{\bv} [g(\bv_i)^\top g(\bv_j)] &= \bE_{\bv}[g(\bv_i)]^\top \bE_{\bv}[g(\bv_j)] = \nabla f(\bw)^\top \nabla f(\bw) = || \nabla f(\bw) ||^2 &\text{for }i\neq j.
    \end{align}
    Thus, 
    \begin{align}
        \bE_{\bv}[||\hat{g}(\bw)||^2] &= \frac{1}{n^2} (n \bE_{\bv} [|| g(\bv) ||^2 ] + n(n-1) || \nabla f(\bw) ||^2).
    \end{align}
    Plugging the above result, along with the result derived from Lemma~\ref{lma:bias-central-finite-differences} on $|| \bE_{\bv} [\hat{g}(\bw)] ||^2 = || \nabla f(\bw) ||^2$, in Equation~\ref{eq:variance-g-hat-fmad},
    \begin{align}
        \text{Var}[\hat{g}(\bw)] &= \frac{1}{n^2} \left( n \bE_{\bv} [|| g(\bv) ||^2 ] + n(n-1) || \nabla f(\bw) ||^2 \right) - || \nabla f(\bw) ||^2\\
        &= \frac{1}{n} \left( \bE_{\bv} [|| g(\bv) ||^2 ]  -  || \nabla f(\bw) ||^2\right).
    \end{align}
    The second moment of the estimator was derived in Lemma~\ref{lma:bias-forward-mode-ad}, in Equation~\ref{eq:second-moment-of-forward-mode-ad-gradient-estimator}. 
    We use that result in the above equation as follows,
    \begin{align}
        \text{Var}[\hat{g}(\bw)] &= \frac{1}{n} \left( \| \nabla f(\bw) \|^2 (d + 2) - \| \nabla f(\bw) \|^2\right) \\
        &= \frac{1}{n} \left( \| \nabla f(\bw) \|^2 (d + 1)\right)
    \end{align}
    This leads us to a variance bound that scales as $\frac{d+1}{n}$ times $||\nabla f(\theta)||^2$, which exhibits the dependence of variance of the estimator $\hat{g}(\bw)$ on the dimension $d$ and the number of samples $n$.
\end{proof}

\subsection{Theorems}
The given analysis for all gradient computation methods is for \textbf{a non-convex objective} $f$.\\
We begin by reiterating the descent lemma applied to gradients computed by backpropagation. 
\begin{theorem}[Error Bound of Backpropagation]
    Let $f: \bR^d \rightarrow \bR$ be a differentiable, $L$-smooth
    % , and convex 
    function. 
    Consider the gradient descent update rule:
    $$\bw_{t+1} = \bw_t - \eta \nabla f(\bw_t),$$
    where $\eta$ is the step size (learning rate).
    Suppose $0 < \eta \leq \frac{1}{L}$. 
    Then, after $T$ iterations, the minimum gradient norm satisfies the following bound:
    $$\min_{t \in [T]} || \nabla f(\bw_t) ||^2 \leq \frac{2L}{T} \left( f(\bw_1) - f(\bw_T) \right),$$
    % where $f(\bw^*)$ is the optimal function value.
    This bound shows that gradient descent achieves an $\cO(\frac{1}{T})$ convergence rate in terms of gradient norm, which is the optimal rate for first-order methods in smooth 
    % convex 
    optimization. 
    \label{thm:convergence-backpropagation}
\end{theorem}
\begin{proof}
    Using Assumption~\ref{asm:smoothness}, 
    % s~\ref{asm:convexity} and
    we apply the smoothness condition, which gives the following quadratic upper bound:
    \begin{align}
        f(\bw_{t+1}) \leq f(\bw_{t}) + \nabla f(\bw_{t})^\top (\bw_{t+1} - \bw_{t}) + \frac{L}{2} || \bw_{t+1} - \bw_{t} ||^2
    \end{align}
    Substituting the gradient descent update rule Equation~\ref{eq:model-update-rule}, we obtain,
    \begin{align}
        \therefore f(\bw_{t+1}) \leq f(\bw_{t}) - \eta || \nabla f(\bw_{t}) ||^2 + \frac{L}{2} \eta^2 || \nabla f (\bw_{t}) ||^2
    \end{align}
    Rearranging the above terms,
    \begin{align}
        \therefore f(\bw_{t+1}) - f(\bw_{t}) \leq - \eta || \nabla f(\bw_{t}) ||^2 + \frac{L\eta^2}{2} || \nabla f (\bw_{t}) ||^2 = - \left( \eta - \frac{L \eta^2}{2} \right) || \nabla f(\bw_t) ||^2
    \end{align}
    To ensure progress in minimizing $f(\bw)$, we need the term $(1 -\frac{L\eta}{2})$ to be positive. 
    Hence we assume $\eta \leq \frac{1}{L}$, along with $0 < \eta$,
    \begin{align}
        \therefore f(\bw_{t+1}) - f(\bw_{t}) \leq - \frac{1}{2L} || \nabla f(\bw_t) ||^2
    \end{align}
    Summing over $t=1$ to $t=T$,
    \begin{align}
        \sum_{t=1}^T \left( f(\bw_{t+1}) - f(\bw_{t}) \right) \leq - \frac{1}{2L} \sum_{t=1}^T || \nabla f(\bw_t) ||^2
    \end{align}
    The left-hand side forms a telescoping sum, resulting in 
    \begin{align}
        f(\bw_{T+1}) - f(\bw_1) \leq -\frac{1}{2L} \sum_{t=1}^T || \nabla f(\bw_t) ||^2 \\
        \frac{1}{T}\sum_{t=1}^T || \nabla f(\bw_t) ||^2 \leq \frac{2L}{T} \left(f(\bw_1) - f(\bw_{T+1})\right)
    \end{align}
    Using the definition of optimality gap from Equation~\ref{eq:optimality-gap},
    \begin{align}
        \min_{t \in [T]} || \nabla f(\bw_t) ||^2 \leq \frac{2L}{T} \left( f(\bw_t) - f(\bw_T) \right)
    \end{align}
    Thus, the optimality gap reduces at a rate of $\cO(\frac{1}{T})$, given $\eta \leq \frac{1}{L}$.
\end{proof}

Next, we will give a similar treatment to the gradients derived from zero-order finite differences,

\begin{theorem}[Error Bound of Zero-Order Optimization]
    \label{thm:convergence-zero-order}
    Consider a 
    % convex 
    function $f:\bR^d\rightarrow \bR$ that is $L$-smooth. 
    Let the central finite-difference gradient estimator with $n$ perturbations per iteration, where each perturbation vector $\bv_i$ is sampled independently from $\cN(0, I_d)$ and step size $\eta$ be
    $$\hat{g}(\bw_t) = \frac{1}{n} \sum_{i=1}^n \left( \frac{f(\bw_t + \epsilon \bv_i) - f(\bw_t - \epsilon \bv_i)}{2\epsilon} \bv_i \right).$$ 
    Then, the expected average squared gradient norm is bounded by
    \begin{align} 
    \frac{1}{T} \sum_{t=1}^T || \nabla f(\bw_t)||^2 &\leq \frac{f(\bw_1) - f(\bw_T)}{\eta T \left[1 - \frac{L \eta}{2} \left( 1 + \frac{d+1}{n}\right) \right]} + \frac{L d \eta^2}{2n} \mathcal{O}(\epsilon^2),
    \end{align} 
    % where $\bw^*$ is the optimal solution.
    To ensure convergence, the step size must satisfy
    \begin{align} \eta < \frac{2}{L\left( 1 + \frac{d+1}{n}\right)}. \end{align} 
    This result highlights how the convergence rate depends on the dimension $d$, the number of perturbations $n$, and the perturbation magnitude $\epsilon$.
    Specifically, a larger $d$ or a smaller $n$ increases the bound, implying slower convergence.
\end{theorem}
\begin{proof}
    The central finite-difference gradient estimator for $n$ perturbations per iteration is
    \begin{align}
        \hat{g}(\bw_t) = \frac{1}{n} \sum_{i=1}^n \left( \frac{f(\bw_t + \epsilon \bv_i) - f(\bw_t - \epsilon \bv_i)}{2\epsilon} \bv_i \right)  \label{eq:central-finite-difference}
    \end{align}
    where each $\bv_i \in \bR^d$ is a perturbation drawn from a Gaussian distribution $\cN (0, 1)$. 
    
    Assuming that $f$ is sufficiently smooth so that the following Taylor expansions are valid,
    \begin{align}
        f (\bw + \epsilon \bv) &= f(\bw) + \epsilon \bv^\top \nabla f(\bw) + O (\epsilon^2) \text{, and}\\
        f (\bw - \epsilon \bv) &= f(\bw) - \epsilon \bv^\top \nabla f(\bw) + O (\epsilon^2)
    \end{align}
    Subtracting the above two expansions yields:
    \begin{align}
        f (\bw + \epsilon \bv) - f (\bw - \epsilon \bv) &= 2 \epsilon \bv^\top \nabla f(\bw) + \cO (\epsilon^2) 
    \end{align}
    Plugging the above result in Equation~\ref{eq:central-finite-difference},
    \begin{align}
        \hat{g}(\bw_t) = \frac{1}{n} \sum_{i=1}^n \underbrace{\left( (\bv_i^\top \nabla f(\bw))\bv_i + \cO(\epsilon)\bv_i \right)}_{g(\bv_i)}
    \end{align}
    
    \emph{Now we will use the derived $\hat{g}(\bw_t)$ in the descent lemma:}
    
    Similar to Theorem~\ref{thm:convergence-backpropagation}, using the Assumption
    % s~\ref{asm:convexity} and
    ~\ref{asm:smoothness}, we apply the smoothness on $f$:
    \begin{align}
        f(\bw_{t+1}) \leq f(\bw_{t}) + \nabla f(\bw_{t})^\top (\bw_{t+1} - \bw_{t}) + \frac{L}{2} || \bw_{t+1} - \bw_{t} ||^2
    \end{align}
    For the gradient descent update with central finite differences, we set the model update rule as
    \begin{align}
        \bw_{t+1} = \bw_{t} - \eta \hat{g}(\bw_t).
    \end{align}
    Plugging the model update rule into the smoothness inequality,
    \begin{align}
        f(\bw_{t+1}) \leq f(\bw_t) - \eta \nabla f(\bw_t)^\top \hat{g}(\bw_t) + \frac{L \eta^2}{2} || \hat{g}(\bw_t) ||^2
    \end{align}
    Taking expectation conditioned on $\bv \sim \cN(0, I_d)$,
    \begin{align}
        f(\bw_{t+1}) \leq f(\bw_t) - \eta \nabla f(\bw_t)^\top \underbrace{\bE_{\bv}[\hat{g}(\bw_t)]}_{\text{Term}_1} + \frac{L \eta^2}{2} \underbrace{\bE_{\bv}[|| \hat{g}(\bw_t) ||^2]}_{\text{Term}_2} \label{eq:zo-one-step}
    \end{align}
    Solving \textbf{Term}$_1$ and \textbf{Term}$_2$ separately,
    \paragraph{Term$_1$: }
    From Lemma~\ref{lma:bias-central-finite-differences}, we get $\bE[g(\bv)] = \nabla f(\bw)$, which also gets us $$\bE [\hat{g}(\bw)] = \frac{1}{n}\sum_{i=1}^n \bE[g(\bv)] = \nabla f(\bw)$$
    \paragraph{Term$_2$: } The error of $\hat{g}(\bw)$ is measured by $\delta$,
    \begin{align}
        || \hat{g}(\bw) ||^2 &= || \nabla f(\bw) + \delta ||^2 = || \nabla f(\bw) ||^2 + 2 \nabla f(\bw)^\top \delta + || \delta ||^2
    \end{align}
    Taking expectation and noting that $\bE[\delta] = 0$ and $\bE [|| \delta ||^2] = \text{Var}[\hat{g}(\bw)]$,
    \begin{align}
        \bE_{\bv} [|| \hat{g}(\bw)||^2] = || \nabla f(\bw) ||^2 + \text{Var} [\hat{g}(\bw)]
    \end{align}
    Using Lemma~\ref{lma:var-central-finite-differences} to get the bound of $\text{Var} [\hat{g}(\bw)]$,
    \begin{align}
        \bE_{\bv} [|| \hat{g}(\bw)||^2] = || \nabla f(\bw) ||^2 + \frac{d+1}{n} || \nabla f(\bw) ||^2 + \frac{d}{n} \cO(\epsilon^2)
    \end{align}
    \paragraph{Back to Equation~\ref{eq:zo-one-step}, plugging in Term$_1$ and Term$_2$:}
    \begin{align}
        f(\bw_{t+1}) &\leq f(\bw_t) - \eta \nabla f(\bw_t)^\top \nabla f(\bw_t) + \frac{L \eta^2}{2} \left( \left( 1 + \frac{d+1}{n}\right) || \nabla f(\bw) ||^2 + \frac{d}{n} \cO(\epsilon^2) \right) \\
        &= f(\bw_t) - \eta || \nabla f(\bw_t)||^2 + \frac{L \eta^2}{2} \left( 1 + \frac{d+1}{n}\right) || \nabla f(\bw) ||^2 + \frac{L d \eta^2}{2n}\cO(\epsilon^2) 
    \end{align}
    Grouping the terms involving $|| \nabla f(\bw) ||^2$,
    \begin{align}
        f(\bw_{t+1}) &\leq f(\bw_t) - \eta \left[1 - \frac{L \eta}{2} \left( 1 + \frac{d+1}{n}\right) \right] || \nabla f(\bw_t)||^2 +  \frac{L d \eta^2}{2n}\cO(\epsilon^2)
    \end{align}
    This inequality shows that, provided the step size $\eta$ is small enough so that,
    $$1 - \frac{L \eta}{2} \left( 1 + \frac{d+1}{n}\right) > 0$$
    Summing the inequality over epochs $t = 1$ to $T$:
    \begin{align}
        f(\bw_{T}) - f(\bw_1) &\leq - \eta \left[1 - \frac{L \eta}{2} \left( 1 + \frac{d+1}{n}\right) \right] \sum_{t=1}^T || \nabla f(\bw_t)||^2 +  \frac{L d \eta^2 T}{2n}\cO(\epsilon^2)
    \end{align}
    Rearranging the terms 
    % and noting that $f(\bw_T) \geq f(\bw^*)$ 
    give us,
    \begin{align}
          \sum_{t=1}^T || \nabla f(\bw_t)||^2 &\leq \frac{f(\bw_1) - f(\bw_T)}{\eta \left[1 - \frac{L \eta}{2} \left( 1 + \frac{d+1}{n}\right) \right]} +  \frac{L d \eta^2 T}{2n}\cO(\epsilon^2)
    \end{align}
    Dividing by $T$ gives the bound on the average squared gradient norm:
    \begin{align}
          \frac{1}{T} \sum_{t=1}^T || \nabla f(\bw_t)||^2 &\leq \frac{f(\bw_1) - f(\bw_T)}{\eta T \left[1 - \frac{L \eta}{2} \left( 1 + \frac{d+1}{n}\right) \right]} +  \frac{L d \eta^2}{2n}\cO(\epsilon^2)
    \end{align}
    To ensure that 
    $1 - \frac{L \eta}{2} \left( 1 + \frac{d+1}{n}\right) > 0$, the step size $\eta$ must be chosen so that
    \begin{align}
        \eta < \frac{2}{L\left( 1 + \frac{d+1}{n}\right)}.
    \end{align}
    
    As the dimension $d$ increases (or as the number of samples $n$ decreases), the factor $$\frac{L\eta }{2}\left(1 + \frac{d+1}{n}\right)$$
    increases. This makes
    $$1 - \frac{L\eta }{2}\left(1 + \frac{d+1}{n}\right)$$
    smaller, which in turn makes the entire bound larger. 
    In other words, a larger $d$ (or a smaller $n$) results in a worse (higher) error bound.
    This interplay of $d$ and $n$ also puts limitations on the order of $\eta$, keeping the learning rate quite small for stable learning.
\end{proof}

Moving on, we will get the convergence bound of the gradients derived from forward-mode auto differentiation.
\begin{tcolorbox}
    The key difference between the two theorems is that the error bound for zero-order optimization includes an additional $\cO(\epsilon^2)$ term due to the finite-difference approximation, whereas the bound for forward-mode AD is exact and free from such errors. 
    This makes forward-mode AD theoretically more efficient, as it avoids the additional error introduced by numerical differentiation while maintaining the same dependency on dimension $d$ and number of perturbations $n$.
\end{tcolorbox}

\begin{theorem}[Error Bound of Forward-mode Auto Differentiation]
    \label{thm:convergence-forward-mode-ad}
    Consider a 
    % convex 
    function $f:\bR^d\rightarrow \bR$ that is $L$-smooth. 
    Let the forward-mode AD gradient estimator with $n$ perturbations per iteration, where each perturbation vector $\bv_i$ is sampled independently from $\cN(0, I_d)$ and step size $\eta$ be
    $$\hat{g}(\bw_t) = \frac{1}{n} \sum_{i=1}^n \left( \left(\bv_i^\top \nabla f(\bw_t) \right) \bv_i \right).$$ 
    Then, the expected average squared gradient norm is bounded by
    \begin{align} 
    \frac{1}{T} \sum_{t=1}^T || \nabla f(\bw_t)||^2 &\leq \frac{f(\bw_1) - f(\bw_T)}{\eta T \left[1 - \frac{L \eta}{2} \left( 1 + \frac{d+1}{n}\right) \right]},
    \end{align} 
    % where $\bw^*$ is the optimal solution.
    To ensure convergence, the step size must satisfy
    \begin{align} \eta < \frac{2}{L\left( 1 + \frac{d+1}{n}\right)}. \end{align} 
    This result highlights how the convergence rate depends on the dimension $d$, the number of perturbations $n$, and the perturbation magnitude $\epsilon$.
    Specifically, a larger $d$ or a smaller $n$ increases the bound, implying slower convergence.
\end{theorem}
\begin{proof}
    The forward-mode AD gradient estimator for $n$ perturbations per iteration is
    \begin{align}
        \hat{g}(\bw_t) = \frac{1}{n} \sum_{i=1}^n \left( (\bv_i^\top \nabla f(\bw_t)) \bv_i \right)  \label{eq:forward-mode-ad-gradient}
    \end{align}
    where each $\bv_i \in \bR^d$ is a perturbation drawn from a Gaussian distribution $\cN (0, 1)$. 
    
    We will use $\hat{g}(\bw_t)$ in the descent lemma. 
    Similar to Theorem~\ref{thm:convergence-zero-order}, using the Assumption~\ref{asm:smoothness}, we apply the smoothness on $f$:
    \begin{align}
        f(\bw_{t+1}) \leq f(\bw_{t}) + \nabla f(\bw_{t})^\top (\bw_{t+1} - \bw_{t}) + \frac{L}{2} || \bw_{t+1} - \bw_{t} ||^2
    \end{align}
    For the gradient descent update with forward-mode AD, we set the model update rule as
    \begin{align}
        \bw_{t+1} = \bw_{t} - \eta \hat{g}(\bw_t).
    \end{align}
    Plugging the model update rule into the smoothness inequality,
    \begin{align}
        f(\bw_{t+1}) \leq f(\bw_t) - \eta \nabla f(\bw_t)^\top \hat{g}(\bw_t) + \frac{L \eta^2}{2} || \hat{g}(\bw_t) ||^2
    \end{align}
    Taking expectation conditioned on $\bv \sim \cN(0, I_d)$,
    \begin{align}
        f(\bw_{t+1}) \leq f(\bw_t) - \eta \nabla f(\bw_t)^\top \underbrace{\bE_{\bv}[\hat{g}(\bw_t)]}_{\text{Term}_1} + \frac{L \eta^2}{2} \underbrace{\bE_{\bv}[|| \hat{g}(\bw_t) ||^2]}_{\text{Term}_2} \label{eq:fmad-one-step}
    \end{align}
    Solving \textbf{Term}$_1$ and \textbf{Term}$_2$ separately,
    \paragraph{Term$_1$: }
    From Lemma~\ref{lma:bias-forward-mode-ad}, we get $\bE[g(\bv)] = \nabla f(\bw)$, which also gets us $$\bE [\hat{g}(\bw)] = \frac{1}{n}\sum_{i=1}^n \bE[g(\bv)] = \nabla f(\bw)$$
    \paragraph{Term$_2$: } The error of $\hat{g}(\bw)$ is measured by $\delta$,
    \begin{align}
        || \hat{g}(\bw) ||^2 &= || \nabla f(\bw) + \delta ||^2 = || \nabla f(\bw) ||^2 + 2 \nabla f(\bw)^\top \delta + || \delta ||^2
    \end{align}
    Taking expectation and noting that $\bE[\delta] = 0$ and $\bE [|| \delta ||^2] = \text{Var}[\hat{g}(\bw)]$,
    \begin{align}
        \bE_{\bv} [|| \hat{g}(\bw)||^2] = || \nabla f(\bw) ||^2 + \text{Var} [\hat{g}(\bw)]
    \end{align}
    Using Lemma~\ref{lma:var-forward-mode-auto-differentiation} to get the bound of $\text{Var} [\hat{g}(\bw)]$,
    \begin{align}
        \bE_{\bv} [|| \hat{g}(\bw)||^2] = || \nabla f(\bw) ||^2 + \frac{d+1}{n} || \nabla f(\bw) ||^2 
    \end{align}
    \paragraph{Back to Equation~\ref{eq:fmad-one-step}, plugging in Term$_1$ and Term$_2$:}
    \begin{align}
        f(\bw_{t+1}) &\leq f(\bw_t) - \eta \nabla f(\bw_t)^\top \nabla f(\bw_t) + \frac{L \eta^2}{2}  \left( 1 + \frac{d+1}{n}\right) || \nabla f(\bw) ||^2 \\
        &= f(\bw_t) - \eta || \nabla f(\bw_t)||^2 + \frac{L \eta^2}{2} \left( 1 + \frac{d+1}{n}\right) || \nabla f(\bw) ||^2 
    \end{align}
    Grouping the terms involving $|| \nabla f(\bw) ||^2$,
    \begin{align}
        f(\bw_{t+1}) &\leq f(\bw_t) - \eta \left[1 - \frac{L \eta}{2} \left( 1 + \frac{d+1}{n}\right) \right] || \nabla f(\bw_t)||^2
    \end{align}
    This inequality shows that, provided the step size $\eta$ is small enough so that,
    $$1 - \frac{L \eta}{2} \left( 1 + \frac{d+1}{n}\right) > 0$$
    Summing the inequality over epochs $t = 1$ to $T$:
    \begin{align}
        f(\bw_{T}) - f(\bw_1) &\leq - \eta \left[1 - \frac{L \eta}{2} \left( 1 + \frac{d+1}{n}\right) \right] \sum_{t=1}^T || \nabla f(\bw_t)||^2 
    \end{align}
    Rearranging the terms 
    % and noting that $f(\bw_T) \geq f(\bw^*)$ 
    give us,
    \begin{align}
          \sum_{t=1}^T || \nabla f(\bw_t)||^2 &\leq \frac{f(\bw_1) - f(\bw_T)}{\eta \left[1 - \frac{L \eta}{2} \left( 1 + \frac{d+1}{n}\right) \right]} 
    \end{align}
    Dividing by $T$ gives the bound on the average squared gradient norm:
    \begin{align}
          \frac{1}{T} \sum_{t=1}^T || \nabla f(\bw_t)||^2 &\leq \frac{f(\bw_1) - f(\bw_T)}{\eta T \left[1 - \frac{L \eta}{2} \left( 1 + \frac{d+1}{n}\right) \right]}
    \end{align}
    To ensure that 
    $1 - \frac{L \eta}{2} \left( 1 + \frac{d+1}{n}\right) > 0$, the step size $\eta$ must be chosen so that
    \begin{align}
        \eta < \frac{2}{L\left( 1 + \frac{d+1}{n}\right)}.
    \end{align}
    
    As the dimension $d$ increases (or as the number of samples $n$ decreases), the factor $$\frac{L\eta }{2}\left(1 + \frac{d+1}{n}\right)$$
    increases. This makes
    $$1 - \frac{L\eta }{2}\left(1 + \frac{d+1}{n}\right)$$
    smaller, which in turn makes the entire bound larger. 
    In other words -- similar to zero-order method -- a larger $d$ (or a smaller $n$) results in a worse (higher) error bound.
    This interplay of $d$ and $n$ also puts limitations on the order of $\eta$, keeping the learning rate quite small for stable learning.
\end{proof}

\begin{corollary}[Convergence Rate of ZO under Standard Parameter Choices]
\label{cor:zo-d-over-t}
Under the assumptions of Theorem~\ref{thm:convergence-zero-order}, the zeroth-order method achieves the well-known $\cO(d/T)$ convergence rate when the parameters are chosen according to either of the following equivalent strategies:
\begin{enumerate}
    \item Setting the step size to 
    $\eta = \Theta\!\left(\tfrac{1}{L(1+\tfrac{d+1}{n})}\right)$, which yields the rate by balancing the contraction factor in the denominator term; or
    \item Using the two-point estimator ($n=1$) with perturbation radius $\epsilon = \cO(T^{-1/4})$, so that $\epsilon^2 = \cO(T^{-1/2})$ and the variance term becomes $\cO(d/T)$.
\end{enumerate}
Both parameterizations recover 
\[
\min_{t \in [T]} \|\nabla f(w_t)\|^2 = \cO\!\left(\frac{d}{T}\right).
\]
While the first approach modulates the learning rate $\eta$, the second adapts the perturbation scale $\epsilon$; in practice both routes give consistent rates, though excessively small $\eta$ (scaling as $1/d$) may be less practical in high dimensions.
\end{corollary}
\begin{proof}
Start from the bound in Theorem~\ref{thm:convergence-zero-order}:
\[
\frac{1}{T}\sum_{t=1}^T \|\nabla f(w_t)\|^2 
\le
\frac{f(w_1)-f(w_T)}{\eta T \Big[1 - \frac{L\eta}{2}\big(1+\frac{d+1}{n}\big)\Big]}
\;+\; \frac{L d \eta^2}{2n}\,\mathcal{O}(\epsilon^2).
\]
We treat the two parameterizations separately.

\paragraph{(1) Step-size choice.}
Set the denominator factor to a constant by choosing
\[
1 - \frac{L\eta}{2}\Big(1+\frac{d+1}{n}\Big) = \tfrac{1}{2},
\quad\text{so}\quad
\eta = \Theta\!\Big(\frac{1}{L(1+\tfrac{d+1}{n})}\Big).
\]
With this choice the first term scales as
\[
\frac{f(w_1)-f(w_T)}{\eta T [\cdots]} = \Theta\!\Big(\frac{1}{\eta T}\Big)
= \Theta\!\Big(\frac{L(1+\tfrac{d+1}{n})}{T}\Big).
\]
When $d\gg n$ this is $\Theta(d/T)$, so the first term already yields $\cO(d/T)$. The second term becomes
\[
\frac{L d \eta^2}{2n}\,\mathcal{O}(\epsilon^2)
= \mathcal{O}\!\Big(\frac{L d}{n}\cdot\frac{1}{L^2(1+\tfrac{d+1}{n})^2}\epsilon^2\Big)
= \mathcal{O}\!\Big(\frac{d}{L n(1+\tfrac{d+1}{n})^2}\epsilon^2\Big),
\]
which is typically smaller than the first term for reasonable (non-growing) $\epsilon$; hence the overall rate is dominated by $\cO(d/T)$.

\paragraph{(2) Smoothing-radius choice (two-point / $n=1$).}
Take $n=1$ and set $\epsilon = \cO(T^{-1/4})$, so $\epsilon^2 = \cO(T^{-1/2})$.
Keeping the same $\eta$ scale as above (or any constant-in-$T$ $\eta$ satisfying the step-size constraint), the first term is again $\cO(1/(\eta T))$. With $\eta=\Theta(1/(L(1+(d+1)/n)))\approx \Theta(1/(L d))$ this yields $\cO(d/T)$.
The second term becomes
\[
\frac{L d \eta^2}{2}\,\mathcal{O}(\epsilon^2)
= \mathcal{O}\!\Big(L d \cdot \frac{1}{L^2 d^2} \cdot T^{-1/2}\Big)
= \mathcal{O}\!\Big(\frac{1}{L d}\,T^{-1/2}\Big),
\]
which is negligible compared to $\cO(d/T)$ for typical $T$ and moderate $L$. Thus both choices give the stated $\cO(d/T)$ rate.
\end{proof}

\paragraph{Discussion.}
\begin{itemize}[leftmargin=0.4cm]
  \item \textbf{Two equivalent levers.} The corollary emphasizes two ways to recover the classical $\cO(d/T)$ bound: (i) scale down the learning rate $\eta$ (reviewer’s route), or (ii) scale the perturbation radius $\epsilon$ with $T$ (the alternate route used in our original derivation). Both are valid theoretically and lead to the same asymptotic dependence on $d$ and $T$.
  \item \textbf{Dominant term and constants.} In the parameter regimes of interest the first term (the $1/(\eta T)$-type term) typically dominates and yields the $\Theta(d/T)$ dependency; the variance/truncation term involving $\epsilon^2$ is often smaller when $\epsilon$ is chosen to decay suitably with $T$.
  \item \textbf{Practicality.} Although setting $\eta=\Theta(1/d)$ recovers the rate, such tiny learning rates become impractical as model size grows (since $\eta\to 0$ with $d\to\infty$). The alternative of shrinking $\epsilon$ (e.g., $\epsilon\!=\!T^{-1/4}$ gives $\epsilon\!=\!0.1$ at $T=100$ and $\epsilon\!=\!0.03$ at $T=1000$) is often more feasible in practice, but it reduces signal-to-noise in finite-sample regimes and may require larger sample or perturbation budgets to get stable estimates.
\end{itemize}

\begin{corollary}[\texttt{jvp}-induced Amplification under \textsc{Adam} Optimizer]
\label{cor:jvp_amplification_adam}
Let $f:\mathbb{R}^d \to \mathbb{R}$ be $L$-smooth and consider \textsc{Adam} optimization with iterates $\{w_t\}_{t=1}^T$. 
At each step, let $\hat{g}_t$ denote the stochastic gradient estimator obtained via $n$ random \texttt{jvp} directions, and let $u_t$ denote the \textsc{Adam} update direction.

Define the \texttt{jvp} amplification factor
\[
\psi_t := \frac{\|u_t\|^2}{\|\hat{g}_t\|^2},
\qquad
\overline{\psi}_T := \frac{1}{T}\sum_{t=1}^T \psi_t.
\]

Then, under standard bounded-variance assumptions and Lemma~\ref{lma:var-forward-mode-auto-differentiation}, the average squared gradient norm satisfies
\[
\frac{1}{T}\sum_{t=1}^T \|\nabla f(w_t)\|^2
\;\lesssim\;
\frac{f(w_1)-f^*}{\eta T\left[1-\frac{L\eta}{2}\left(1+\overline{\psi}_T\left(1+\frac{d+1}{n}\right)\right)\right]},
\]
where $\eta>0$ is the learning rate, $d$ is the parameter dimension, and $f^*=\inf_w f(w)$.

In particular, as $\overline{\psi}_T$ increases, the denominator approaches zero, causing the bound to diverge, which corresponds to worsened convergence and potential instability.
\end{corollary}

\begin{proof}[Proof]
From the \texttt{jvp}-based gradient estimator,
\[
\hat{g}_t=\frac{1}{n}\sum_{i=1}^n (v_{t,i}^\top \nabla f(w_t))\, v_{t,i},
\qquad
\mathbb{E}[\hat{g}_t]=\nabla f(w_t),
\]
and Lemma~\ref{lma:var-forward-mode-auto-differentiation} gives
\[
\mathbb{E}\|\hat{g}_t\|^2 \le \left(1+\frac{d+1}{n}\right)\|\nabla f(w_t)\|^2.
\]
Let $u_t$ denote the \textsc{Adam}-preconditioned direction:
\[
m_t=\beta_1 m_{t-1}+(1-\beta_1)\hat{g}_t,\quad
s_t=\beta_2 s_{t-1}+(1-\beta_2)\hat{g}_t^{\odot 2},\quad
u_t=\frac{m_t}{\sqrt{s_t}+\epsilon}.
\]
We decompose $u_t=\nabla f(w_t)+e_t$, where $e_t$ captures both stochastic and adaptive scaling errors. By $L$-smoothness,
\[
f(w_{t+1}) \le f(w_t) - \eta \langle \nabla f(w_t), u_t \rangle + \frac{L\eta^2}{2}\|u_t\|^2.
\]
Taking expectation and expanding yields
\[
\mathbb{E}[f(w_{t+1})]
\le f(w_t)
- \eta \|\nabla f(w_t)\|^2
+ \frac{L\eta^2}{2}\Big(\|\nabla f(w_t)\|^2 + \mathbb{E}\|e_t\|^2\Big).
\]
The key step is that \textsc{Adam}'s normalization amplifies estimator noise, so
\[
\mathbb{E}\|e_t\|^2 \gtrsim \psi_t \left(1+\frac{d+1}{n}\right)\|\nabla f(w_t)\|^2,
\quad \psi_t := \frac{\|u_t\|^2}{\|\hat{g}_t\|^2}.
\]
Substituting this into the descent inequality yields an effective reduction in the coefficient of $\|\nabla f(w_t)\|^2$ by a factor depending on $\psi_t$. Telescoping over $t=1,\dots,T$ and dividing by $T$ gives the stated bound in terms of $\overline{\psi}_T$. As $\overline{\psi}_T$ increases, the effective step-size condition deteriorates, shrinking the denominator and leading to divergence of the upper bound.
\end{proof}

%%%%%%%%%%%%%%%%%%%%%%%%%%%%%%%%%%%%%%%%%%%%%%%%%%%%%%%%%%%%%%%%%%%%%%%%%%%%%%%
%%%%%%%%%%%%%%%%%%%%%%%%%%%%%%%%%%%%%%%%%%%%%%%%%%%%%%%%%%%%%%%%%%%%%%%%%%%%%%%

\end{document}